\documentclass{article}

\usepackage{microtype}
\usepackage{graphicx}
\usepackage{subfigure}
\usepackage{booktabs}   

\usepackage{hyperref}

\usepackage{amsmath,bm}
\usepackage{amsthm}
\usepackage{cite}
\usepackage{enumerate}
\usepackage{color, soul}
\usepackage{psfrag}
\usepackage{adjustbox}
\usepackage{amssymb}
\usepackage{multirow}
\usepackage{afterpage}
\usepackage{float}

\usepackage{enumitem}

\usepackage[dvipsnames]{xcolor}
\colorlet{darkgreen}{green!65!black}
\colorlet{darkblue}{blue!75!black}
\colorlet{darkred}{red!80!black}
\definecolor{lightblue}{HTML}{0071bc}
\definecolor{lightgreen}{HTML}{39b54a}

\newenvironment{Itemize}%
{\begin{itemize}[leftmargin=*]%
\setlength{\itemsep}{0pt}%
\setlength{\topsep}{0pt}%
\setlength{\partopsep}{0pt}%
\setlength{\parskip}{0pt}}%
{\end{itemize}}
{\begin{enumerate}%
\setlength{\itemsep}{0pt}%
\setlength{\topsep}{0pt}%
\setlength{\partopsep}{0pt}%
\setlength{\parskip}{0pt}}%
{\end{enumerate}}
\global\long\def\real{\mathbb{R}}

\numberwithin{theorem}{section}
\numberwithin{lemma}{section}
\numberwithin{definition}{section}
\numberwithin{proposition}{section}
\numberwithin{remark}{section}
\numberwithin{cor}{section}

\usepackage[accepted]{icml2021}

\icmltitlerunning{Delving into Deep Imbalanced Regression}

\begin{document}

\twocolumn[
\icmltitle{Delving into Deep Imbalanced Regression}



\icmlsetsymbol{equal}{*}

\begin{icmlauthorlist}
\icmlauthor{Yuzhe Yang}{mit}
\icmlauthor{Kaiwen Zha}{mit}
\icmlauthor{Ying-Cong Chen}{mit}
\icmlauthor{Hao Wang}{rutgers}
\icmlauthor{Dina Katabi}{mit}
\end{icmlauthorlist}

\icmlaffiliation{mit}{MIT Computer Science \& Artificial Intelligence Laboratory}
\icmlaffiliation{rutgers}{Department of Computer Science, Rutgers University}

\icmlcorrespondingauthor{Yuzhe Yang}{yuzhe@mit.edu}

\icmlkeywords{Machine Learning, ICML}

\vskip 0.3in
]



\printAffiliationsAndNotice{}  

\begin{abstract}

Real-world data often exhibit imbalanced distributions, where certain target values have significantly fewer observations. Existing techniques for dealing with imbalanced data focus on targets with categorical indices, i.e., different classes. However, many tasks involve continuous targets, where hard boundaries between classes do not exist.
We define Deep Imbalanced Regression (DIR) as learning from such imbalanced data with continuous targets, dealing with potential missing data for certain target values, and generalizing to the entire target range.
Motivated by the intrinsic difference between categorical and continuous label space, we propose distribution smoothing for both labels and features, which explicitly acknowledges the effects of nearby targets, and calibrates both label and learned feature distributions.
We curate and benchmark large-scale DIR datasets from common real-world tasks in computer vision, natural language processing, and healthcare domains. Extensive experiments verify the superior performance of our strategies.
Our work fills the gap in benchmarks and techniques for practical imbalanced regression problems.
Code and data are available at: {\fontsize{9.5}{11.5}\selectfont \url{https://github.com/YyzHarry/imbalanced-regression}}.

\end{abstract}

\vspace{-0.6cm}
\section{Introduction}
\label{sec:intro}
\begin{figure}[tb]
\begin{center}
\includegraphics[width=0.99\linewidth]{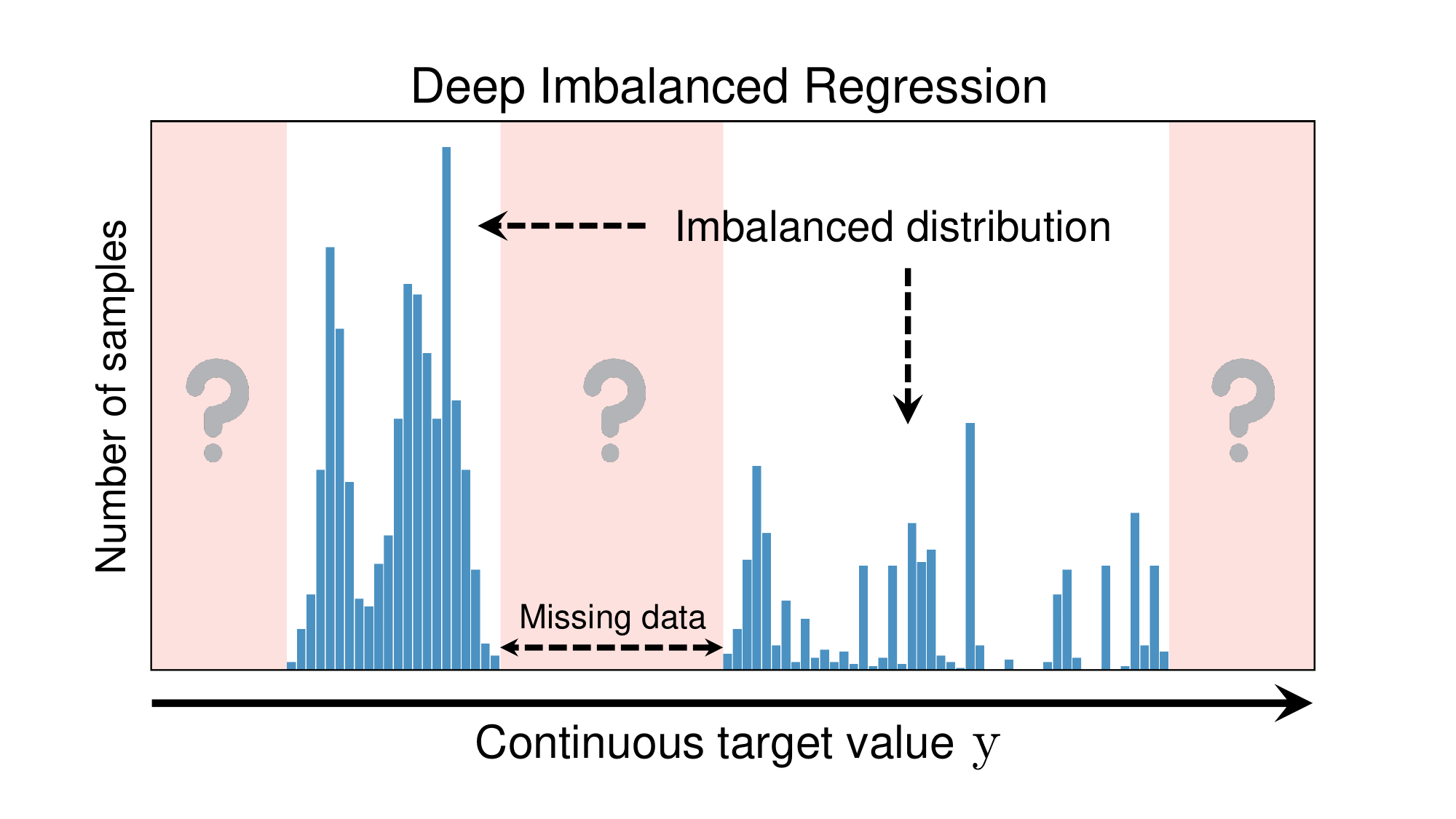}
\end{center}
\vspace{-0.45cm}
\caption{Deep Imbalanced Regression (DIR) aims to learn from imbalanced data with continuous targets, tackle potential missing data for certain regions, and generalize to the entire target range.}
\label{fig:teaser-intro}
\vspace{-0.5cm}
\end{figure}

Data imbalance is ubiquitous and inherent in the real world. Rather than preserving an ideal uniform distribution over each category, the data often exhibit skewed distributions with a long tail~\cite{buda2018systematic,liu2019large}, where certain target values have significantly fewer observations.
This phenomenon poses great challenges for deep recognition models, and has motivated many prior techniques for addressing data imbalance~\cite{huang2019deep,cao2019learning,liu2019large,cui2019class,tang2020long}.

Existing solutions for learning from imbalanced data, however, focus on targets with categorical indices, i.e., the targets are different classes. However, many real-world tasks involve continuous and even infinite target values. For example, in vision applications, one needs to infer the age of different people based on their visual appearances, where age is a continuous target and can be highly imbalanced. Treating different ages as distinct classes is unlikely to yield the best results because it does not take advantage of the similarity between people with nearby ages. Similar issues happen in medical applications since many health metrics including heart rate, blood pressure, and oxygen saturation, are continuous and often have skewed distributions across patient populations.

In this work, we systematically investigate \emph{Deep Imbalanced Regression} (DIR) arising in real-world settings (see Fig.~\ref{fig:teaser-intro}). We define DIR as learning continuous targets from natural imbalanced data, dealing with potentially missing data for certain target values, and generalizing to a test set that is balanced over the entire range of continuous target values. This definition is analogous to the class imbalance problem \cite{liu2019large}, but focuses on the continuous setting.

DIR brings new challenges distinct from its classification counterpart. First, given continuous (potentially infinite) target values, the hard boundaries between classes no longer exist, causing ambiguity when directly applying traditional imbalanced classification methods such as re-sampling and re-weighting. Moreover, continuous labels inherently possess a meaningful distance between targets, which has implication for how we should interpret data imbalance. For example, say two target labels $t_1$ and $t_2$ have a small number of observations in training data. However, $t_1$ is in a highly represented neighborhood (i.e., there are many samples in the range $ [t_1-\Delta, t_1+\Delta]$), while $t_2$ is in a weakly represented neighborhood. In this case, $t_1$ does not suffer from the same level of imbalance as $t_2$.  Finally, unlike classification, certain target values may have no data at all, which motivates the need for target extrapolation \& interpolation.

In this paper, we propose two simple yet effective methods for addressing DIR: label distribution smoothing (LDS) and feature distribution smoothing (FDS). A key idea underlying both approaches is to leverage the similarity between nearby targets by employing a kernel distribution to perform explicit distribution smoothing in the label and feature spaces. Both techniques can be easily embedded into existing deep networks and allow optimization in an end-to-end fashion.
We verify that our techniques not only successfully calibrate for the intrinsic underlying imbalance, but also provide large and consistent gains when combined with other methods.

To support practical evaluation of imbalanced regression, we curate and benchmark large-scale DIR datasets for common real-world tasks in computer vision, natural language processing, and healthcare. They range from single-value prediction such as age, text similarity score, health condition score, to dense-value prediction such as depth. We further set up benchmarks for proper DIR performance evaluation.

Our contributions are as follows:
\vspace{-0.35cm}
\begin{Itemize}
\item We formally define the DIR task as learning from imbal-anced data with continuous targets, and generalizing to the entire target range. DIR provides thorough and unbiased evaluation of learning algorithms in practical settings.
\item We develop two simple, effective, and interpretable algorithms for DIR, LDS and FDS, which exploit the similarity between nearby targets in both label and feature space.
\item We curate benchmark DIR datasets in different domains: computer vision, natural language processing, and healthcare. We set up strong baselines as well as benchmarks for proper DIR performance evaluation.
\item Extensive experiments on large-scale DIR datasets verify the consistent and superior performance of our strategies.
\end{Itemize}

\section{Related Work}
\label{sec:related-work}
\textbf{Imbalanced Classification.}
Much prior work has focused on the imbalanced classification problem (also referred to as long-tailed recognition~\cite{liu2019large}). Past solutions can be divided into data-based and model-based solutions:
Data-based solutions either over-sample the minority class or under-sample the majority~\cite{chawla2002smote,he2008adasyn,garcia2009evolutionary}. For example, SMOTE generates synthetic samples for minority classes by linearly interpolating samples in the same class~\cite{chawla2002smote}. Model-based solutions include re-weighting or adjusting the loss function to compensate for class imbalance \cite{huang2019deep,huang2016learning,cui2019class,cao2019learning,dong2019imbalanced}, and leveraging relevant learning paradigms, including transfer learning \cite{yin2019feature}, metric learning \cite{zhang2017range}, meta-learning \cite{shu2019meta}, and two-stage training \cite{kang2020decoupling}. Recent studies have also discovered that semi-supervised learning and self-supervised learning lead to better imbalanced classification results \cite{yang2020rethinking}. 
In contrast to these past work, we identify the limitations of applying class imbalance methods to regression problems, and introduce new techniques particularly suitable for learning continuous target values.

\textbf{Imbalanced Regression.}
Regression over imbalanced data is not as well explored. Most of the work on this topic is a direct adaptation of the SMOTE algorithm to regression scenarios~\cite{torgo2013smoter,branco2017smogn,branco2018rebagg}. Synthetic samples are created for pre-defined rare target regions by either directly interpolating both inputs and targets \cite{torgo2013smoter}, or {using} Gaussian noise augmentation \cite{branco2017smogn}. A bagging-based ensemble method that incorporates multiple data pre-processing steps has also been introduced~\cite{branco2018rebagg}. However, there exist several intrinsic drawbacks for these methods. First, they fail to take the distance between targets into account, and rather heuristically divide the dataset into rare and frequent sets, then plug in classification-based methods. Moreover, modern data is of extremely high dimension (e.g., images and physiological signals); linear interpolation of two samples of such data does not lead to meaningful new synthetic samples.
Our methods are intrinsically different from past work in their approach. They can be combined with existing methods to improve their performance, as we show in Sec.~\ref{sec:experiment}. Further, our approaches are tested on large-scale real-world datasets in computer vision, NLP, and healthcare.

\vspace{-0.06cm}
\section{Methods}
\label{sec:method}
\textbf{Problem Setting.}
Let $\{ ( \mathbf{x}_i, y_i )\}_{i=1}^{N}$ be a training set, where $\mathbf{x}_i\in\real^{d}$ denotes the input and $y_i\in\real$ is the label, which is a continuous target.
We introduce an additional structure for the label space $\mathcal{Y}$, where we divide $\mathcal{Y}$ into $B$ groups (bins) with equal intervals, i.e., $[y_0, y_1), [y_1, y_2),\dots, [y_{B-1}, y_B)$. Throughout the paper, we use $b\in\mathcal{B}$ to denote the group index of the target value, where $\mathcal{B} = \{1, \dots,B\} \subset \mathbb{Z}^+$ is the index space.
In practice, the defined bins reflect a minimum resolution we care for grouping data in a regression task.
For instance, in age estimation, we could define $\delta y \triangleq y_{b+1} - y_{b} = 1$, showing a minimum age difference of $1$ is of interest.
Finally, we denote $\mathbf{z} = f(\mathbf{x};\theta)$ as the feature for $\mathbf{x}$, where $f(\mathbf{x};\theta)$ is parameterized by a deep neural network model with parameter $\theta$.
The final prediction $\hat{y}$ is given by a regression function $g(\cdot)$ that operates over $\mathbf{z}$.

\subsection{Label Distribution Smoothing}
\label{sec:lds}

\begin{figure}[!t]
\centering
\subfigure[CIFAR-100 (subsampled)]{
    \label{fig:label-error-cifar100}
    \includegraphics[height=0.19\textwidth]{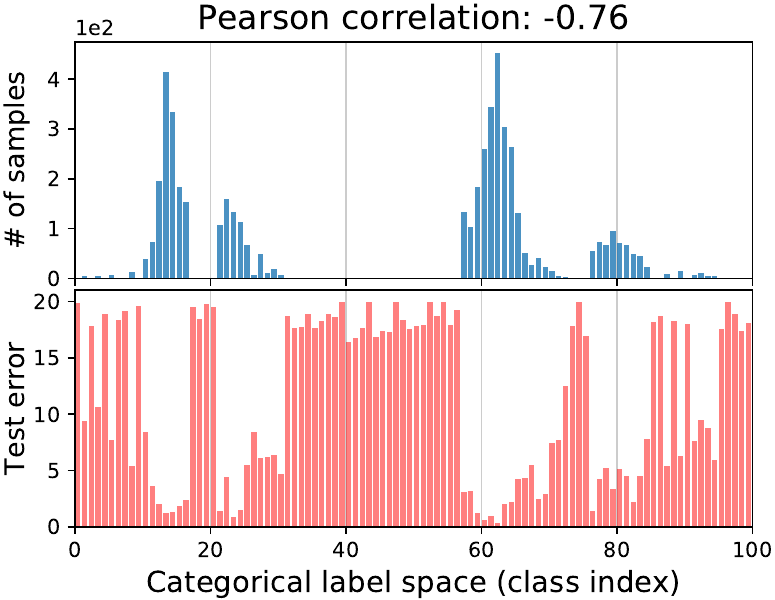}
}
\hspace{-2ex}
\subfigure[IMDB-WIKI (subsampled)]{
    \label{fig:label-error-imdb}
    \includegraphics[height=0.19\textwidth]{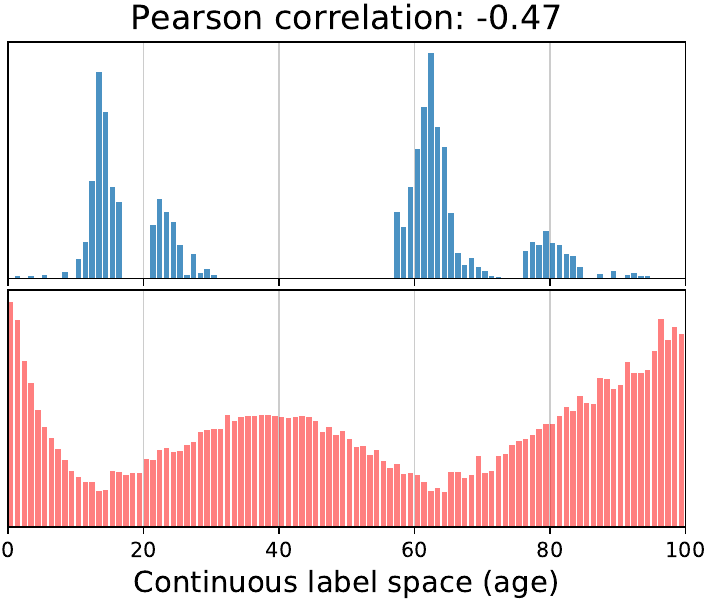}
}
\vspace{-0.5cm}
\caption{Comparison on the test error distribution (bottom) using same training label distribution (top) on two different datasets: (a) CIFAR-100, a classification task with categorical label space. (b) IMDB-WIKI, a regression task with continuous label space.}
\label{fig:lds-motivate-label-error}
\vspace{-0.3cm}
\end{figure}

We start by showing an example to demonstrate the difference between classification and regression when imbalance comes into the picture.

\textbf{Motivating Example.}
We employ two datasets: (1) CIFAR-100~\cite{krizhevsky2009learning}, which is a 100-class classification dataset, and (2) the IMDB-WIKI dataset~\cite{imdb-wiki}, which is a large-scale image dataset for age estimation from visual appearance.
The two datasets have intrinsically different label space: CIFAR-100 exhibits \emph{categorical label space} where the target is class index, while IMDB-WIKI has a \emph{continuous label space} where the target is age. We limit the age range to $0\sim 99$ so that the two datasets have the same label range, and subsample them to simulate data imbalance, while ensuring they have exactly the same label density distribution (Fig.~\ref{fig:lds-motivate-label-error}). We make both test sets balanced. We then train a plain ResNet-50 model on the two datasets, and plot their test error distributions.

We observe from Fig.~\ref{fig:label-error-cifar100} that the error distribution \emph{correlates} with label density distribution. Specifically, the test error as a function of class index has a high negative Pearson correlation with the label density distribution (i.e., $-0.76$) in the categorical label space. The phenomenon is expected, as majority classes with more samples are better learned than minority classes. Interestingly however, as Fig.~\ref{fig:label-error-imdb} shows, the error distribution is very different for IMDB-WIKI with continuous label space, even when the label density distribution is the same as CIFAR-100. In particular, the error distribution is much smoother and no longer correlates well with the label density distribution ($-0.47$).

The reason why this example is interesting is that all {imbalanced learning} methods, directly or indirectly, operate by compensating for the imbalance in the \emph{empirical} label density distribution. This works well for class imbalance, but for continuous labels the empirical density does not accurately reflect the imbalance as seen by the neural network. Hence, compensating for data imbalance based on empirical label density is inaccurate for the continuous label space.

\begin{figure}[!t]
\begin{center}
 \includegraphics[width=0.99\linewidth]{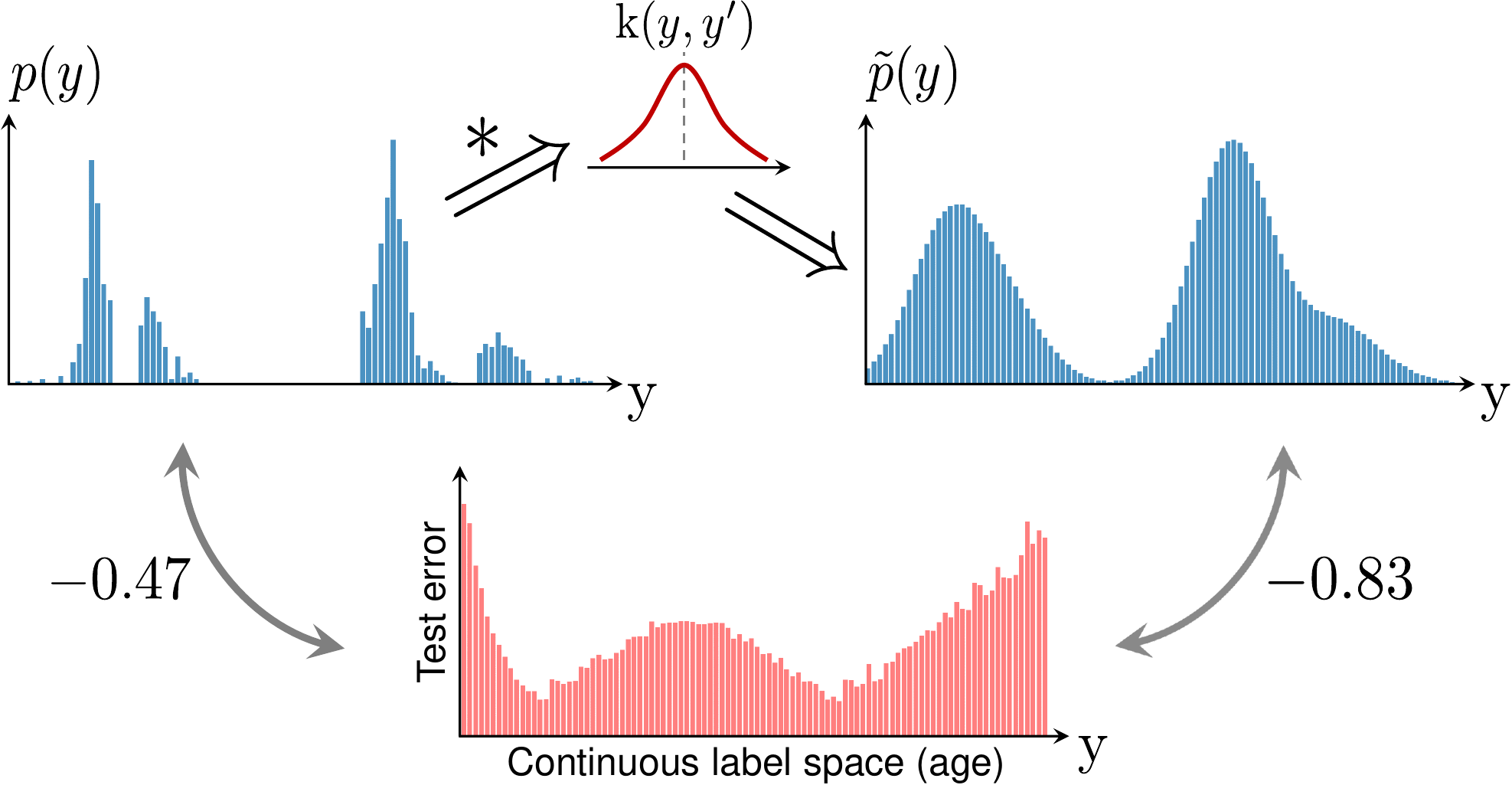}
\end{center}
\vspace{-0.4cm}
\caption{Label distribution smoothing~(LDS) convolves a symmetric kernel with the empirical label density to estimate the effective label density distribution that accounts for the continuity of labels.}
\label{fig:lds-motivate-smooth}
\vspace{-0.3cm}
\end{figure}

\textbf{LDS for Imbalanced Data Density Estimation.}
The above example shows that, in the continuous case, the {empirical} label distribution does not reflect the real label density distribution.
This is because of the dependence between data samples at nearby labels (e.g., images of close ages). In fact, there is a significant literature in statistics on how to estimate the expected density in such cases~\cite{parzen1962estimation}. Thus, Label Distribution Smoothing (LDS) advocates the use of kernel density estimation to learn the effective imbalance in datasets that corresponds to continuous targets. 

LDS convolves a symmetric kernel with the empirical density distribution to extract a kernel-smoothed version that accounts for the overlap in information of data samples of nearby labels. A symmetric kernel is any kernel that satisfies: $\mathrm{k}(y, y') = \mathrm{k}(y', y)$ and $\nabla_{y} \mathrm{k}(y, y') + \nabla_{y'} \mathrm{k}(y', y) = 0$, $\forall y, y'\in \mathcal{Y}$. Note that a Gaussian or a Laplacian kernel is a symmetric kernel, while $\mathrm{k}(y, y') = yy'$ is not. The symmetric kernel characterizes the similarity between target values $y'$ and any $y$ w.r.t. their distance in the target space. Thus, LDS computes the \emph{effective label density distribution} as:
\begin{equation}
    \tilde{p}(y')\triangleq \int_{\mathcal{Y}}\mathrm{k}(y, y') p(y)dy,
\end{equation}
where $p(y)$ is the number of appearances of label of $y$ in the training data, and $\tilde{p}(y')$ is the effective density of label $y'$.  

Fig.~\ref{fig:lds-motivate-smooth} illustrates LDS and how it smooths the label density distribution. Further, it shows that the resulting label density computed by LDS correlates well with the error distribution ($-0.83$). This demonstrates that LDS captures the real imba-lance that affects regression problems.

Now that the effective label density is available, techniques for addressing class imbalance problems can be directly adapted to the DIR context. For example, a straightforward adaptation can be the cost-sensitive re-weighting method, where we re-weight the loss function by multiplying it by the inverse of the LDS estimated label density for each target. We show in Sec.~\ref{sec:experiment} that LDS can be seamlessly incorporated with a wide range of techniques to boost DIR performance.

\subsection{Feature Distribution Smoothing}
\label{sec:fds}

We are motivated by the intuition that continuity in the target space should create a corresponding continuity in the feature space. That is, if the model works properly and the data is balanced, one expects the feature statistics corresponding to nearby targets to be close to each other. 

\begin{figure}[!t]
\begin{center}
 \includegraphics[width=0.99\linewidth]{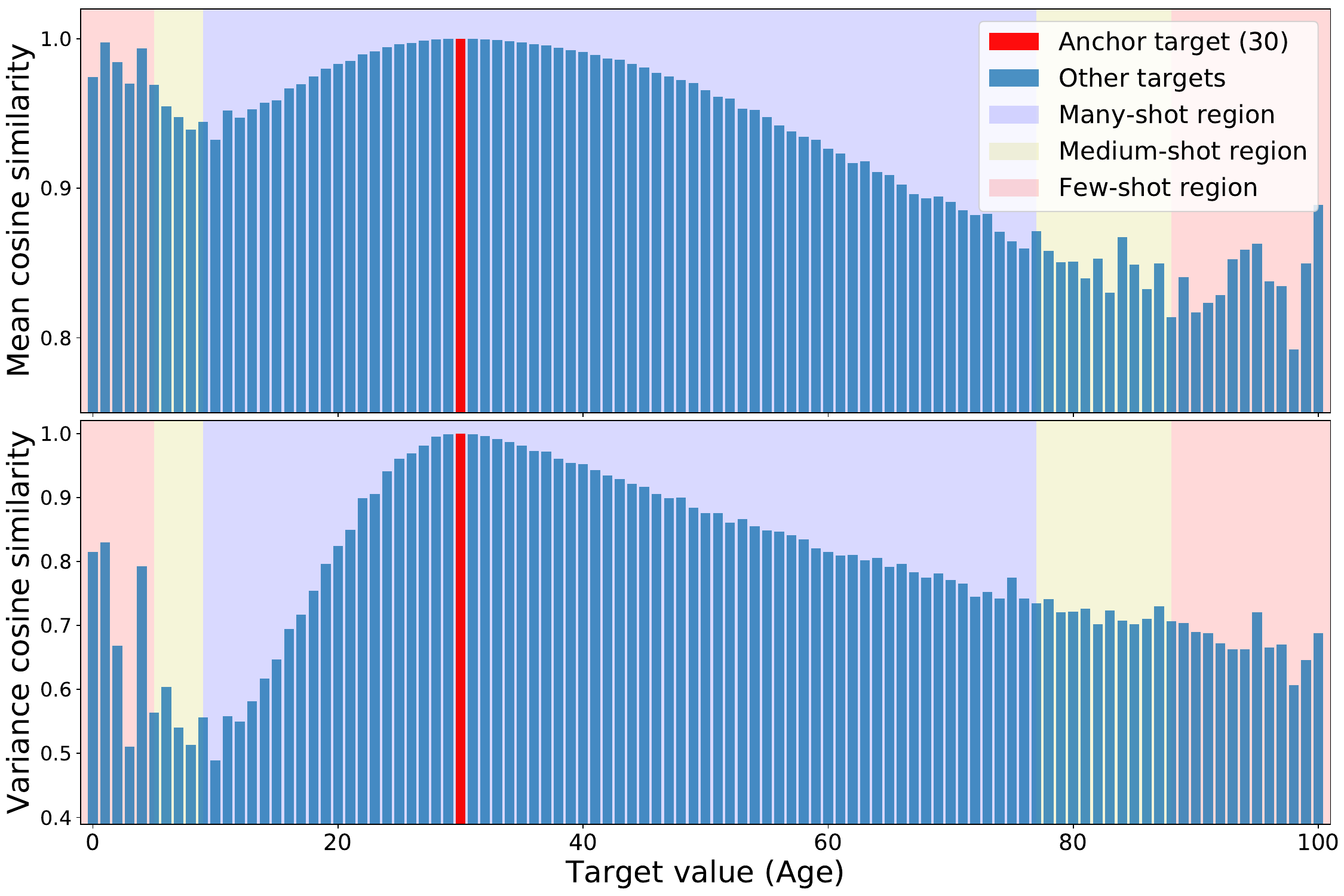}
\end{center}
\vspace{-0.5cm}
\caption{Feature statistics similarity for age 30. \textbf{Top:} Cosine simi-larity of the feature mean at a particular age w.r.t. its value at the anchor age. \textbf{Bottom:} Cosine similarity of the feature variance at a particular age w.r.t. its value at the anchor age. The color of the background refers to the data density in a particular target range. The figure shows that nearby ages have close similarities; However, it also shows that there is unjustified similarity between images at  ages 0 to 6 and age 30, due to data imbalance.}
\label{fig:fds-motivate}
\vspace{-0.4cm}
\end{figure}

\textbf{Motivating Example.}
We use an illustrative example to highlight the impact of data imbalance on feature statistics in DIR. Again, we use a plain model trained on the images in the IMDB-WIKI dataset to infer a person's age from visual appearance.
We  focus on the learned feature space, i.e., $\mathbf{z}$. We use a minimum bin size of $1$, i.e., $y_{b+1} - y_{b} = 1$, and group features with the same target value in the same bin. We then compute the feature statistics (i.e., mean and variance) with respect to the data in each bin, which we denote as $\{\boldsymbol{\mu}_b,\boldsymbol{\sigma}_b\}_{b=1}^{B}$. To visualize the similarity between feature statistics, we select an anchor bin $b_0$, and calculate the cosine similarity of the feature statistics between $b_0$ and all other bins. The results are summarized in Fig.~\ref{fig:fds-motivate} for $b_0=30$. The figure also shows the regions with different data densities using the colors purple, yellow, and pink.

Fig.~\ref{fig:fds-motivate} shows that the feature statistics around $b_0=30$ are highly similar to their values at $b_0=30$. Specifically, the cosine similarity of the feature mean and feature variance for all bins between age 25 and 35 are within a few percent from their values at age 30 (the anchor age). Further, the similarity gets higher for tighter ranges around the anchor. Note that bin 30 falls in the high shot region. In fact, it is among the few bins that have the most samples. So, the figure confirms the intuition that when there is enough data, and for continuous targets, the feature statistics are similar to nearby bins. Interestingly, the figure also shows the problem with regions that have very few data samples, like the age range 0 to 6 years (shown in pink). Note that the mean and variance in this range show unexpectedly high similarity to age 30. In fact, it is shocking that the feature statistics at age 30 are more similar to age 1 than age 17. This unjustified similarity is due to data imbalance. Specifically, since there are not enough images for ages 0 to 6, this range thus inherits its priors from the range with the maximum amount of data, which is the range around age 30.

\textbf{FDS Algorithm.}
Inspired by these observations, we propose feature distribution smoothing (FDS), which performs distribution smoothing on the feature space, i.e., transfers the feature statistics between nearby target bins. This procedure aims to calibrate the potentially biased estimates of feature distribution, especially for underrepresented target values (e.g., medium- and few-shot groups) in training data.

FDS is performed by first estimating the statistics of each bin. Without loss of generality, we substitute variance with covariance to reflect also the relationship between the various feature elements within  $\mathbf{z}$:
\begin{align}
    \boldsymbol{\mu}_b & = \frac{1}{N_b}\sum_{i=1}^{N_b} \mathbf{z}_i, \label{eq:estimate-run-stats-mu} \\
    \boldsymbol{\Sigma}_b & = \frac{1}{N_b-1} \sum_{i=1}^{N_b} (\mathbf{z}_i - \boldsymbol{\mu}_b) (\mathbf{z}_i -  \boldsymbol{\mu}_b)^\top, \label{eq:estimate-run-stats-sigma}
\end{align}
where $N_b$ is the total number of samples in $b$-th bin. Given the feature statistics, we employ again a symmetric kernel $\mathrm{k}(y_{b}, y_{b'})$ to smooth the distribution of the feature mean and covariance over the target bins $\mathcal{B}$. This results in a smoothed version of the statistics:
\begin{align}
    \tilde{\boldsymbol{\mu}}_b & = \sum_{b'\in\mathcal{B}} \mathrm{k}(y_{b}, y_{b'}) \boldsymbol{\mu}_{b'}, \label{eq:smooth-stats-mu} \\
    \boldsymbol{\widetilde{\Sigma}}_b & = \sum_{b'\in\mathcal{B}} \mathrm{k}(y_{b}, y_{b'}) \boldsymbol{\Sigma}_{b'}. \label{eq:smooth-stats-sigma}
\end{align}
With both $\{{\boldsymbol{\mu}}_b, \boldsymbol{{\Sigma}}_b\}$ and $\{\tilde{\boldsymbol{\mu}}_b, \boldsymbol{\widetilde{\Sigma}}_b\}$, we then follow the standard whitening and re-coloring procedure~\cite{sun2016return} to calibrate the feature representation for each input sample:
\begin{align}
   &\tilde{\mathbf{z}} = \boldsymbol{\widetilde{\Sigma}}_b^{\frac{1}{2}} \boldsymbol{\Sigma}_b^{-\frac{1}{2}} (\mathbf{z} - \boldsymbol{\mu}_b) + \tilde{\boldsymbol{\mu}}_b. \label{eq:calibrate-stats}
\end{align}
We integrate FDS into deep networks by inserting a feature calibration layer after the final feature map. To train the model, we employ a \emph{momentum update} of the running statistics $\{{\boldsymbol{\mu}}_b, \boldsymbol{{\Sigma}}_b\}$ across each epoch. Correspondingly, the smoothed statistics $\{\tilde{\boldsymbol{\mu}}_b, \boldsymbol{\widetilde{\Sigma}}_b\}$ are updated across different epochs but fixed within each training epoch. The momentum update, which performs an exponential moving average (EMA) of running statistics, results in more stable and accurate estimations of the feature statistics during training. The calibrated features $\tilde{\mathbf{z}}$ are then passed to the final regression function and used to compute the loss.

\begin{figure}[!t]
\begin{center}
 \includegraphics[width=0.99\linewidth]{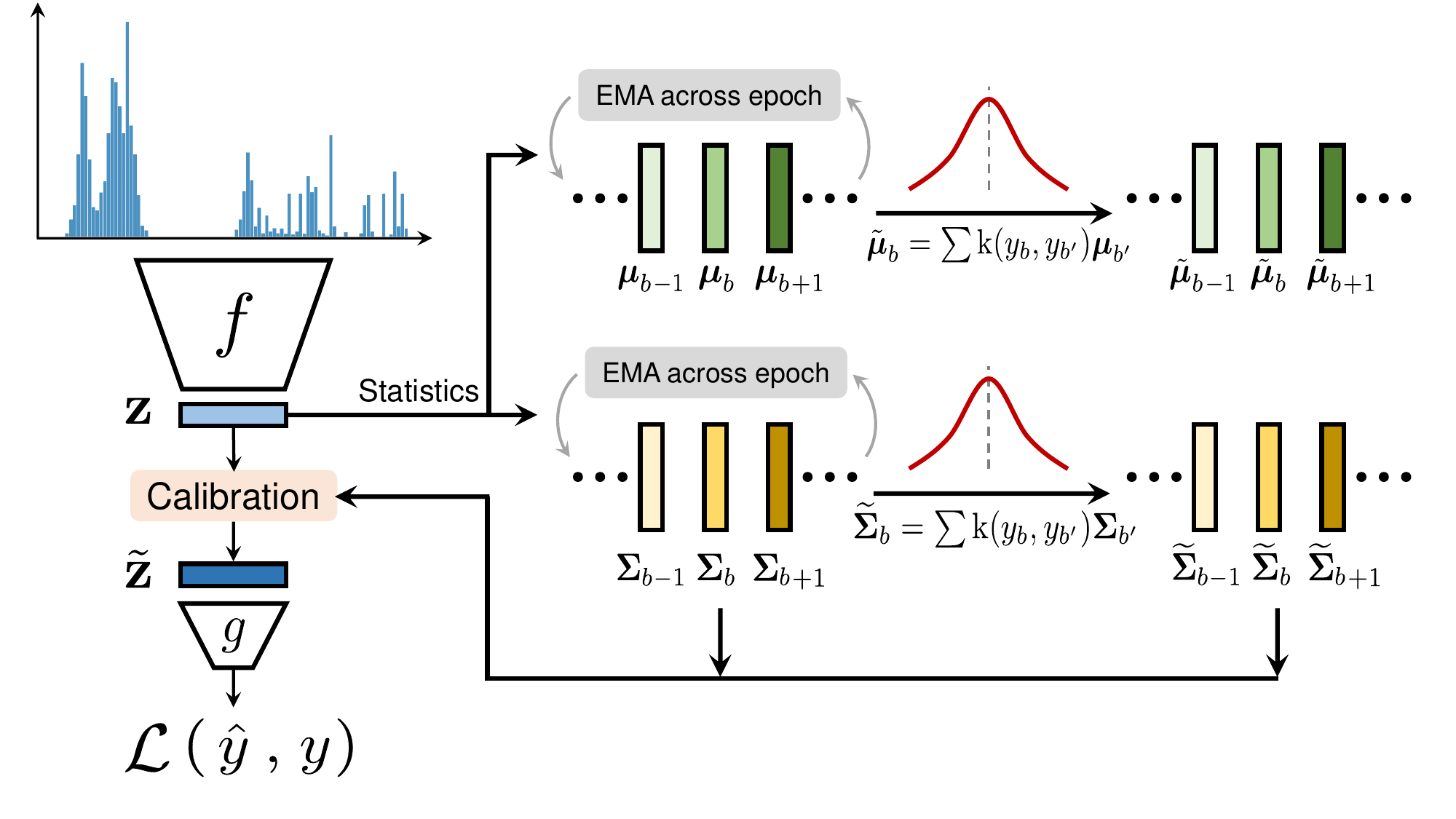}
\end{center}
\vspace{-0.4cm}
\caption{Feature distribution smoothing (FDS) introduces a feature calibration layer that uses kernel smoothing to smooth the distributions of feature mean and covariance over the target space.}
\label{fig:teaser-fds}
\vspace{-0.3cm}
\end{figure}

We note that FDS can be integrated with any neural network model, as well as any past work on improving label imbalance. In Sec.~\ref{sec:experiment}, we integrate FDS with a variety of prior techniques for addressing data imbalance, and demonstrate that it consistently improves performance.

\section{Benchmarking DIR}
\label{sec:experiment}
\begin{figure*}[!t]
\begin{center}
 \includegraphics[width=0.99\linewidth]{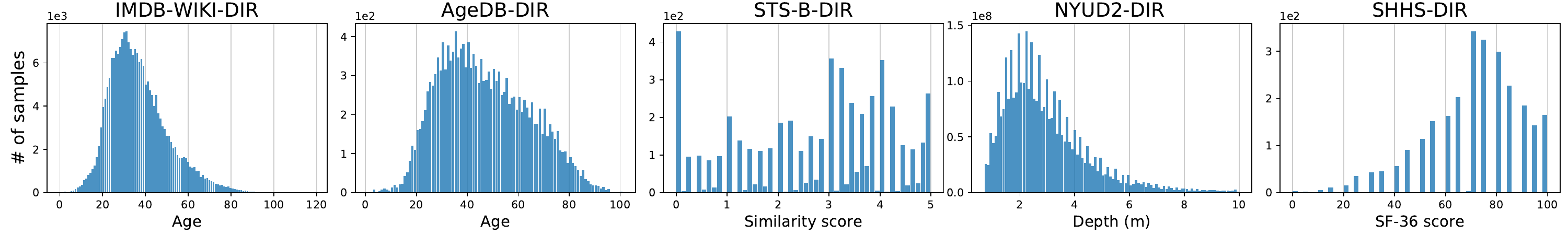}
\end{center}
\vspace{-0.6cm}
\caption{Overview of training set label distribution for five DIR datasets. They range from single-value prediction such as age, textual similarity score, and health condition score, to dense-value prediction such as depth estimation. More details are provided in Appendix~\ref{appendix:dataset-details}.}
\label{fig:dataset-info}
\vspace{-0.4cm}
\end{figure*}

\textbf{Datasets.}
We curate five DIR benchmarks that span computer vision, natural language processing, and healthcare. Fig.~\ref{fig:dataset-info} shows the label density distribution of these datasets, and their level of imbalance.
\begin{itemize}[leftmargin=*]
    \item \emph{IMDB-WIKI-DIR (age)}: We construct IMDB-WIKI-DIR using the IMDB-WIKI dataset~\cite{imdb-wiki}, which contains 523.0K face images and the corresponding ages. We filter out unqualified images, and manually construct balanced validation and test set over the supported ages. The length of each bin is $1$ year, with a minimum age of $0$ and a maximum age of $186$. The number of images per bin varies between $1$ and $7149$, exhibiting significant data imbalance. Overall, the curated dataset has 191.5K images for training, 11.0K images for validation and testing.
    \item \emph{AgeDB-DIR (age)}: AgeDB-DIR is constructed in a similar manner from the AgeDB dataset~\cite{moschoglou2017agedb}. It contains 12.2K images for training, with a minimum age of $0$ and a maximum age of $101$, and maximum bin density of $353$ images and minimum bin density of $1$. The validation and test set are balanced with 2.1K images.
    \item \emph{STS-B-DIR (text similarity score)}: We construct STS-B-DIR from the Semantic Textual Similarity Benchmark \cite{cer2017semeval, wang2018glue}, which is a collection of sentence pairs drawn from news headlines, video and image captions, and natural language inference data. Each pair is annotated by multiple annotators with an averaged continuous similarity score from $0$ to $5$. From the original training set of 7.2K pairs, we create a training set with 5.2K pairs, and balanced validation set and test set of 1K pairs each. The length of each bin is $0.1$. 
    \item \emph{NYUD2-DIR (depth)}: We create NYUD2-DIR based on the NYU Depth Dataset V2~\cite{Silberman:ECCV12}, which provides images and depth maps for different indoor scenes. The depth maps have an upper bound of $10$ meters and we set the bin length as $0.1$ meter. Following standard practices~\cite{hu2019revisiting, bhat2020adabins}, we use 50K images for training and 654 images for testing. We randomly select $9357$ test pixels for each bin to make the test set balanced.
    \item \emph{SHHS-DIR (health condition score)}: We create SHHS-DIR based on the SHHS dataset \cite{quan1997sleep}, which contains full-night Polysomnography (PSG) from $2651$ subjects. Available PSG signals include Electroencephalography (EEG), Electrocardiography (ECG), and breathing signals (airflow, abdomen, and thorax), which are used as inputs. The dataset also includes the 36-Item Short Form Health Survey (SF-36) \cite{ware1992mos} for each subject, where a General Health score is extracted. The score is used as the target value with a minimum score of $0$ and maximum of $100$.
\end{itemize}

\textbf{Network Architectures.}
We employ ResNet-50~\cite{he2016deep} as our backbone network for IMDB-WIKI-DIR and AgeDB-DIR.
Following~\cite{wang2018glue}, we adopt the same BiLSTM + GloVe word embeddings baseline for STS-B-DIR. For NYUD2-DIR, we use ResNet-50-based encoder-decoder architecture introduced in~\cite{hu2019revisiting}. Finally, for SHHS-DIR, we use the same CNN-RNN architecture with ResNet block for PSG signals as in~\cite{wang2019bidirectional}.

\textbf{Baselines.}
Since the literature has only a few proposals for DIR, in addition to past work on imbalanced regression \cite{torgo2013smoter,branco2017smogn}, we adapt a few imbalanced classification methods for regression, and propose a strong set of baselines.   Below, we describe the baselines, and how we can combine LDS with each method. For FDS,  it can be directly integrated with any baseline as a calibration layer, as described in Sec.~\ref{sec:fds}.
\begin{itemize}[leftmargin=*]
    \item \emph{Vanilla model}: We use term \textbf{VANILLA} to denote a model that does not include any technique for dealing with imbalanced data. To combine the vanilla model with LDS, we re-weight the loss function by multiplying it by the inverse of the LDS estimated density for each target bin. 
    \item \emph{Synthetic samples}: We choose existing methods for imbalanced regression, including \textbf{SMOTER}~\cite{torgo2013smoter} and \textbf{SMOGN}~\cite{branco2017smogn}. SMOTER first defines frequent and rare regions using the original label density, and creates synthetic samples for pre-defined rare regions by linearly interpolating both inputs and targets. SMOGN further adds Gaussian noise to SMOTER. {We note that LDS can be directly used for a better estimation of label density when dividing the target space.}
    \item \emph{Error-aware loss}: Inspired by the Focal loss~\cite{lin2017focal} for classification, we propose a regression version called \textbf{Focal-R}, where the scaling factor is replaced by a continuous function that maps the absolute error into $[0,1]$. Precisely, Focal-R loss based on $L_1$ distance can be written as $\frac{1}{n} \sum_{i=1}^{n} \sigma(|\beta e_i|)^{\gamma} e_i$, where $e_i$ is the $L_1$ error for $i$-th sample, $\sigma(\cdot)$ is the \texttt{Sigmoid} function, and $\beta,\gamma$ are hyper-parameters. To combine Focal-R with LDS, we multiply the loss with the inverse frequency of the estimated label density.
    \item \emph{Two-stage training}: Following~\cite{kang2020decoupling} where feature and classifier are decoupled and trained in two stages, we propose a regression version called regressor re-training (\textbf{RRT}), where in the first stage we train the encoder normally, and in the second stage freeze the encoder and re-train the regressor $g(\cdot)$ with inverse re-weighting. When adding LDS, the re-weighting in the second stage is based on the label density estimated through LDS.
    \item \emph{Cost-sensitive re-weighting}: Since we divide the target space into finite bins, classic re-weighting methods can be directly plugged in. We adopt two re-weighting schemes based on the label distribution: inverse-frequency weighting (\textbf{INV}) and its square-root weighting variant (\textbf{SQINV}). When combining with LDS, instead of using the original label density, we use the LDS estimated target density.  
\end{itemize}

\textbf{Evaluation Process and Metrics.}
Following~\cite{liu2019large}, we divide the target space into three disjoint subsets: \emph{many-shot region} (bins with over 100 training samples), \emph{medium-shot region} (bins with 20$\sim$100 training samples), and \emph{few-shot region} (bins with under 20 training samples), and report results on these subsets, as well as overall performance. 
We also refer to regions with no training samples as \emph{zero-shot}, and investigate the ability of our techniques to  generalize to zero-shot regions in Sec.~{\ref{sec:exp-further-analysis}}.
For metrics, we use common metrics for regression, such as the mean-average-error (MAE), mean-squared-error (MSE), and Pearson correlation. We further propose another metric, called error Geometric Mean (\textbf{GM}), and is defined as $(\prod_{i=1}^n e_i)^{\frac{1}{n}}$ for better prediction fairness.

\subsection{Main Results}
\label{sec:exp-main-results}
We report the main results in this section for all DIR datasets. All training details, 
hyper-parameter settings, and additional results are provided in Appendix~\ref{appendix:experiment-details} and \ref{appendix:additional-results}.

\textbf{Inferring Age from Images: IMDB-WIKI-DIR \& AgeDB-DIR.}
We report the performance of different methods in Table~\ref{table:imdb-wiki} and \ref{table:agedb}, respectively. For each dataset, we group the baselines into four sections to reflect their different strategies. First, as both tables indicate, when applied to modern high-dimensional data like images, SMOTER and SMOGN can actually degrade the performance in comparison to the vanilla model.
Moreover, within each group, adding either LDS, FDS, or both leads to performance gains, while LDS $+$ FDS often achieves the best results. Finally, when compared to the vanilla model, using our LDS and FDS maintains or slightly improves the performance overall and on the many-shot regions, while substantially boosting the performance for the medium-shot and few-shot regions.

\begin{table}[t]
\setlength{\tabcolsep}{2.5pt}
\caption{Benchmarking results on IMDB-WIKI-DIR.}
\vspace{-4pt}
\label{table:imdb-wiki}
\small
\begin{center}
\resizebox{0.49\textwidth}{!}{
\begin{tabular}{l|cccc|cccc}
\toprule[1.5pt]
Metrics      & \multicolumn{4}{c|}{MAE~$\downarrow$}     & \multicolumn{4}{c}{GM~$\downarrow$}  \\ \midrule
Shot         & All  & Many & Med. & Few   & All  & Many & Med. & Few   \\ \midrule\midrule
\textsc{Vanilla}      & 8.06 & 7.23 & 15.12  & 26.33 & 4.57 & 4.17 & 10.59  & 20.46 \\ \midrule\midrule
\textsc{SmoteR}~\cite{torgo2013smoter}       & 8.14 & 7.42 & 14.15  & 25.28 & 4.64 & \textbf{4.30} & 9.05   & 19.46 \\[1.2pt]
\textsc{SMOGN}~\cite{branco2017smogn}        & 8.03 & \textbf{7.30} & 14.02  & 25.93 & 4.63 & \textbf{4.30} & 8.74   & 20.12 \\[1.2pt]
\textsc{SMOGN} + \textbf{\textsc{LDS}}   & 8.02 & 7.39 & 13.71 & 23.22 & 4.63 & 4.39 & 8.71 & 15.80 \\[1.2pt]
\textsc{SMOGN} + \textbf{\textsc{FDS}}   & 8.03 & 7.35 & 14.06 & 23.44 & 4.65 & 4.33 & 8.87 & 16.00 \\[1.2pt]
\textsc{SMOGN} + \textbf{\textsc{LDS}} + \textbf{\textsc{FDS}}   & \textbf{7.97} & 7.38 & \textbf{13.22} & \textbf{22.95} & \textbf{4.59} & 4.39 & \textbf{7.84} & \textbf{14.94} \\ \midrule\midrule
\textsc{Focal-R}      & 7.97 & 7.12 & 15.14  & 26.96 & 4.49 & 4.10 & 10.37  & 21.20 \\[1.2pt]
\textsc{Focal-R} + \textbf{\textsc{LDS}} & {7.90}  & \textbf{7.10}  &  14.72   & 25.84  & \textbf{4.47}   & \textbf{4.09}    & {10.11}     & {19.14}     \\[1.2pt]
\textsc{Focal-R} + \textbf{\textsc{FDS}}   & 7.96 & 7.14 & 14.71 & 26.06 & 4.51 & 4.12 & 10.16 & 19.56 \\[1.2pt]
\textsc{Focal-R} + \textbf{\textsc{LDS}} + \textbf{\textsc{FDS}}   & \textbf{7.88} & \textbf{7.10} & \textbf{14.08}  & \textbf{25.75}   & \textbf{4.47} & 4.11 & \textbf{9.32} & \textbf{18.67} \\ \midrule\midrule
\textsc{RRT}          & 7.81 & 7.07 & 14.06  & 25.13 & 4.35 & 4.03 & 8.91   & 16.96 \\[1.2pt]
\textsc{RRT} + \textbf{\textsc{LDS}} & {7.79} &  7.08  & {13.76}  & {24.64} & {4.34} & \textbf{4.02} & {8.72} & {16.92} \\[1.2pt]
\textsc{RRT} + \textbf{\textsc{FDS}}   & \textbf{7.65}  & \textbf{7.02} & 12.68 & 23.85 & \textbf{4.31} & 4.03 & 7.58 & 16.28 \\[1.2pt]
\textsc{RRT} + \textbf{\textsc{LDS}} + \textbf{\textsc{FDS}}   & \textbf{7.65}  & 7.06  & \textbf{12.41}  & \textbf{23.51}  & \textbf{4.31} & {4.07} & \textbf{7.17} & \textbf{15.44} \\ \midrule\midrule
\textsc{SQInv}      & 7.87 & 7.24 & 12.44  & 22.76 & 4.47 & 4.22 & 7.25   & 15.10 \\[1.2pt]
\textsc{SQInv} + \textbf{\textsc{LDS}} & {7.83} & 7.31 & \textbf{12.43}  & {22.51} & {4.42} & {4.19} & {7.00}  & {13.94} \\[1.2pt]
\textsc{SQInv} + \textbf{\textsc{FDS}}   & 7.83 & 7.23 & 12.60  & 22.37  & 4.42 & 4.20 & \textbf{6.93} & 13.48  \\[1.2pt]
\textsc{SQInv} + \textbf{\textsc{LDS}} + \textbf{\textsc{FDS}}   & \textbf{7.78} & \textbf{7.20} & {12.61} & \textbf{22.19} & \textbf{4.37}    & \textbf{4.12} & 7.39  & \textbf{12.61}  \\ \midrule\midrule
\textsc{\textbf{Ours~(best)} vs. Vanilla}   & \textcolor{darkgreen}{\textbf{+0.41}} & \textcolor{darkgreen}{\textbf{+0.21}} & \textcolor{darkgreen}{\textbf{+2.71}} & \textcolor{darkgreen}{\textbf{+4.14}} & \textcolor{darkgreen}{\textbf{+0.26}} & \textcolor{darkgreen}{\textbf{+0.15}} & \textcolor{darkgreen}{\textbf{+3.66}} & \textcolor{darkgreen}{\textbf{+7.85}} \\
\bottomrule[1.5pt]
\end{tabular}
}
\end{center}
\vspace{-0.5cm}
\end{table}

\begin{table}[t]
\setlength{\tabcolsep}{2.5pt}
\caption{Benchmarking results on AgeDB-DIR.}
\vspace{-4pt}
\label{table:agedb}
\small
\begin{center}
\resizebox{0.49\textwidth}{!}{
\begin{tabular}{l|cccc|cccc}
\toprule[1.5pt]
Metrics      & \multicolumn{4}{c|}{MAE~$\downarrow$}     & \multicolumn{4}{c}{GM~$\downarrow$}  \\ \midrule
Shot         & All  & Many & Med. & Few   & All  & Many & Med. & Few   \\ \midrule\midrule
\textsc{Vanilla}      & 7.77 & 6.62 & 9.55   & 13.67 & 5.05 & 4.23 & 7.01   & 10.75 \\ \midrule\midrule
\textsc{SmoteR}~\cite{torgo2013smoter}       & 8.16  & 7.39   & 8.65 & 12.28 & 5.21    & 4.65    & 5.69 & 8.49    \\[1.2pt]
\textsc{SMOGN}~\cite{branco2017smogn}        & 8.26  & 7.64   & 9.01 & 12.09 & 5.36    & 4.90    & 6.19 & 8.44    \\[1.2pt]
\textsc{SMOGN} + \textbf{\textsc{LDS}}   & 7.96 & 7.44 & 8.64  & 11.77 & 5.03 & 4.68 & 5.69  & 7.98  \\[1.2pt]
\textsc{SMOGN} + \textbf{\textsc{FDS}}   & 8.06 & 7.52 & 8.75  & 11.89 & 5.02 & 4.66 & 5.63  & 8.02  \\[1.2pt]
\textsc{SMOGN} + \textbf{\textsc{LDS}} + \textbf{\textsc{FDS}}   & \textbf{7.90} & \textbf{7.32}  & \textbf{8.51}  & \textbf{11.19}  & \textbf{4.98}  & \textbf{4.64}  & \textbf{5.41} & \textbf{7.35} \\ \midrule\midrule
\textsc{Focal-R}      & 7.64 & 6.68 & 9.22   & 13.00 & 4.90 & 4.26 & 6.39   & 9.52  \\[1.2pt]
\textsc{Focal-R} + \textbf{\textsc{LDS}}      & 7.56 & \textbf{6.67} & 8.82   & 12.40 & 4.82 & 4.27 & 5.87   & 8.83  \\[1.2pt]
\textsc{Focal-R} + \textbf{\textsc{FDS}}      & 7.65 & 6.89 & 8.70   & \textbf{11.92} & 4.83 & 4.32 & 5.89   & \textbf{8.04}  \\[1.2pt]
\textsc{Focal-R} + \textbf{\textsc{LDS}} + \textbf{\textsc{FDS}} & \textbf{7.47} & 6.69 & \textbf{8.30}  & 12.55 & \textbf{4.71} & \textbf{4.25} & \textbf{5.36}  & 8.59  \\ \midrule\midrule
\textsc{RRT}          & 7.74 & 6.98 & 8.79   & 11.99 & 5.00 & 4.50 & 5.88   & 8.63  \\[1.2pt]
\textsc{RRT} + \textbf{\textsc{LDS}}     & {7.72} & 7.00 & {8.75}   & {11.62} & {4.98} & 4.54 & {5.71}   & {8.27}  \\[1.2pt]
\textsc{RRT} + \textbf{\textsc{FDS}}     & {7.70} & \textbf{6.95} & {8.76}   & {11.86} & {4.82} & \textbf{4.32} & {5.83}   & {8.08}  \\[1.2pt]
\textsc{RRT} + \textbf{\textsc{LDS}} + \textbf{\textsc{FDS}}     & \textbf{7.66} & 6.99 & \textbf{8.60}   & \textbf{11.32} & \textbf{4.80} & 4.42 & \textbf{5.53}   & \textbf{6.99}  \\ \midrule\midrule
\textsc{SQInv}      & 7.81 & 7.16 & 8.80   & 11.20 & 4.99 & 4.57 & 5.73   & 7.77  \\[1.2pt]
\textsc{SQInv} + \textbf{\textsc{LDS}} & 7.67 & \textbf{6.98} & 8.86   & 10.89 & 4.85 & {4.39} &  5.80  & {7.45}  \\[1.2pt]
\textsc{SQInv} + \textbf{\textsc{FDS}} & 7.69 & 7.10 & 8.86   & \textbf{9.98} & 4.83 & {4.41} &  5.97  & \textbf{6.29}  \\[1.2pt]
\textsc{SQInv} + \textbf{\textsc{LDS}} + \textbf{\textsc{FDS}} & \textbf{7.55} &  7.01  & \textbf{8.24}   & 10.79 & \textbf{4.72} & \textbf{4.36} & \textbf{5.45}  & {6.79}  \\ \midrule\midrule
\textsc{\textbf{Ours~(best)} vs. Vanilla}   & \textcolor{darkgreen}{\textbf{+0.30}} & \textcolor{lightblue}{\textbf{-0.05}} & \textcolor{darkgreen}{\textbf{+1.31}} & \textcolor{darkgreen}{\textbf{+3.69}} & \textcolor{darkgreen}{\textbf{+0.34}} & \textcolor{lightblue}{\textbf{-0.02}} & \textcolor{darkgreen}{\textbf{+1.65}} & \textcolor{darkgreen}{\textbf{+4.46}} \\
\bottomrule[1.5pt]
\end{tabular}
}
\end{center}
\vspace{-0.5cm}
\end{table}

\begin{table}[!htbp]
\setlength{\tabcolsep}{2.5pt}
\caption{Benchmarking results on STS-B-DIR.}
\vspace{-4pt}
\label{table:sts-b}
\small
\begin{center}
\resizebox{0.49\textwidth}{!}{
\begin{tabular}{l|cccc|cccc}
\toprule[1.5pt]
Metrics      & \multicolumn{4}{c|}{MSE~$\downarrow$}       & \multicolumn{4}{c}{Pearson correlation~(\%)~$\uparrow$}    \\ \midrule
Shot         & All   & Many  & Med. & Few   & All   & Many  & Med. & Few   \\ \midrule\midrule
\textsc{Vanilla}      & 0.974 & 0.851 & 1.520  & 0.984 & 74.2 & 72.0 & 62.7  & 75.2 \\ \midrule\midrule
\textsc{SmoteR}~\cite{torgo2013smoter}       & 1.046     & 0.924     & 1.542      & 1.154     & 72.6     & 69.3     & 65.3      & 70.6     \\[1.2pt]
\textsc{SMOGN}~\cite{branco2017smogn}        & 0.990     & 0.896     & 1.327      & 1.175     & 73.2     & 70.4     & 65.5      & 69.2     \\[1.2pt]
\textsc{SMOGN} + \textbf{\textsc{LDS}}   & 0.962     & 0.880     & 1.242      & 1.155     & 74.0     & 71.5     & 65.2      & 69.8     \\[1.2pt]
\textsc{SMOGN} + \textbf{\textsc{FDS}}   & 0.987    & 0.945    & \textbf{1.101}      & 1.153     & 73.0    & 69.6    & \textbf{68.5}      & 69.9     \\[1.2pt]
\textsc{SMOGN} + \textbf{\textsc{LDS}} + \textbf{\textsc{FDS}}   &  \textbf{0.950}   & \textbf{0.851}    & 1.327      & \textbf{1.095}     & \textbf{74.6}    & \textbf{72.1}    & 65.9      & \textbf{71.7}     \\ \midrule\midrule
\textsc{Focal-R}      & 0.951 & 0.843 & 1.425  & 0.957 & 74.6 & 72.3 & 61.8  & 76.4 \\[1.2pt]
\textsc{Focal-R} + \textbf{\textsc{LDS}} & 0.930     & \textbf{0.807}     & 1.449      & 0.993     & \textbf{75.7}     & \textbf{73.9}     & 62.4      & 75.4 \\[1.2pt]
\textsc{Focal-R} + \textbf{\textsc{FDS}} & \textbf{0.920} & 0.855 & \textbf{1.169}  & 1.008     & 75.1     & 72.6   & \textbf{66.4}      & 74.7 \\[1.2pt]
\textsc{Focal-R} + \textbf{\textsc{LDS}} + \textbf{\textsc{FDS}} & 0.940 & 0.849 & 1.358  & \textbf{0.916}     & 74.9     & 72.2     & 66.3      & \textbf{77.3} \\ \midrule\midrule
\textsc{RRT}         & 0.964 & 0.842 & 1.503  & 0.978 & 74.5 & 72.4 & 62.3  & 75.4 \\[1.2pt]
\textsc{RRT} + \textbf{\textsc{LDS}}     & {0.916} & 0.817 & {1.344}  & 0.945 & {75.7} & {73.5} & {64.1}  & {76.6} \\[1.2pt]
\textsc{RRT} + \textbf{\textsc{FDS}} & 0.929 & 0.857 & \textbf{1.209}  & 1.025     & 74.9     & 72.1     & \textbf{67.2}      & 74.0 \\[1.2pt]
\textsc{RRT} + \textbf{\textsc{LDS}} + \textbf{\textsc{FDS}} & \textbf{0.903} & \textbf{0.806} & 1.323  & \textbf{0.936}     & \textbf{76.0}     &\textbf{73.8}     & 65.2      & \textbf{76.7} \\ \midrule\midrule
\textsc{Inv}      & 1.005 & 0.894 & 1.482  & 1.046 & 72.8 & 70.3 & 62.5  & 73.2 \\[1.2pt]
\textsc{Inv} + \textbf{\textsc{LDS}} & 0.914 & 0.819 & {1.319}  & {0.955} & {75.6} & {73.4} & {63.8}  & {76.2} \\[1.2pt]
\textsc{Inv} + \textbf{\textsc{FDS}} & 0.927 & 0.851 & \textbf{1.225}  & 1.012     & 75.0     & 72.4     & \textbf{66.6}      & 74.2 \\[1.2pt]
\textsc{Inv} + \textbf{\textsc{LDS}} + \textbf{\textsc{FDS}} & \textbf{0.907} & \textbf{0.802} & 1.363  & \textbf{0.942}     & \textbf{76.0}     & \textbf{74.0}     & 65.2      & \textbf{76.6} \\ \midrule\midrule
\textsc{\textbf{Ours~(best)} vs. Vanilla}   & \textcolor{darkgreen}{\textbf{+.071}} & \textcolor{darkgreen}{\textbf{+.049}} & \textcolor{darkgreen}{\textbf{+.419}} & \textcolor{darkgreen}{\textbf{+.068}} & \textcolor{darkgreen}{\textbf{+1.8}} & \textcolor{darkgreen}{\textbf{+2.0}} & \textcolor{darkgreen}{\textbf{+5.8}} & \textcolor{darkgreen}{\textbf{+2.1}} \\
\bottomrule[1.5pt]
\end{tabular}
}
\end{center}
\vspace{-0.5cm}
\end{table}

\textbf{Inferring Text Similarity Score: STS-B-DIR.}
Table~\ref{table:sts-b} shows the results, where similar observations can be made on STS-B-DIR. Again, both SMOTER and SMOGN perform worse than the vanilla model. In contrast, both LDS and FDS consistently and substantially improve the results for various methods, especially in medium- and few-shot regions. The advantage is even more profound under \emph{Pearson correlation}, which is commonly used for this NLP task.

\textbf{Inferring Depth: NYUD2-DIR.}
For NYUD2-DIR, which is a dense regression task, we verify from Table~\ref{table:nyu2} that adding LDS and FDS significantly improves the results. We note that the vanilla model can inevitably overfit to the many-shot regions during training. FDS and LDS help alleviate this effect, and generalize better to all regions, with minor degradation in the many-shot region but significant boosts for other regions.

\begin{table}[!t]
\setlength{\tabcolsep}{2.5pt}
\caption{Benchmarking results on NYUD2-DIR.}
\vspace{-4pt}
\label{table:nyu2}
\small
\begin{center}
\resizebox{0.49\textwidth}{!}{
\begin{tabular}{l|cccc|cccc}
\toprule[1.5pt]
Metrics      & \multicolumn{4}{c|}{RMSE~$\downarrow$}       & \multicolumn{4}{c}{$\delta_1$~$\uparrow$}    \\ \midrule
Shot         & All   & Many  & Med. & Few   & All   &Many   & Med. & Few   \\ \midrule\midrule
\textsc{Vanilla}      & 1.477 & 0.591 & 0.952  & 2.123 & 0.677 & 0.777 &  0.693  & 0.570 \\ \midrule\midrule
\textsc{Vanilla} + \textbf{\textsc{LDS}} &  1.387 &  0.671  &    0.913 & 1.954  & 0.672  & 0.701  & 0.706 & 0.630 \\[1.2pt]
\textsc{Vanilla} + \textbf{\textsc{FDS}} & 1.442 & \textbf{0.615} & 0.940  & 2.059  & 0.681  & \textbf{0.760} & 0.695 & 0.596 \\[1.2pt]
\textsc{Vanilla} + \textbf{\textsc{LDS}} + \textbf{\textsc{FDS}} & \textbf{1.338} & 0.670  & \textbf{0.851} & \textbf{1.880}  & \textbf{0.705} & 0.730  & \textbf{0.764}  & \textbf{0.655}  \\ \midrule\midrule
\textsc{\textbf{Ours~(best)} vs. Vanilla}   & \textcolor{darkgreen}{\textbf{+.139}} & \textcolor{lightblue}{\textbf{-.024}} & \textcolor{darkgreen}{\textbf{+.101}} & \textcolor{darkgreen}{\textbf{+.243}} & \textcolor{darkgreen}{\textbf{+.028}} & \textcolor{lightblue}{\textbf{-.017}} & \textcolor{darkgreen}{\textbf{+.071}} & \textcolor{darkgreen}{\textbf{+.085}} \\
\bottomrule[1.5pt]
\end{tabular}
}
\end{center}
\vspace{-0.5cm}
\end{table}

\begin{table}[tb]
\setlength{\tabcolsep}{2.5pt}
\caption{Benchmarking results on SHHS-DIR.}
\vspace{-4pt}
\label{table:shhs}
\small
\begin{center}
\resizebox{0.49\textwidth}{!}{
\begin{tabular}{l|cccc|cccc}
\toprule[1.5pt]
Metrics      & \multicolumn{4}{c|}{MAE~$\downarrow$}       & \multicolumn{4}{c}{GM~$\downarrow$}    \\ \midrule
Shot         & All   & Many  & Med. & Few   & All   & Many  & Med. & Few   \\ \midrule\midrule
\textsc{Vanilla}      & 15.36 & 12.47 & 13.98  & 16.94 & 10.63 & 8.04 &  9.59  & 12.20 \\ \midrule\midrule
\textsc{Focal-R}      & 14.67 & 11.70  & 13.69  & 17.06 & 9.98 & 7.93 & 8.85 & 11.95 \\[1.2pt]
\textsc{Focal-R} + \textbf{\textsc{LDS}} & 14.49  & 12.01  & 12.43 & 16.57 & 9.98 & 7.89 & 8.59 & 11.40  \\[1.2pt]
\textsc{Focal-R} + \textbf{\textsc{FDS}} & 14.18 &  \textbf{11.06}  & 13.56 & 15.99 & 9.45 & \textbf{6.95} & 8.81 & 11.13 \\[1.2pt]
\textsc{Focal-R} + \textbf{\textsc{LDS}} + \textbf{\textsc{FDS}} & \textbf{14.02} & 11.08 &  \textbf{12.24} & \textbf{15.49}  & \textbf{9.32} & 7.18  & \textbf{8.10}  & \textbf{10.39}  \\ \midrule\midrule
\textsc{RRT}         & 14.78 & 12.43 & 14.01 & 16.48 & 10.12 & 8.05 & 9.71 & 11.96 \\[1.2pt]
\textsc{RRT} + \textbf{\textsc{LDS}}     & {14.56} & 12.08 & {13.44}  & 16.45 & {9.89} & {7.85} & {9.18}  & {11.82} \\[1.2pt]
\textsc{RRT} + \textbf{\textsc{FDS}} & 14.36 & 11.97 & {13.33}  & 16.08 & 9.74 & 7.54 & 9.20  & 11.31 \\[1.2pt]
\textsc{RRT} + \textbf{\textsc{LDS}} + \textbf{\textsc{FDS}} & \textbf{14.33} & \textbf{11.96} & \textbf{12.47}  & \textbf{15.92}     & \textbf{9.63}     & \textbf{7.35}     & \textbf{8.74} & \textbf{11.17} \\ \midrule\midrule
\textsc{Inv}      & 14.39 & 11.84 & 13.12  & 16.02 & 9.34 & 7.73 & 8.49  & 11.20 \\[1.2pt]
\textsc{Inv} + \textbf{\textsc{LDS}} & 14.14 & 11.66 & {12.77}  & {16.05} & {9.26} & {7.64} & {8.18}  & {11.32} \\[1.2pt]
\textsc{Inv} + \textbf{\textsc{FDS}} & 13.91 & \textbf{11.12} & {12.29}  & 15.53 & 8.94  & \textbf{6.91}  & {7.79}      & 10.65 \\[1.2pt]
\textsc{Inv} + \textbf{\textsc{LDS}} + \textbf{\textsc{FDS}} & \textbf{13.76} & \textbf{11.12} & \textbf{12.18}  & \textbf{15.07}     & \textbf{8.70}     & 6.94     & \textbf{7.60}      & \textbf{10.18} \\ \midrule\midrule
\textsc{\textbf{Ours~(best)} vs. Vanilla}   & \textcolor{darkgreen}{\textbf{+1.60}} & \textcolor{darkgreen}{\textbf{+1.41}} & \textcolor{darkgreen}{\textbf{+1.80}} & \textcolor{darkgreen}{\textbf{+1.87}} & \textcolor{darkgreen}{\textbf{+1.93}} & \textcolor{darkgreen}{\textbf{+1.13}} & \textcolor{darkgreen}{\textbf{+1.99}} & \textcolor{darkgreen}{\textbf{+2.02}} \\
\bottomrule[1.5pt]
\end{tabular}
}
\end{center}
\vspace{-0.1cm}
\end{table}

\begin{figure}[!tb]
\begin{center}
 \includegraphics[width=0.99\linewidth]{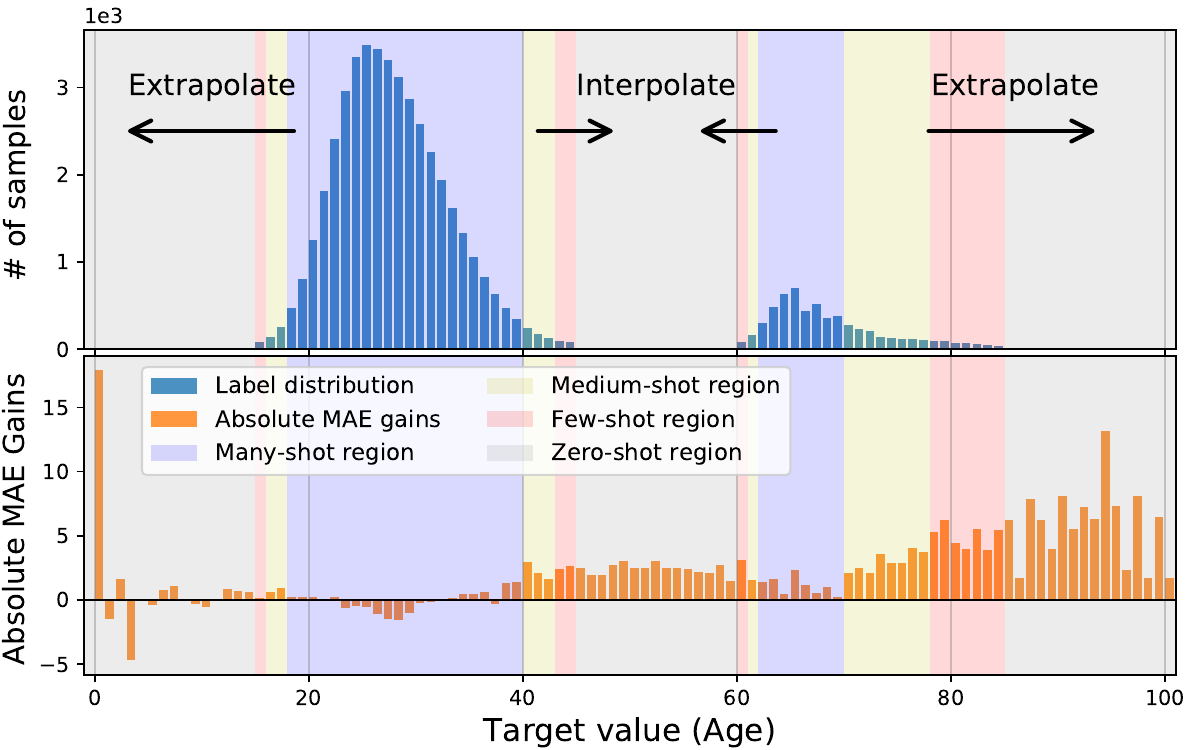}
\end{center}
\vspace{-0.5cm}
\caption{The absolute MAE gains of LDS $+$ FDS over the vanilla model, on a curated subset of IMDB-WIKI-DIR with certain target values having no training data. We establish notable performance gains w.r.t. all regions, especially for extrapolation \& interpolation.}
\label{fig:interp-extrap}
\vspace{-0.3cm}
\end{figure}

\textbf{Inferring Health Score: SHHS-DIR.}
Table~\ref{table:shhs} reports the results on SHHS-DIR. Since SMOTER and SMOGN are not directly applicable to this medical data, we skip them for this dataset. The results again confirm the effectiveness of both FDS and LDS when applied for real-world imbalanced regression tasks, where by combining FDS and LDS we often get the highest gains over all tested regions.

\begin{figure*}[t]
\centering
\subfigure[Feature statistics similarity for age $0$, without FDS]{
    \label{fig:feat_sim_vanilla}
    \includegraphics[height=0.265\textwidth]{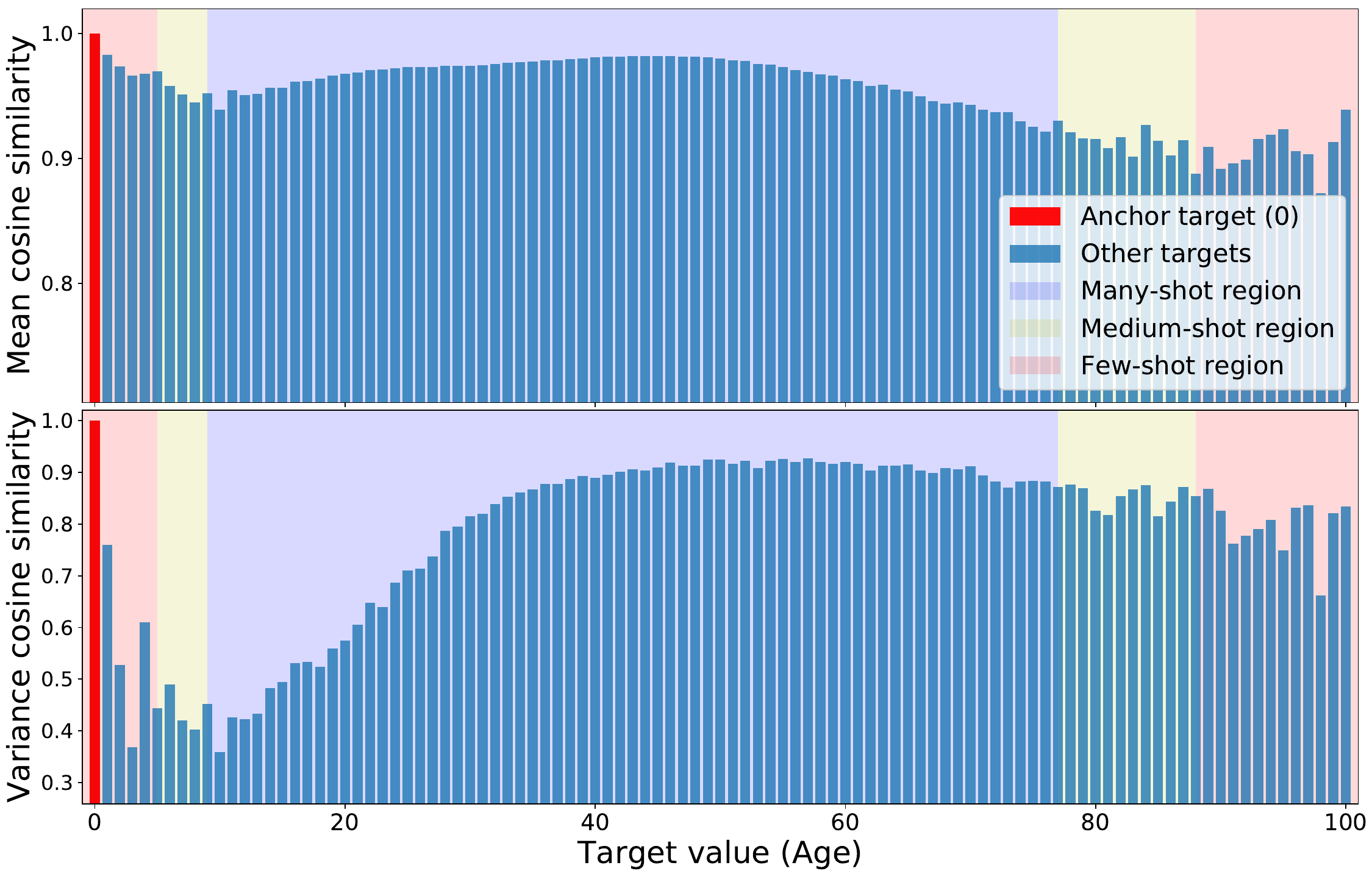}
}
\hspace{-2.2ex}
\subfigure[Feature statistics similarity for age $0$, with FDS]{
    \label{fig:feat_sim_ours}
    \includegraphics[height=0.265\textwidth]{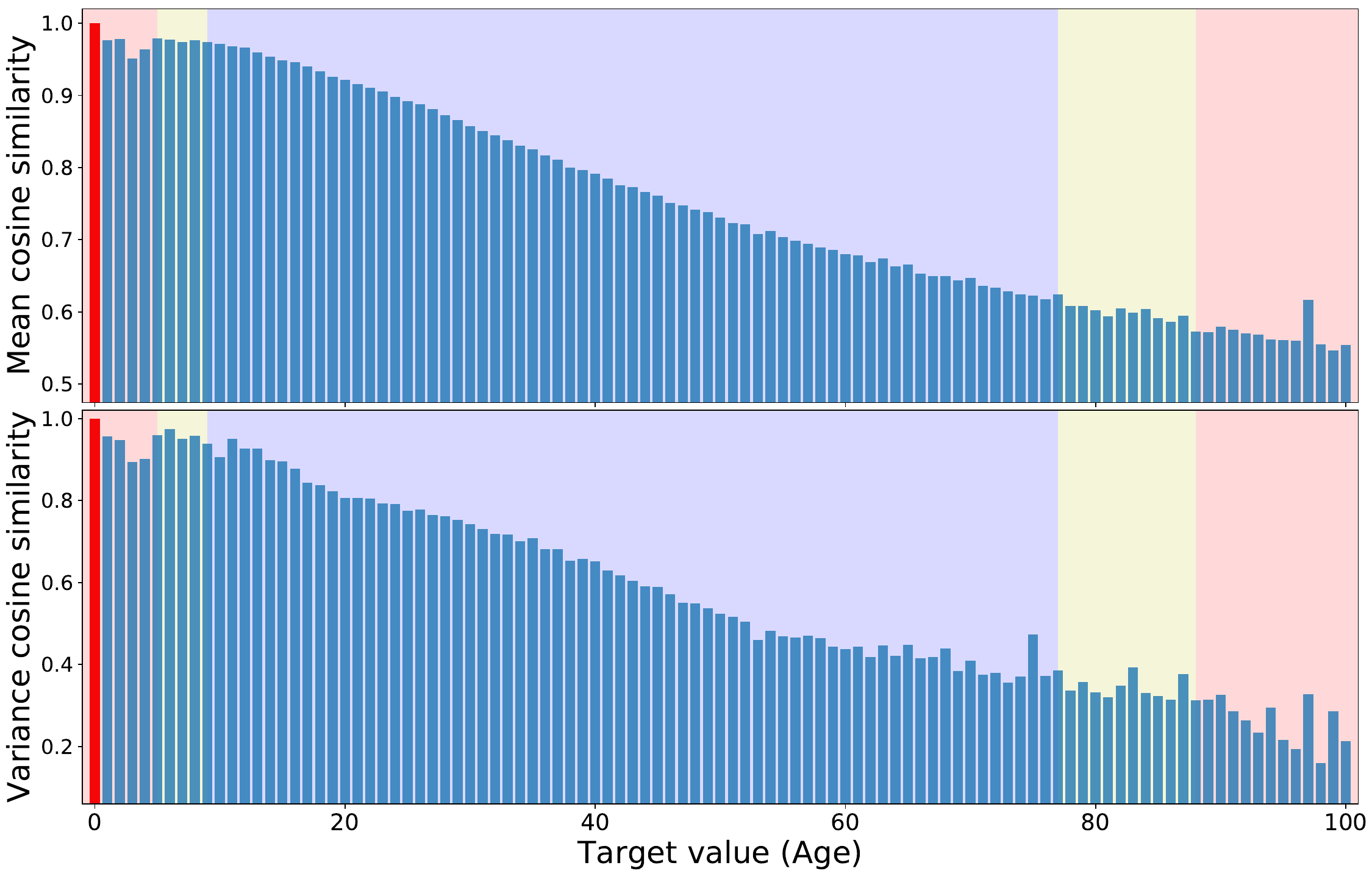}
}
\hspace{-1.8ex}
\subfigure[Statistics change]{
    \label{fig:train_stats_difference}
    \includegraphics[height=0.265\textwidth]{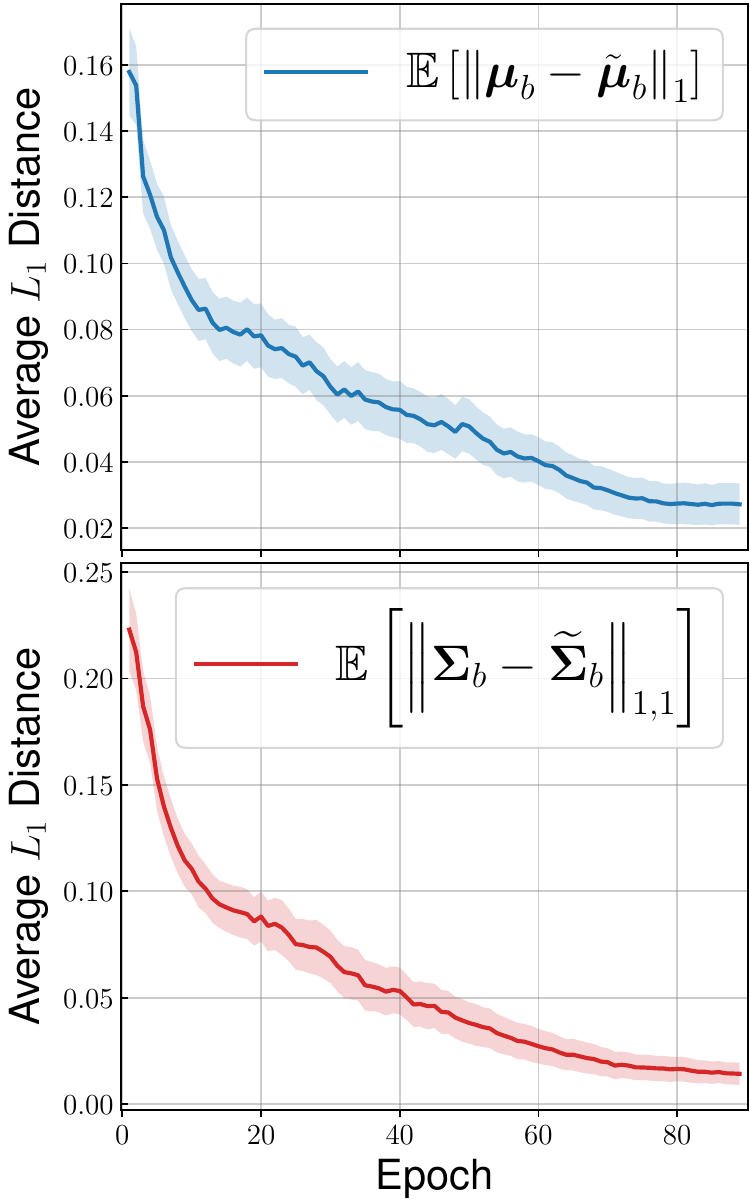}
}
\vspace{-0.55cm}
\caption{Analysis on how FDS works. \textbf{(a) \& (b)} Feature statistics similarity for anchor age $0$, using model trained without and with FDS. \textbf{(c)} $L_1$ distance between the running statistics $\{{\boldsymbol{\mu}}_b, \boldsymbol{{\Sigma}}_b\}$ and the smoothed statistics $\{\tilde{\boldsymbol{\mu}}_b, \boldsymbol{\widetilde{\Sigma}}_b\}$ during training.
}
\label{fig:fds-analysis}
\vspace{-0.3cm}
\end{figure*}

\begin{table}[t]
\setlength{\tabcolsep}{1.5pt}
\caption{Interpolation \& extrapolation results on the curated subset of IMDB-WIKI-DIR. Using LDS and FDS, the generalization results on zero-shot regions can be consistently improved.}
\vspace{-8pt}
\label{table:interp-extrap}
\small
\begin{center}
\resizebox{0.49\textwidth}{!}{
\begin{tabular}{l|cccc|cccc}
\toprule[1.5pt]
Metrics      & \multicolumn{4}{c|}{MAE~$\downarrow$}       & \multicolumn{4}{c}{GM~$\downarrow$}    \\ \midrule
Shot         & All   & w/ data  & Interp. & Extrap.   & All   & w/ data  & Interp. & Extrap.   \\ \midrule\midrule
\textsc{Vanilla}      & 11.72 & 9.32 & 16.13  & 18.19 & 7.44 & 5.33 &  14.41  & 16.74 \\ \midrule\midrule
\textsc{Vanilla} + \textbf{\textsc{LDS}} &  10.54 &  8.31  &    14.14 & 17.38  & 6.50  & 4.67  & 12.13 & 15.36 \\[1.2pt]
\textsc{Vanilla} + \textbf{\textsc{FDS}} & 11.40 & 8.97 & 15.83  & 18.01  & 7.18  & 5.12 & 14.02 & 16.48 \\[1.2pt]
\textsc{Vanilla} + \textbf{\textsc{LDS}} + \textbf{\textsc{FDS}} & \textbf{10.27} & \textbf{8.11}  & \textbf{13.71} & \textbf{17.02}  & \textbf{6.33} & \textbf{4.55}  & \textbf{11.71}  & \textbf{15.13}  \\ \midrule\midrule
\textsc{\textbf{Ours~(best)} vs. Vanilla}   & \textcolor{darkgreen}{\textbf{+1.45}} & \textcolor{darkgreen}{\textbf{+1.21}} & \textcolor{darkgreen}{\textbf{+2.42}} & \textcolor{darkgreen}{\textbf{+1.17}} & \textcolor{darkgreen}{\textbf{+1.11}} & \textcolor{darkgreen}{\textbf{+0.78}} & \textcolor{darkgreen}{\textbf{+2.70}} & \textcolor{darkgreen}{\textbf{+1.61}} \\
\bottomrule[1.5pt]
\end{tabular}
}
\end{center}
\vspace{-0.5cm}
\end{table}

\subsection{Further Analysis}
\label{sec:exp-further-analysis}

\textbf{Extrapolation \& Interpolation.}
In real-world DIR tasks, certain target values can have no data at all (e.g., see SHHS-DIR and STS-B-DIR in Fig.~\ref{fig:dataset-info}). This motivates the need for target extrapolation and interpolation. We curate a subset from the training set of IMDB-WIKI-DIR, which has no training data in certain regions (Fig.~\ref{fig:interp-extrap}), but evaluate on the original testset for zero-shot generalization analysis.

As Table~\ref{table:interp-extrap} shows, compared to the vanilla model, LDS and FDS can both improve the results not only on regions that have data, but also achieve larger gains on those without data. Specifically, substantial improvements are established for both target interpolation and extrapolation, where interpolation enjoys larger boosts.

We further visualize the absolute MAE gains of our method over vanilla model in Fig.~\ref{fig:interp-extrap}. Our method provides a comprehensive treatment to the many, medium, few, as well as zero-shot regions, achieving remarkable performance gains.

\textbf{Understanding FDS.}
We investigate how FDS influences the feature statistics. In Fig.~\ref{fig:feat_sim_vanilla} and \ref{fig:feat_sim_ours} we plot the similarity of the feature statistics for anchor age $0$, using model trained without and with FDS. As the figure indicates, since age $0$ lies in the few-shot region, the feature statistics can have a large bias, i.e., age $0$ shares large similarity with region $40\sim 80$ as in Fig.~\ref{fig:feat_sim_vanilla}. In contrast, when FDS is added, the statistics are better calibrated, resulting in a high similarity only in its neighborhood, and a gradually decreasing similarity score as target value becomes larger.
We further visualize the $L_1$ distance between the running statistics $\{{\boldsymbol{\mu}}_b, \boldsymbol{{\Sigma}}_b\}$ and the smoothed statistics $\{\tilde{\boldsymbol{\mu}}_b, \boldsymbol{\widetilde{\Sigma}}_b\}$ during training in Fig.~\ref{fig:train_stats_difference}. Interestingly, the average $L_1$ distance becomes smaller and gradually diminishes as the training evolves, indicating that the model learns to generate features that are more accurate even without smoothing, and finally the smoothing module can be removed during inference.
We provide more results for different anchor ages in Appendix~\ref{appendix:visualize-feat-sim-all-ages}, where similar effects can be observed.

\textbf{Ablation: Kernel type for LDS \& FDS (Appendix~\ref{appendix:ablation-kernel-type}).}
We study the effects of different kernel types for LDS and FDS when applying distribution smoothing.
We select three different kernel types, i.e., Gaussian, Laplacian, and Triangular kernel, and evaluate their influences on both LDS and FDS. In general, all kernel types lead to notable gains~(e.g., $3.7\%\sim 6.2 \%$ relative MSE gains on STS-B-DIR), with the Gaussian kernel often delivering the best results.

\textbf{Ablation: Different regression loss functions (Appendix \ref{appendix:ablation-diff-training-loss}).}
We investigate the influence of different training loss functions on LDS and FDS. We select three common losses used for regression tasks, i.e., $L_1$ loss, MSE loss, and the Huber loss~(also referred to as smoothed $L_1$ loss). We find that similar results are obtained for all losses, indicating that both LDS and FDS are robust to different loss functions.

\textbf{Ablation: Hyper-parameter for LDS \& FDS (Appendix \ref{appendix:ablation-hyper-param}).}
We investigate the effects of hyper-parameters on both LDS and FDS. As we mainly employ the Gaussian kernel for distribution smoothing, we extensively study different choices of the kernel size $l$ and standard deviation $\sigma$. Interestingly, we find LDS and FDS are surprisingly robust to different hyper-parameters in a given range, and obtain similar gains. For example, on STS-B-DIR with $l\in\{\texttt{5,9,15}\}$ and $\sigma\in\{\texttt{1,2,3}\}$, overall MSE gains range from $3.3\%$ to $6.2 \%$, with $l=\texttt{5}$ and $\sigma=\texttt{2}$ exhibiting the best results.

\textbf{Ablation: Robustness to diverse skewed label densities (Appendix \ref{appendix:ablation-diff-skewed-label-density}).}
We curate different imbalanced distributions for IMDB-WIKI-DIR by combining different number of disjoint skewed Gaussian distributions over the target space, with potential missing data in certain target regions, and evaluate the robustness of FDS and LDS to the distribution change. We verify that even under different imbalanced label distributions, LDS and FDS consistently boost the performance across all regions compared to the vanilla model, with relative MAE gains ranging from $8.8\%$ to $12.4\%$.

\textbf{Comparisons to imbalanced classification methods (Appendix \ref{appendix:compare-imb-classification}).}
Finally, to gain more insights on the intrinsic difference between imbalanced classification \& imbalanced regression problems, we directly apply existing imbalanced classification schemes on several appropriate DIR datasets, and show empirical comparisons with imbalanced regression approaches. We demonstrate in Appendix \ref{appendix:compare-imb-classification} that LDS and FDS outperform imbalanced classification schemes by a large margin, where the errors for few-shot regions can be reduced by up to $50\%$ to $60\%$. Interestingly, the results also show that imbalanced classification schemes often perform \emph{worse} than even the vanilla regression model, which confirms that regression requires different approaches for data imbalance than simply applying classification methods.
We note that imbalanced classification methods could fail on regression problems for several reasons. First, they ignore the similarity between data samples that are close w.r.t. the continuous target. Moreover, classification cannot extrapolate or interpolate in the continuous label space, therefore unable to deal with missing data in certain target regions.

\section{Conclusion}
\label{sec:conclusion}
We introduce the DIR task that learns from natural imbalanced data with continuous targets, and generalizes to the entire target range. We propose two simple and effective algorithms for DIR that exploit the similarity between nearby targets in both label and feature spaces.
Extensive results on five curated large-scale real-world DIR benchmarks confirm the superior performance of our methods.
Our work fills the gap in benchmarks and techniques for practical DIR tasks.


\bibliography{imbalance}
\bibliographystyle{icml2021}

\onecolumn
\newpage
\appendix
\begin{center}
\LARGE{\textbf{Supplementary Material}}
\end{center}

\section{Pseudo Code for LDS \& FDS}
\label{appendix:pseudo-code}

We provide the pseudo code of the proposed LDS and FDS algorithms in Algorithm~\ref{alg:lds} and \ref{alg:fds}, respectively. For LDS, we provide an illustrative example which combines LDS with the loss inverse re-weighting scheme.

\begin{algorithm}[!th]
   \caption{Label Distribution Smoothing (LDS)}
   \label{alg:lds}
\begin{algorithmic}
   \STATE {\bfseries Input:} Training set $\mathcal{D}=\{ ( \mathbf{x}_i, y_i )\}_{i=1}^{N}$, bin size $\Delta b$, symmetric kernel distribution $k(y, y')$
   \STATE Calculate the empirical label density distribution $p(y)$ based on $\Delta b$ and $\mathcal{D}$
   \STATE Calculate the effective label density distribution $\tilde{p}(y') \triangleq \int_{\mathcal{Y}}\mathrm{k}(y, y') p(y)dy$
   \STATE
   \STATE \texttt{/*}~~\texttt{Example:}~~\texttt{Combine}~~\texttt{LDS}~~\texttt{with}~~\texttt{loss}~~\texttt{inverse}~~\texttt{re-weighting}~~\texttt{*/}
   \FORALL{$( \mathbf{x}_i, y_i )\in \mathcal{D}$}
   \STATE Compute weight for each sample as $w_i = \frac{c}{\tilde{p}(y_i)} \propto \frac{1}{\tilde{p}(y_i)}$ (constant $c$ as scaling factor)
   \ENDFOR
   \FOR{number of training iterations}
   \STATE Sample a mini-batch $\{ (\mathbf{x}_i, y_i, w_i) \}_{i=1}^m$ from $\mathcal{D}$
   \STATE Forward $\{ \mathbf{x}_i\}_{i=1}^m$ and get corresponding predictions $\{ \hat{y}_i\}_{i=1}^m$
   \STATE Do one training step using the weighted loss $\frac{1}{m} \sum_{i=1}^m w_i \mathcal{L}(\hat{y}_i, y_i)$
   \ENDFOR
\end{algorithmic}
\end{algorithm}

\begin{algorithm}[!th]
   \caption{Feature Distribution Smoothing (FDS)}
   \label{alg:fds}
\begin{algorithmic}
   \STATE {\bfseries Input:} Training set $\mathcal{D}=\{ ( \mathbf{x}_i, y_i )\}_{i=1}^{N}$, bin index space $\mathcal{B}$, symmetric kernel distribution $k(y, y')$, encoder $f$, regression function $g$, total training epochs $E$, FDS momentum $\alpha$
   \FORALL{$b\in \mathcal{B}$}
   \STATE Initialize the running statistics $\{{\boldsymbol{\mu}}_b^{(0)}, \boldsymbol{{\Sigma}}_b^{(0)}\}$ and the smoothed statistics $\{\tilde{\boldsymbol{\mu}}_b^{(0)}, \boldsymbol{\widetilde{\Sigma}}_b^{(0)}\}$
   \ENDFOR
   \FOR{$e = 0$ \textbf{to} $E$}
   \REPEAT
   \STATE Sample a mini-batch $\{ (\mathbf{x}_i, y_i) \}_{i=1}^m$ from $\mathcal{D}$
   \FOR{$i=1$ \textbf{to} $m$ (in parallel)}
   \STATE $\mathbf{z}_i = f(\mathbf{x}_i)$
   \STATE $\tilde{\mathbf{z}}_i = \left(\boldsymbol{\widetilde{\Sigma}}_b^{(e)}\right)^{\frac{1}{2}} \left(\boldsymbol{\Sigma}_b^{(e)}\right)^{-\frac{1}{2}} (\mathbf{z}_i - \boldsymbol{\mu}_b^{(e)}) + \tilde{\boldsymbol{\mu}}_b^{(e)}$~~~~~~~~\texttt{/*}~~\texttt{Feature}~~\texttt{statistics}~~\texttt{calibration}~~\texttt{*/}
   \STATE $\hat{y}_i = g(\tilde{\mathbf{z}}_i)$
   \ENDFOR
   \STATE Do one training step with loss $\frac{1}{m} \sum_{i=1}^m \mathcal{L}(\hat{y}_i, y_i)$
   \UNTIL{iterate over all training samples at current epoch $e$}
    \STATE \texttt{/*}~~\texttt{Update}~~\texttt{running}~~\texttt{statistics}~~\texttt{\&}~~\texttt{smoothed}~~\texttt{statistics}~~\texttt{*/}
   \FORALL{$b\in \mathcal{B}$}
   \STATE Estimate current running statistics of $b$-th bin $\{{\boldsymbol{\mu}}_b, \boldsymbol{{\Sigma}}_b\}$ using Eqn.~(\ref{eq:estimate-run-stats-mu}) and (\ref{eq:estimate-run-stats-sigma})
   \STATE $\boldsymbol{\mu}_b^{(e+1)} \leftarrow \alpha\times \boldsymbol{\mu}_b^{(e)} + (1-\alpha)\times \boldsymbol{\mu}_b$
   \STATE $\boldsymbol{\Sigma}_b^{(e+1)} \leftarrow \alpha\times \boldsymbol{\Sigma}_b^{(e)} + (1-\alpha)\times \boldsymbol{\Sigma}_b$
   \ENDFOR
   \STATE Update smoothed statistics $\{\tilde{\boldsymbol{\mu}}_b^{(e+1)}, \boldsymbol{\widetilde{\Sigma}}_b^{(e+1)}\}_{b\in \mathcal{B}}$ based on $\{\boldsymbol{\mu}_b^{(e+1)}, \boldsymbol{\Sigma}_b^{(e+1)}\}_{b\in \mathcal{B}}$ using Eqn.~(\ref{eq:smooth-stats-mu}) and (\ref{eq:smooth-stats-sigma})
   \ENDFOR
\end{algorithmic}
\end{algorithm}

\section{Details of DIR Datasets}
\label{appendix:dataset-details}

In this section, we provide the detailed information of the five curated DIR datasets we used in our experiments. Table~\ref{table:dir-dataset-overview} provides an overview of the five datasets.

\begin{table}[!t]
\setlength{\tabcolsep}{4pt}
\caption{Overview of the five curated DIR datasets used in our experiments.}
\vspace{-3pt}
\label{table:dir-dataset-overview}
\small
\begin{center}
\resizebox{\textwidth}{!}{
\begin{tabular}{c|c|c|c|c|c|c|c|c}
\toprule[1.5pt]
Dataset       & \multicolumn{1}{c|}{Target type} & \multicolumn{1}{c|}{Target range} & \multicolumn{1}{c|}{Bin size} & \multicolumn{1}{c|}{Max bin density} & \multicolumn{1}{c|}{Min bin density} & \multicolumn{1}{c|}{\# Training set} & \multicolumn{1}{c|}{\# Val. set} & \multicolumn{1}{c}{\# Test set} \\ \midrule\midrule
IMDB-WIKI-DIR & Age & 0 $\sim$ 186 & 1 & 7,149  & 1 & 191,509 & 11,022 & 11,022 \\ \midrule
AgeDB-DIR     & Age & 0 $\sim$ 101 & 1 & 353    & 1 & 12,208  & 2,140  & 2,140  \\ \midrule
STS-B-DIR     & Text similarity score & 0 $\sim$ 5  & 0.1 &  428  & 1  & 5,249  &  1,000  &     1,000  \\ \midrule
NYUD2-DIR     & Depth & 0.7 $\sim$ 10 & 0.1  &  $1.46 \times 10^8$ & $1.13 \times 10^6$ & 50,688 ($3.51 \times 10^9$)  & $-$ & 654 ($8.70\times 10^5$)    \\ \midrule
SHHS-DIR      & Health condition score & 0 $\sim$ 100 & 1  & 275 & 0 & 1,892 & 369 & 369 \\
\bottomrule[1.5pt]
\end{tabular}}
\end{center}
\vspace{-0.45cm}
\end{table}

\subsection{IMDB-WIKI-DIR}
The original IMDB-WIKI dataset~\cite{imdb-wiki} is a large-scale face image dataset for age estimation from single input image. The original version contains 523.0K face images and the corresponding ages, where 460.7K face images are collected from the IMDB website and 62.3K images from the Wikipedia website.
We construct IMDB-WIKI-DIR by first filtering out unqualified images with low face scores~\cite{imdb-wiki}, and then manually creating balanced validation and test set over the supported ages. Overall, the curated dataset has 191.5K images for training, and 11.0K images for validation and testing, respectively. We make the length of each bin to be 1 year, with a minimum age of 0 and a maximum age of 186. The number of images per bin varies between 1 and 7,149, exhibiting significant data imbalance.

As for the data pre-processing, the images are first resized to $224\times 224$. During training, we follow the standard data augmentation scheme~\cite{he2016deep} to do zero-padding with 16 pixels on each side, and then random crop back to the original image size. We then randomly flip the images horizontally and normalize them into $[0, 1]$.

\subsection{AgeDB-DIR}
The original AgeDB dataset~\cite{moschoglou2017agedb} is a manually collected in-the-wild age database with accurate and noise-free labels. Similar to IMDB-WIKI, the task is also to estimate age from visual appearance.
The original dataset contains 16,488 images in total.
We construct AgeDB-DIR in a similar manner as IMDB-WIKI-DIR, where the training set contains 12,208 images, with a minimum age of 0 and a maximum age of 101, and maximum bin density of 353 images and minimum bin density of 1. The validation set and test set are made balanced with 2,140 images.
Similarly, the images in AgeDB are resized to $224\times 224$, and go through the same data pre-processing schedule as in the IMDB-WIKI-DIR dataset.

\subsection{STS-B-DIR}
The original Semantic Textual Similarity Benchmark (STS-B)~\cite{cer2017semeval}, also included in the GLUE benchmark~\cite{wang2018glue}, is a collection of sentence pairs drawn from news headlines, video and image captions, and natural language inference data. Each pair is human-annotated by multiple annotators with an averaged continuous similarity score from 0 to 5. The task is to predict these scores from the sentence pairs. From the original training set of 7.2K pairs, we create a training set with 5.2K pairs, and balanced validation set and test set of 1K pairs each for STS-B-DIR. We make the length of each bin to be 0.1, and the number of training pairs per bin varies between 1 and 428. 

As for the data pre-processing, the sentences are first tokenized using NLTK toolkit~\cite{loper2002nltk} with a maximum length of 40. We then count the frequencies of all words (tokens) of all splits, build the word vocabulary based on the word frequency, and finally use the 300D GloVe word embeddings (840B Common Crawl version)~\cite{pennington2014glove} to embed words in the vocabulary into 300-dimensional vectors. Following~\cite{wang2018glue}, we use AllenNLP~\cite{Gardner2017AllenNLP} open source library to facilitate the data processing, as well as model training and evaluation.

\subsection{NYUD2-DIR}
We create NYUD2-DIR based on the NYU Depth Dataset V2~\cite{Silberman:ECCV12}, which provides images and depth maps for different indoor scenes. Our task is to predict the depth maps from the RGB scene images. The depth maps have an upper bound of 10 meters and a lower bound of 0.7 meters. Following standard practices~\cite{hu2019revisiting, bhat2020adabins}, we use 50K images for training and 654 images for testing. We set the bin length to 0.1 meter and the number of pixels per bin varies between $1.13\times 10^6$ and $1.46\times 10^8$. Besides, we randomly select 9,357 test pixels (the minimum number of bin pixels in the test set) for each bin from 654 test images to make the test set balanced, with a total of $8.70\times 10^5$ test pixels in the NYUD2-DIR test set, as indicated in Table~\ref{table:dir-dataset-overview}.

Following~\cite{hu2019revisiting}, for both training and evaluation phases, we first downsample images (both RGB and depth) from original size $640\times 480$ to $320\times 240$ using bilinear interpolation, then conduct center crop to obtain images of size $304\times 228$, and finally normalize them into $[0,1]$. Note that our pixel statistics are calculated and selected based on this resolution. For training, we further downsample the depth maps to $114\times 152$ to fit the size of outputs. Additionally, we also employ the following data argumentation methods during training: (1) Flip: randomly flip both RGB and depth images horizontally with probability of 0.5; (2) Rotation: rotate both RGB and depth images by a random degree from -5 to 5; (3) Color Jitter: randomly scale the brightness, contrast, and saturation values of the RGB images by $c\in [0.6, 1.4]$.

\subsection{SHHS-DIR}
We create SHHS-DIR based on the SHHS dataset \cite{quan1997sleep}, which contains full-night Polysomnography (PSG) signals from 2,651 subjects. The signal length for each subject varies from 7,278 seconds to 45,448 seconds. Available PSG signals include Electroencephalography (EEG), Electrocardiography (ECG), and breathing signals (airflow, abdomen, and thorax). In the experiments, we consider all of these PSG signals as high-dimensional information, and use them as inputs.
Specifically, we first preprocess both EEG and ECG signals to transform them from time domain to the frequency domain using the short-time Fourier transform (STFT), and get the dense EEG spectrograms $\mathbf{x}_e \in \mathbb{R}^{64\times l_i}$ and ECG spectrograms $\mathbf{x}_c \in \mathbb{R}^{22\times l_i}$, where $l_i \in [7278, 45448]$ is the signal length for the $i$-th subject. For the breathing signals, we use the original time series with a sampling rate of 10Hz, resulting in the high-dimensional input as $\mathbf{x}_b \in \mathbb{R}^{3\times 10 l_i}$, where the three different breathing sources are concatenated as different channels.

The dataset also includes the 36-Item Short Form Health Survey (SF-36) \cite{ware1992mos} for each subject, where a General Health score is extracted. We use the score as the target value, and formulate the task as predicting the General Health score for different subjects from their PSG signals~(i.e., $\mathbf{x}_e,\mathbf{x}_c,\mathbf{x}_b$).
The training set of SHHS-DIR contains 1,892 samples (subjects), and the validation set and test set are made balanced over the health score with 369 samples each.
We set the length of each bin to be 1, with a minimum score of 0 and a maximum score of 100.
The number of samples per bin varies between 0 and 275, indicating the missing data issue in certain target bins.

\section{Experimental Settings}
\label{appendix:experiment-details}

\subsection{Implementation Details}
\label{appendix:experiment-details-implementation}

\textbf{IMDB-WIKI-DIR \& AgeDB-DIR.}
We use ResNet-50 model~\cite{he2016deep} for all IMDB-WIKI-DIR and AgeDB-DIR experiments. We train all models for 90 epochs using the Adam optimizer~\cite{kingma2014adam}, with an initial learning rate of $10^{-3}$ and then decayed by 0.1 at the 60-th and 80-th epoch, respectively. We mainly employ the $L_1$ loss throughout the experiments, and fix the batch size as 256.

For both LDS and FDS, we use the Gaussian kernel for distribution smoothing, with the kernel size $l=5$ and the standard deviation $\sigma=2$.
We study different choices of kernel types, training losses, and hyper-parameter values in Sec.~\ref{appendix:ablation-kernel-type}, \ref{appendix:ablation-diff-training-loss}, and \ref{appendix:ablation-hyper-param}.
For the implementation of FDS, we simply use the feature variance instead of covariance for better computational efficiency. The momentum of FDS is fixed as 0.9.
As for the baseline methods, we set $\beta=0.2$ and $\gamma=1$ for \textsc{Focal-R}. For \textsc{RRT}, in the second training stage, we employ an initial learning rate of $10^{-4}$ with total training epochs of 30. For \textsc{SmoteR} and \textsc{SMOGN}, we divide the target range based on a manually defined relevance method, under-sample majority regions, and over-sample minority regions by either interpolating with selected nearest neighbors~\cite{torgo2013smoter} or also adding Gaussian noise perturbation~\cite{branco2017smogn}. We use pixel-wise Euclidean distance to define the image distance, which is further used to determine nearest neighbors, and set Gaussian perturbation ratio as 0.1 for \textsc{SMOGN}.

\textbf{STS-B-DIR.}
Following~\cite{wang2018glue}, we use 300D GloVe word embeddings (840B Common Crawl version)~\cite{pennington2014glove} and a two-layer, 1500D (per direction) BiLSTM with max pooling to encode the paired sentences into independent vectors $u$ and $v$, and then pass $[u;v;|u-v|;uv]$ to a regressor. We train all models using the Adam optimizer with a fixed learning rate $10^{-4}$. We validate the model every 10 epochs, use MSE as the validation metric, and stop training when performance does not improve, i.e., validation error does not decrease, after 10 validation checks. We employ the MSE loss throughout the experiments and fix the batch size as 128. 

We use the same hyper-parameter settings for both LDS and FDS as in the IMDB-WIKI-DIR experiments. For the baselines, we employ MSE-based \textsc{Focal-R} and set $\beta=20$ and $\gamma=1$. For \textsc{RRT}, the hyper-parameter settings remain the same between the first and the second training stage. For \textsc{SmoteR} and \textsc{SMOGN}, we use the Euclidean distance between the word embeddings to measure the sentence distance and do interpolation or Gaussian noise argumentation based on the word embeddings. We set Gaussian perturbation ratio as 0.1 and the number of neighbors $k=7$.
For STS-B-DIR, we define \emph{many-shot region} as bins with over 100 training samples, \emph{medium-shot region} with 30$\sim$100 training samples, and \emph{few-shot region} with under 30 training samples.

\textbf{NYUD2-DIR.}
We use ResNet-50-based encoder-decoder architecture proposed by~\cite{hu2019revisiting} for all NYUD2-DIR experiments, which consists of an encoder, a decoder, a multi-scale feature fusion module, and a refinement module. We train all models for 20 epochs using Adam optimizer with an initial learning rate of $10^{-4}$ and then decayed by 0.1 every 5 epochs. To better evaluate the performance of our methods, we simply use the MSE loss as the depth loss without adding the gradient and surface normal losses as in~\cite{hu2019revisiting}. We fix the batch size as 32 for all experiments. We use the same hyper-parameter settings for both LDS and FDS as in the IMDB-WIKI-DIR experiments.
For NYUD2-DIR, \emph{many-shot region} is defined as bins with over $2.6\times 10^7$ training pixels, \emph{medium-shot region} as bins with $1.0\times 10^7 \sim 2.6\times 10^7$ training pixels, and \emph{few-shot region} as bins with under $1.0\times 10^7$ training pixels.

\textbf{SHHS-DIR.}
Following~\cite{wang2019bidirectional}, we use a CNN-RNN network architecture for SHHS-DIR experiments. The network first employs three encoders with the same architecture to encode the high-dimensional EEG $\mathbf{x}_e$, ECG $\mathbf{x}_c$, and breathing signals $\mathbf{x}_b$ into fixed-length vectors (each with 256 dimensions). The encodings are then concatenated and sent to a 3-layer MLP regression network to produce the output value.
Each of the encoder uses the ResNet block~\cite{he2016deep} with 1D convolution as the CNN components, and employs the simple recurrent units~(SRU)~\cite{lei2018simple} as the RNN components.
We train all models for 80 epochs using the Adam optimizer with a learning rate of $10^{-3}$, and remain all other hyper-parameters the same as~\cite{wang2019bidirectional}. We use the same hyper-parameter settings for both LDS and FDS, as well as other baseline methods as in the IMDB-WIKI-DIR experiments.

\subsection{Evaluation Metrics}

We describe in detail all the evaluation metrics we used in our experiments.

\textbf{MAE.} The mean absolute error~(MAE) is defined as $\frac{1}{N}\sum_{i=1}^{N}|y_i-\hat{y}_i|$, which represents the averaged absolute difference between the ground truth and predicted values over all samples.

\textbf{MSE \& RMSE.} The mean squared error~(MSE) is defined as $\frac{1}{N}\sum_{i=1}^{N}(y_i-\hat{y}_i)^2$, which represents the averaged squared difference between the ground truth and predicted values over all samples. 
The root mean squared error~(RMSE) is computed by simply taking the square root of MSE.

\textbf{GM.} We propose another evaluation metric for regression, called error Geometric Mean (\textbf{GM}), and is defined as $(\prod_{i=1}^N e_i)^{\frac{1}{N}}$, where $e_i\triangleq|y_i-\hat{y}_i|$ represents the $L_1$ error of each sample. GM aims to characterize the fairness (uniformity) of model predictions using the geometric mean instead of the arithmetic mean over the prediction errors.

\textbf{Pearson correlation \& Spearman correlation.} Following the common evaluation practice as in the STS-B~\cite{cer2017semeval} and the GLUE benchmark~\cite{wang2018glue}, we employ Pearson correlation as well as Spearman correlation for performance evaluation on STS-B-DIR, where Pearson correlation evaluates the linear relationship between predictions and corresponding ground truth values, and Spearman correlation evaluates the monotonic rank-order relationship.

\textbf{Mean $\log_{10}$ error \& Threshold accuracy.}
For NYUD2-DIR, we further use several standard depth estimation evaluation metrics proposed by~\cite{eigen2014depth}: 
Mean $\log_{10}$ error ($\log_{10}$), which is expressed as $\frac{1}{N} \sum_{i=1}^{N}\left|\log _{10} d_{i}-\log _{10} g_{i}\right|$; Threshold accuracy ($\delta_i$), which is defined as the percentage of $d_i$ such that $\max \left(\frac{d_{i}}{g_{i}}, \frac{g_{i}}{d_{i}}\right)=\delta_i < 1.25^i$ ($i=1,2,3$). Here, $g_i$ denotes the value of a pixel in the ground truth depth image, $d_i$ represents the value of its corresponding pixel in the predicted depth image, and $N$ is the total number of evaluation pixels.

\section{Additional Results}
\label{appendix:additional-results}

We provide complete evaluation results on the five DIR datasets, where more baselines and evaluation metrics are included in addition to the reported results in the main paper.

\subsection{Complete Results on IMDB-WIKI-DIR}

We include more baseline methods for comparison on IMDB-WIKI-DIR. Specifically, the following two baselines are added for comparison in the group of \emph{Synthetic samples} strategies:
\begin{itemize}[leftmargin=*]
    \item \textbf{Mixup}~\cite{zhang2018mixup}: \textsc{Mixup} trains a deep model using samples created by the convex combinations of pairs of inputs and corresponding labels. It has shown promising results on improving the generalization of deep models as a regularization technique.
    \item \textbf{Manifold-Mixup}~(\textsc{M-Mixup})~\cite{verma2019manifold}: \textsc{M-Mixup} extends the idea of \textsc{Mixup} from input space to the hidden representation space, where the linear interpolations are performed in (multiple) deep hidden layers.
\end{itemize}
We note that both \textsc{Mixup} and \textsc{M-Mixup} are not tailored for imbalanced regression problems, but share similarities with \textsc{SmoteR} and \textsc{SMOGN} as synthetic samples are constructed. The differences lie in the fact that \textsc{Mixup} and \textsc{M-Mixup} create virtual samples (either in input space or feature space) on the fly during network training, while \textsc{SmoteR} and \textsc{SMOGN} operate on a newly generated and fixed dataset for training.
We set $\alpha=0.2$ for \textsc{Mixup} in implementation, and set $\alpha=0.2$ as well and eligible layers $\mathcal{S}=\{0, 1, 2, 3\}$ for \textsc{M-Mixup}.
In addition, for \textsc{Inv} which re-weights the loss based on the inverse frequency in the empirical label distribution, we further clip the maximum weight to be at most 200$\times$ larger than the minimum weight to avoid extreme loss values.

\begin{table*}[t]
\setlength{\tabcolsep}{6pt}
\caption{Complete evaluation results on IMDB-WIKI-DIR.}
\vspace{-2pt}
\label{table:appendix-complete-imdb-wiki}
\small
\begin{center}
\resizebox{\textwidth}{!}{
\begin{tabular}{l|cccc|cccc|cccc}
\toprule[1.5pt]
Metrics      & \multicolumn{4}{c|}{MSE~$\downarrow$}     & \multicolumn{4}{c|}{MAE~$\downarrow$}     & \multicolumn{4}{c}{GM~$\downarrow$}  \\ \midrule
Shot         & All  & Many & Med. & Few   & All  & Many & Med. & Few   & All  & Many & Med. & Few   \\ \midrule\midrule
\textsc{Vanilla}      &  138.06 &  108.70  & 366.09  & 964.92 & 8.06 & 7.23 & 15.12  & 26.33 & 4.57 & 4.17 & 10.59  & 20.46 \\[1.2pt]
\textsc{Vanilla} + \textbf{\textsc{LDS}}   & 131.65 & 109.04 & \textbf{298.98} & 829.35  & {7.83} & 7.31 & \textbf{12.43}  & {22.51} & {4.42} & {4.19} & \textbf{7.00}  & {13.94} \\[1.2pt]
\textsc{Vanilla} + \textbf{\textsc{FDS}}   & 133.81 & 107.51 & 332.90 & 916.18 & 7.85 & \textbf{7.18} & 13.35 & 24.12 & 4.47 & 4.18 & 8.18 & 15.18 \\[1.2pt]
\textsc{Vanilla} + \textbf{\textsc{LDS}} + \textbf{\textsc{FDS}} & \textbf{129.35} & \textbf{106.52} & 311.49 & \textbf{811.82} & \textbf{7.78} & {7.20} & {12.61} & \textbf{22.19} & \textbf{4.37}    & \textbf{4.12} & 7.39  & \textbf{12.61} \\ \midrule\midrule
\textsc{Mixup}~\cite{zhang2018mixup}   &  141.11 & 109.13 & 389.95 &    1037.98   & 8.22  & \textbf{7.29} & 16.23 & 28.11  & 4.68  & \textbf{4.22} & 12.28 &  23.55 \\[1.2pt]
\textsc{M-Mixup}~\cite{verma2019manifold}  & 137.45  & \textbf{108.33} & 363.72 & 957.53 & 8.22  & 7.39 & 15.24 & 26.70 & 4.80  & 4.39 & 10.85 & 21.86  \\[1.2pt]
\textsc{SmoteR}~\cite{torgo2013smoter}   & 138.75  & 111.55 & 346.09 &  935.89     & 8.14 & 7.42 & 14.15  & 25.28 & 4.64 & {4.30} & 9.05   & 19.46 \\[1.2pt]
\textsc{SMOGN}~\cite{branco2017smogn}    & 136.09  & 109.15 & 339.09 &  944.20     & 8.03 & {7.30} & 14.02  & 25.93 & 4.63 & {4.30} & 8.74   & 20.12 \\[1.2pt]
\textsc{SMOGN} + \textbf{\textsc{LDS}}   & 137.31 & 111.79 & 333.15 & 823.07 & 8.02 & 7.39 & 13.71 & 23.22 & 4.63 & 4.39 & 8.71 & 15.80 \\[1.2pt]
\textsc{SMOGN} + \textbf{\textsc{FDS}}   & 137.82 & 109.42 & 340.65 & 847.96 & 8.03 & 7.35 & 14.06 & 23.44 & 4.65 & 4.33 & 8.87 & 16.00 \\[1.2pt]
\textsc{SMOGN} + \textbf{\textsc{LDS}} + \textbf{\textsc{FDS}}   & \textbf{135.26} & 110.91 & \textbf{326.52} & \textbf{808.45} & \textbf{7.97} & 7.38 & \textbf{13.22} & \textbf{22.95} & \textbf{4.59} & 4.39 & \textbf{7.84} & \textbf{14.94} \\ \midrule\midrule
\textsc{Focal-R}  & 136.98 & 106.87 & 368.60 & 1002.90 & 7.97 & 7.12 & 15.14  & 26.96 & 4.49 & 4.10 & 10.37  & 21.20 \\[1.2pt]
\textsc{Focal-R} + \textbf{\textsc{LDS}} & 132.81 & 105.62 & 354.37 & 949.03 & {7.90}  & \textbf{7.10}  &  14.72   & 25.84  & \textbf{4.47}   & \textbf{4.09}    & {10.11}     & {19.14}     \\[1.2pt]
\textsc{Focal-R} + \textbf{\textsc{FDS}} & 133.74 & 105.35 & 351.00 & 958.91 & 7.96 & 7.14 & 14.71 & 26.06 & 4.51 & 4.12 & 10.16 & 19.56 \\[1.2pt]
\textsc{Focal-R} + \textbf{\textsc{LDS}} + \textbf{\textsc{FDS}} & \textbf{132.58} & \textbf{105.33} & \textbf{338.65} & \textbf{944.92} & \textbf{7.88} & \textbf{7.10} & \textbf{14.08}  & \textbf{25.75}   & \textbf{4.47} & 4.11 & \textbf{9.32} & \textbf{18.67} \\ \midrule\midrule
\textsc{RRT}    & 132.99 & 105.73 & 341.36 & 928.26 & 7.81 & 7.07 & 14.06  & 25.13 & 4.35 & 4.03 & 8.91   & 16.96 \\[1.2pt]
\textsc{RRT} + \textbf{\textsc{LDS}} & 132.91 & 105.97 & 338.98 & 916.98 & {7.79} &  7.08  & {13.76}  & {24.64} & {4.34} & \textbf{4.02} & {8.72} & {16.92} \\[1.2pt]
\textsc{RRT} + \textbf{\textsc{FDS}} & 129.88 & \textbf{104.63} & 310.69 & 890.04 & \textbf{7.65}  & \textbf{7.02} & 12.68 & 23.85 & \textbf{4.31} & 4.03 & 7.58 & 16.28 \\[1.2pt]
\textsc{RRT} + \textbf{\textsc{LDS}} + \textbf{\textsc{FDS}}  & \textbf{129.14} & 105.92 & \textbf{306.69} & \textbf{880.13} & \textbf{7.65}  & 7.06  & \textbf{12.41}  & \textbf{23.51}  & \textbf{4.31} & {4.07} & \textbf{7.17} & \textbf{15.44} \\ \midrule\midrule
\textsc{Inv}   & 139.48 & 116.72 & 305.19 & 869.50 & 8.17 & 7.64 & 12.46 & 22.83 & 4.70 & 4.51 & 6.94 & 13.78 \\[1.2pt]
\textsc{SQInv}   & 134.36 & 111.23 & 308.63 &  834.08 & 7.87 & 7.24 & 12.44  & 22.76 & 4.47 & 4.22 & 7.25   & 15.10 \\[1.2pt]
\textsc{SQInv} + \textbf{\textsc{LDS}} & 131.65 & 109.04 & \textbf{298.98} & 829.35 & {7.83} & 7.31 & \textbf{12.43}  & {22.51} & {4.42} & {4.19} & {7.00}  & {13.94} \\[1.2pt]
\textsc{SQInv} + \textbf{\textsc{FDS}} & 132.64 & 109.28 & 311.35 & 851.06 & 7.83 & 7.23 & 12.60  & 22.37  & 4.42 & 4.20 & \textbf{6.93} & 13.48  \\[1.2pt]
\textsc{SQInv} + \textbf{\textsc{LDS}} + \textbf{\textsc{FDS}} & \textbf{129.35} & \textbf{106.52} & 311.49 & \textbf{811.82} & \textbf{7.78} & \textbf{7.20} & {12.61} & \textbf{22.19} & \textbf{4.37}    & \textbf{4.12} & 7.39  & \textbf{12.61}  \\ \midrule\midrule
\textsc{\textbf{Ours~(best)} vs. Vanilla} & \textcolor{darkgreen}{\textbf{+8.92}} & \textcolor{darkgreen}{\textbf{+4.07}} & \textcolor{darkgreen}{\textbf{+67.11}} & \textcolor{darkgreen}{\textbf{+156.47}} & \textcolor{darkgreen}{\textbf{+0.41}} & \textcolor{darkgreen}{\textbf{+0.21}} & \textcolor{darkgreen}{\textbf{+2.71}} & \textcolor{darkgreen}{\textbf{+4.14}} & \textcolor{darkgreen}{\textbf{+0.26}} & \textcolor{darkgreen}{\textbf{+0.15}} & \textcolor{darkgreen}{\textbf{+3.66}} & \textcolor{darkgreen}{\textbf{+7.85}} \\
\bottomrule[1.5pt]
\end{tabular}
}
\end{center}
\vspace{-0.4cm}
\end{table*}

We show the complete results in Table~\ref{table:appendix-complete-imdb-wiki}. As the table illustrates, both \textsc{Mixup} and \textsc{M-Mixup} can improve the performance in the many-shot region, but lead to negligible improvements in the medium-shot and few-shot regions. In contrast, adding both FDS and LDS can substantially improve the results, especially for the underrepresented regions. Finally, FDS and LDS lead to remarkable improvements when compared to the \textsc{Vanilla} model across all evaluation metrics.

\subsection{Complete Results on AgeDB-DIR}

We provide complete evaluation results for AgeDB-DIR in Table~\ref{table:appendix-complete-agedb}. Similar to IMDB-WIKI-DIR, within each group of techniques, adding either LDS, FDS, or both can lead to performance gains, while LDS $+$ FDS often achieves the best results.
Overall, for different groups of strategies, both FDS and LDS consistently boost the performance, where the larger gains come from the medium-shot and few-shot regions.

\begin{table*}[t]
\setlength{\tabcolsep}{7.5pt}
\caption{Complete evaluation results on AgeDB-DIR.}
\vspace{-2pt}
\label{table:appendix-complete-agedb}
\small
\begin{center}
\resizebox{\textwidth}{!}{
\begin{tabular}{l|cccc|cccc|cccc}
\toprule[1.5pt]
Metrics      & \multicolumn{4}{c|}{MSE~$\downarrow$}     & \multicolumn{4}{c|}{MAE~$\downarrow$}     & \multicolumn{4}{c}{GM~$\downarrow$}  \\ \midrule
Shot         & All  & Many & Med. & Few  & All  & Many & Med. & Few   & All  & Many & Med. & Few   \\ \midrule\midrule
\textsc{Vanilla}  & 101.60 & 78.40 & 138.52 & 253.74 & 7.77 & {6.62} & 9.55   & 13.67 & 5.05 & {4.23} & 7.01   & 10.75 \\[1.2pt]
\textsc{Vanilla} + \textbf{\textsc{LDS}}   & 102.22 & 83.62 & 128.73 & \textbf{204.64} & 7.67 & {6.98} & 8.86   & 10.89 & 4.85 & {4.39} &  5.80  & {7.45} \\[1.2pt]
\textsc{Vanilla} + \textbf{\textsc{FDS}}   & \textbf{98.55} & \textbf{75.06} & 123.58 & 235.70 & \textbf{7.55} & \textbf{6.50} & 8.97  & 13.01 & 4.75 & \textbf{4.03} &  6.42  & 9.93 \\[1.2pt]
\textsc{Vanilla} + \textbf{\textsc{LDS}} + \textbf{\textsc{FDS}} & 99.46 & 84.10 & \textbf{112.20} & 209.27 & \textbf{7.55} &  7.01  & \textbf{8.24}   & \textbf{10.79} & \textbf{4.72} & 4.36 & \textbf{5.45}  & \textbf{6.79} \\ \midrule\midrule
\textsc{SmoteR}~\cite{torgo2013smoter}   & 114.34 & 93.35 & 129.89 & 244.57 & 8.16  & 7.39   & 8.65 & 12.28 & 5.21    & 4.65    & 5.69 & 8.49    \\[1.2pt]
\textsc{SMOGN}~\cite{branco2017smogn}    & 117.29 & 101.36 & 133.86 & 232.90 & 8.26  & 7.64   & 9.01 & 12.09 & 5.36    & 4.90    & 6.19 & 8.44    \\[1.2pt]
\textsc{SMOGN} + \textbf{\textsc{LDS}}   & 110.43 & 93.73 & 124.19 & 229.35 & 7.96 & 7.44 & 8.64  & 11.77 & 5.03 & 4.68 & 5.69  & 7.98  \\[1.2pt]
\textsc{SMOGN} + \textbf{\textsc{FDS}}   & 112.42 & 97.68 & 131.37 & 233.30 & 8.06 & 7.52 & 8.75  & 11.89 & 5.02 & 4.66 & 5.63  & 8.02  \\[1.2pt]
\textsc{SMOGN} + \textbf{\textsc{LDS}} + \textbf{\textsc{FDS}}   & \textbf{108.41} & \textbf{91.58} & \textbf{120.28} & \textbf{218.59} & \textbf{7.90} & \textbf{7.32}  & \textbf{8.51}  & \textbf{11.19}  & \textbf{4.98}  & \textbf{4.64}  & \textbf{5.41} & \textbf{7.35} \\ \midrule\midrule
\textsc{Focal-R}   & 101.26 & 77.03 & 131.81 & 252.47 & 7.64 & 6.68 & 9.22   & 13.00 & 4.90 & 4.26 & 6.39   & 9.52  \\[1.2pt]
\textsc{Focal-R} + \textbf{\textsc{LDS}}   &  98.80 &  77.14 & 125.53 & 229.36 & 7.56 & \textbf{6.67} & 8.82   & 12.40 & 4.82 & 4.27 & 5.87   & 8.83  \\[1.2pt]
\textsc{Focal-R} + \textbf{\textsc{FDS}}   & 100.14  & 80.97 & 121.84 & \textbf{221.15} & 7.65 & 6.89 & 8.70   & \textbf{11.92} & 4.83 & 4.32 & 5.89   & \textbf{8.04}  \\[1.2pt]
\textsc{Focal-R} + \textbf{\textsc{LDS}} + \textbf{\textsc{FDS}}  & \textbf{96.70} & \textbf{76.11} & \textbf{115.86} & 238.25 & \textbf{7.47} & 6.69 & \textbf{8.30}  & 12.55 & \textbf{4.71} & \textbf{4.25} & \textbf{5.36}  & 8.59  \\ \midrule\midrule
\textsc{RRT}   & 102.89 & 83.37 & 125.66 & 224.27 & 7.74 & 6.98 & 8.79   & 11.99 & 5.00 & 4.50 & 5.88   & 8.63  \\[1.2pt]
\textsc{RRT} + \textbf{\textsc{LDS}} & 102.63 & 83.93 & 126.01 & 214.66 & {7.72} & 7.00 & {8.75}   & {11.62} & {4.98} & 4.54 & {5.71}   & {8.27}  \\[1.2pt]
\textsc{RRT} + \textbf{\textsc{FDS}} & 102.09 & 84.49 & 122.89 & 224.05 & {7.70} & \textbf{6.95} & {8.76}   & {11.86} & {4.82} & \textbf{4.32} & {5.83}   & {8.08}  \\[1.2pt]
\textsc{RRT} + \textbf{\textsc{LDS}} + \textbf{\textsc{FDS}} & \textbf{101.74} & \textbf{83.12} & \textbf{121.08} & \textbf{210.78} & \textbf{7.66} & 6.99 & \textbf{8.60}   & \textbf{11.32} & \textbf{4.80} & 4.42 & \textbf{5.53}   & \textbf{6.99}  \\ \midrule\midrule
\textsc{Inv}   & 110.24  & 91.93 & 130.68 & 211.92 & 7.97 & 7.31 & 8.81 & 11.62 & 5.05 & 4.64 & 5.75 & 8.20 \\[1.2pt]
\textsc{SQInv}   & 105.14 & 87.21 & 127.66 & 212.30 & 7.81 & 7.16 & 8.80   & 11.20 & 4.99 & 4.57 & 5.73   & 7.77  \\[1.2pt]
\textsc{SQInv} + \textbf{\textsc{LDS}} & 102.22 & \textbf{83.62} & {128.73} & 204.64 & 7.67 & \textbf{6.98} & 8.86   & 10.89 & 4.85 & {4.39} &  5.80  & {7.45}  \\[1.2pt]
\textsc{SQInv} + \textbf{\textsc{FDS}} & 101.67 & 86.49 & 129.61 & \textbf{167.75} & 7.69 & 7.10 & 8.86   & \textbf{9.98} & 4.83 & {4.41} &  5.97  & \textbf{6.29}  \\[1.2pt]
\textsc{SQInv} + \textbf{\textsc{LDS}} + \textbf{\textsc{FDS}} & \textbf{99.46} & 84.10 & \textbf{112.20} & 209.27 & \textbf{7.55} &  7.01  & \textbf{8.24}   & 10.79 & \textbf{4.72} & \textbf{4.36} & \textbf{5.45}  & {6.79}  \\ \midrule\midrule
\textsc{\textbf{Ours~(best)} vs. Vanilla}  & \textcolor{darkgreen}{\textbf{+4.90}} & \textcolor{darkgreen}{\textbf{+3.34}} & \textcolor{darkgreen}{\textbf{+26.32}} & \textcolor{darkgreen}{\textbf{+85.99}} & \textcolor{darkgreen}{\textbf{+0.30}} & \textcolor{darkgreen}{\textbf{+0.12}} & \textcolor{darkgreen}{\textbf{+1.31}} & \textcolor{darkgreen}{\textbf{+3.69}} & \textcolor{darkgreen}{\textbf{+0.34}} & \textcolor{darkgreen}{\textbf{+0.20}} & \textcolor{darkgreen}{\textbf{+1.65}} & \textcolor{darkgreen}{\textbf{+4.46}} \\
\bottomrule[1.5pt]
\end{tabular}
}
\end{center}
\vspace{-0.65cm}
\end{table*}

\begin{table}[H]
\setlength{\tabcolsep}{6.5pt}
\caption{Complete evaluation results on STS-B-DIR.}
\vspace{-2pt}
\label{table:appendix-complete-sts-b}
\small
\begin{center}
\resizebox{\textwidth}{!}{
\begin{tabular}{l|cccc|cccc|cccc|cccc}
\toprule[1.5pt]
Metrics      & \multicolumn{4}{c|}{MSE~$\downarrow$}       
& \multicolumn{4}{c|}{MAE~$\downarrow$}
& \multicolumn{4}{c|}{Pearson correlation~(\%)~$\uparrow$}
& \multicolumn{4}{c}{Spearman correlation~(\%)~$\uparrow$} \\ \midrule
Shot         & All   & Many  & Med. & Few   & All   & Many  & Med. & Few & All   & Many  & Med. & Few & All   & Many  & Med. & Few   \\ \midrule\midrule
\textsc{Vanilla}      & 0.974 & 0.851 & 1.520  & 0.984 & 0.794 & 0.740 & 1.043 & 0.771 & 74.2 & 72.0 & 62.7  & 75.2 & 74.4 & 68.8 & 50.5 & \textbf{75.0}\\[1.2pt]
\textsc{Vanilla} + \textbf{\textsc{LDS}} & 0.914 & 0.819 & 1.319 & 0.955 & 0.773 & 0.729 & 0.970 & 0.772  & 75.6 & 73.4 & 63.8 & 76.2 & 76.1 &  70.4 & \textbf{55.6} & 74.3 \\[1.2pt]
\textsc{Vanilla} + \textbf{\textsc{FDS}} & 0.916 & 0.875 & \textbf{1.027} & 1.086 & 0.767 & 0.746 & \textbf{0.840} & 0.811  & 75.5 & 73.0 & \textbf{67.0} & 72.8 & 75.8 & 69.9 & 54.4 & 72.0 \\[1.2pt]
\textsc{Vanilla} + \textbf{\textsc{LDS}} + \textbf{\textsc{FDS}} & \textbf{0.907} & \textbf{0.802} & 1.363  & \textbf{0.942}  & \textbf{0.766} & \textbf{0.718} & 0.986 & \textbf{0.755}   & \textbf{76.0}     & \textbf{74.0}     & 65.2      & \textbf{76.6} & \textbf{76.4} & \textbf{70.7} & 54.9 & 74.9 \\ \midrule\midrule
\textsc{SmoteR}~\cite{torgo2013smoter}       & 1.046     & 0.924     & 1.542      & 1.154     & 0.834 & 0.782 & 1.052 & 0.861 & 72.6     & 69.3     & 65.3      & 70.6 & 72.6 & 65.6 & \textbf{55.6} & 69.1    \\[1.2pt]
\textsc{SMOGN}~\cite{branco2017smogn}        & 0.990     & 0.896     & 1.327      & 1.175  & 0.798 & 0.755 & 0.967 & 0.848   & 73.2     & 70.4     & 65.5      & 69.2  & 73.2 & 67.0 & 55.1 & 67.0   \\[1.2pt]
\textsc{SMOGN} + \textbf{\textsc{LDS}}   & 0.962     & 0.880     & 1.242      & 1.155    & 0.787 & 0.748 & 0.944 & 0.837  & 74.0     & 71.5     & 65.2      & 69.8   & 74.3 &  68.5 & 53.6 & 67.1 \\[1.2pt]
\textsc{SMOGN} + \textbf{\textsc{FDS}}   & 0.987    & 0.945    & \textbf{1.101}      & 1.153     & 0.796 & 0.776 & \textbf{0.864} & 0.838 & 73.0    & 69.6    & \textbf{68.5}      & 69.9   & 72.9 & 66.0 & 54.3 & 68.0  \\[1.2pt]
\textsc{SMOGN} + \textbf{\textsc{LDS}} + \textbf{\textsc{FDS}}   &  \textbf{0.950}   & \textbf{0.851}    & 1.327      & \textbf{1.095} & \textbf{0.785} & \textbf{0.738} & 0.987 & \textbf{0.799}     & \textbf{74.6}    & \textbf{72.1}    & 65.9      & \textbf{71.7}   & \textbf{75.0} & \textbf{68.9} & 54.4 & \textbf{70.3}  \\ \midrule\midrule
\textsc{Focal-R}      & 0.951 & 0.843 & 1.425  & 0.957 & 0.790 & 0.739 & 1.028 & 0.759 & 74.6 & 72.3 & 61.8  & 76.4 & 75.0 & 69.4 & 51.9 & 75.5\\[1.2pt]
\textsc{Focal-R} + \textbf{\textsc{LDS}} & 0.930     & \textbf{0.807}     & 1.449      & 0.993    & 0.781 & \textbf{0.723} & 1.031 & 0.801 & \textbf{75.7}     & \textbf{73.9}     & 62.4      & 75.4 & \textbf{76.2} & \textbf{71.2} & 50.7 & 74.7\\[1.2pt]
\textsc{Focal-R} + \textbf{\textsc{FDS}} & \textbf{0.920} & 0.855 & \textbf{1.169}  & 1.008 & \textbf{0.775} & 0.743 & \textbf{0.903} & 0.804 & 75.1     & 72.6   & \textbf{66.4}      & 74.7 & 75.4 & 69.4 & \textbf{52.7} & 75.4 \\[1.2pt]
\textsc{Focal-R} + \textbf{\textsc{LDS}} + \textbf{\textsc{FDS}} & 0.940 & 0.849 & 1.358  & \textbf{0.916}     & 0.785 & 0.737 & 0.984 & \textbf{0.732} & 74.9     & 72.2     & 66.3      & \textbf{77.3} & 75.1 & 69.2 & 52.5 & \textbf{76.4}\\ \midrule\midrule
\textsc{RRT}         & 0.964 & 0.842 & 1.503  & 0.978 & 0.793 & 0.739 & 1.044 & 0.768 & 74.5 & 72.4 & 62.3  & 75.4 & 74.7 & 69.2 & 51.3 & \textbf{74.7}\\[1.2pt]
\textsc{RRT} + \textbf{\textsc{LDS}}     & {0.916} & 0.817 & {1.344}  & 0.945 & 0.772 & 0.727 & 0.980 & \textbf{0.756} & {75.7} & {73.5} & {64.1}  & {76.6} & 76.1 & 70.4 & 53.2 & 74.2\\[1.2pt]
\textsc{RRT} + \textbf{\textsc{FDS}} & 0.929 & 0.857 & \textbf{1.209}  & 1.025  & 0.769 & 0.736 & \textbf{0.905} & 0.795   & 74.9     & 72.1     & \textbf{67.2}      & 74.0 & 75.0 & 69.1 & 52.8 & 74.6\\[1.2pt]
\textsc{RRT} + \textbf{\textsc{LDS}} + \textbf{\textsc{FDS}} & \textbf{0.903} & \textbf{0.806} & 1.323  & \textbf{0.936}  & \textbf{0.764} & \textbf{0.719} & 0.965 & 0.760   &  \textbf{76.0}     &\textbf{73.8}     & 65.2      & \textbf{76.7}  & \textbf{76.4} & \textbf{70.8} & \textbf{54.7} & \textbf{74.7}\\ \midrule\midrule
\textsc{Inv}      & 1.005 & 0.894 & 1.482  & 1.046 & 0.805 & 0.761 & 1.016 & 0.780 & 72.8 & 70.3 & 62.5  & 73.2 & 73.1 & 67.2 & 54.1 & 71.4\\[1.2pt]
\textsc{Inv} + \textbf{\textsc{LDS}} & 0.914 & 0.819 & 1.319 & 0.955 & 0.773 & 0.729 & 0.970 & 0.772  & 75.6 & 73.4 & 63.8 & 76.2 & 76.1 &  70.4 & \textbf{55.6} & 74.3 \\[1.2pt]
\textsc{Inv} + \textbf{\textsc{FDS}} & 0.927 & 0.851 & \textbf{1.225}  & 1.012   & 0.771 & 0.740 & \textbf{0.914} & 0.756  & 75.0     & 72.4     & \textbf{66.6}      & 74.2  & 75.2 & 69.2 & 55.2 & 74.8\\[1.2pt]
\textsc{Inv} + \textbf{\textsc{LDS}} + \textbf{\textsc{FDS}} & \textbf{0.907} & \textbf{0.802} & 1.363  & \textbf{0.942}  & \textbf{0.766} & \textbf{0.718} & 0.986 & \textbf{0.755}   & \textbf{76.0}     & \textbf{74.0}     & 65.2      & \textbf{76.6} & \textbf{76.4} & \textbf{70.7} & 54.9 & \textbf{74.9}\\ \midrule\midrule
\textsc{\textbf{Ours~(best)} vs. Vanilla}   & \textcolor{darkgreen}{\textbf{+.071}} & \textcolor{darkgreen}{\textbf{+.049}} & \textcolor{darkgreen}{\textbf{+.419}} & \textcolor{darkgreen}{\textbf{+.068}} &
\textcolor{darkgreen}{\textbf{+.030}} & \textcolor{darkgreen}{\textbf{+.022}} & \textcolor{darkgreen}{\textbf{+.203}} & \textcolor{darkgreen}{\textbf{+.039}} &
\textcolor{darkgreen}{\textbf{+1.8}} & \textcolor{darkgreen}{\textbf{+2.0}} & \textcolor{darkgreen}{\textbf{+5.8}} & \textcolor{darkgreen}{\textbf{+2.1}} &
\textcolor{darkgreen}{\textbf{+2.0}} & \textcolor{darkgreen}{\textbf{+2.4}} & \textcolor{darkgreen}{\textbf{+5.1}} & \textcolor{darkgreen}{\textbf{+1.4}}\\
\bottomrule[1.5pt]
\end{tabular}
}
\end{center}
\vspace{-0.4cm}
\end{table}

\subsection{Complete Results on STS-B-DIR}
We present complete results on STS-B-DIR in Table~\ref{table:appendix-complete-sts-b}, where more metrics, such as MAE and Spearman correlation are added for further evaluation. In summary, across all the metrics used, by adding LDS and FDS we can substantially improve the results, particularly for the medium-shot and few-shot regions.
The advantage is even more profound under \emph{Pearson correlation}, which is commonly used for this task.

\subsection{Complete Results on NYUD2-DIR}
Table~\ref{table:appendix-complete-nyud2} shows the complete evaluation results on NYUD2-DIR. As described before, we further add common metrics for depth estimation evaluation, including $\log_{10}$, $\delta_1$, $\delta_2$, and $\delta_3$. The table reveals the following results. First, either FDS or LDS alone can improve the overall depth regression results, where LDS is more effective for improving performance in the few-shot region.
Furthermore, when combined together, LDS \& FDS can alleviate the overfitting phenomenon to many-shot regions of the vanilla model, and generalize better to all regions.

\begin{table}[t]
\setlength{\tabcolsep}{2.5pt}
\caption{Complete evaluation results on NYUD2-DIR.}
\vspace{-2pt}
\label{table:appendix-complete-nyud2}
\small
\begin{center}
\resizebox{\textwidth}{!}{
\begin{tabular}{l|cccc|cccc|cccc|cccc|cccc}
\toprule[1.5pt]
Metrics      & \multicolumn{4}{c|}{RMSE~$\downarrow$}       &
\multicolumn{4}{c|}{$\log_{10}$~$\downarrow$} & 
\multicolumn{4}{c|}{$\delta_1$~$\uparrow$} & \multicolumn{4}{c|}{$\delta_2$~$\uparrow$} & \multicolumn{4}{c}{$\delta_3$~$\uparrow$}   \\ \midrule
Shot         & All   & Many  & Med. & Few & All   & Many  & Med. & Few & All   &Many   & Med. & Few  & All   & Many  & Med. & Few & All   & Many  & Med. & Few \\ \midrule\midrule
\textsc{Vanilla}      & 1.477 & 0.591 & 0.952  & 2.123 & 0.086 & 0.066 & 0.082& 0.107 & 0.677 & 0.777 &  0.693  & 0.570 & 0.899 & 0.956 & 0.906 & 0.840 & 0.969 & 0.990& 0.975 & 0.946 \\ \midrule\midrule
\textsc{Vanilla} + \textbf{\textsc{LDS}} &  1.387 &  0.671  &    0.913 & 1.954  & 0.086 & 0.079 & 0.079 & 0.097 & 0.672  & 0.701  & 0.706 & 0.630 & 0.907 & 0.932 & 0.929 & 0.875 & 0.976 & 0.984 & 0.982 & 0.964 \\[1.2pt]
\textsc{Vanilla} + \textbf{\textsc{FDS}} & 1.442 & \textbf{0.615} & 0.940  & 2.059  & 0.084 & \textbf{0.069} & 0.080 & 0.101 & 0.681  & \textbf{0.760} & 0.695 & 0.596 & 0.903 & \textbf{0.952} & 0.918 & 0.849 & 0.975 & \textbf{0.989} & 0.976 & 0.960 \\[1.2pt]
\textsc{Vanilla} + \textbf{\textsc{LDS}} + \textbf{\textsc{FDS}} & \textbf{1.338} & 0.670  & \textbf{0.851} & \textbf{1.880}  &  \textbf{0.080} & 0.074 & \textbf{0.070} & \textbf{0.090} & \textbf{0.705} & 0.730  & \textbf{0.764}  & \textbf{0.655} & \textbf{0.916} & 0.939 & \textbf{0.941} & \textbf{0.884} & \textbf{0.979} & 0.984 & \textbf{0.983} & \textbf{0.971} \\ \midrule\midrule
\textsc{\textbf{Ours~(best)} vs. Vanilla}   & \textcolor{darkgreen}{\textbf{+.139}} & \textcolor{lightblue}{\textbf{-.024}} & \textcolor{darkgreen}{\textbf{+.101}} & \textcolor{darkgreen}{\textbf{+.243}} &
\textcolor{darkgreen}{\textbf{+.006}} & \textcolor{lightblue}{\textbf{-.003}} & \textcolor{darkgreen}{\textbf{+.012}} & \textcolor{darkgreen}{\textbf{+.017}} &
\textcolor{darkgreen}{\textbf{+.028}} & \textcolor{lightblue}{\textbf{-.017}} & \textcolor{darkgreen}{\textbf{+.071}} & \textcolor{darkgreen}{\textbf{+.085}} &
\textcolor{darkgreen}{\textbf{+.017}} & \textcolor{lightblue}{\textbf{-.004}} & \textcolor{darkgreen}{\textbf{+.035}} & \textcolor{darkgreen}{\textbf{+.044}} &
\textcolor{darkgreen}{\textbf{+.010}} & \textcolor{lightblue}{\textbf{-.001}} & \textcolor{darkgreen}{\textbf{+.008}} & \textcolor{darkgreen}{\textbf{+.025}}\\
\bottomrule[1.5pt]
\end{tabular}
}
\end{center}
\vspace{-0.5cm}
\end{table}

\subsection{Complete Results on SHHS-DIR}
We report the complete results on SHHS-DIR in Table~\ref{table:appendix-complete-shhs}.
The results again confirm the effectiveness of both LDS and FDS beyond the success on typical image data and text data, as superior performance is demonstrated when applied for real-world imbalanced regression tasks with healthcare data as inputs (i.e., PSG signals). We verify that by combining LDS and FDS, the highest performance gains are established over all tested regions.

\begin{table}[t]
\setlength{\tabcolsep}{7.5pt}
\caption{Complete evaluation results on SHHS-DIR.}
\vspace{-2pt}
\label{table:appendix-complete-shhs}
\small
\begin{center}
\resizebox{\textwidth}{!}{
\begin{tabular}{l|cccc|cccc|cccc}
\toprule[1.5pt]
Metrics      & \multicolumn{4}{c|}{MSE~$\downarrow$}       & \multicolumn{4}{c|}{MAE~$\downarrow$}       & \multicolumn{4}{c}{GM~$\downarrow$}    \\ \midrule
Shot         & All   & Many  & Med. & Few   & All   & Many  & Med. & Few   & All   & Many  & Med. & Few   \\ \midrule\midrule
\textsc{Vanilla}  & 369.18 & 269.37 & 311.45 & 417.31 & 15.36 & 12.47 & 13.98  & 16.94 & 10.63 & 8.04 &  9.59  & 12.20 \\[1.2pt]
\textsc{Vanilla} + \textbf{\textsc{LDS}} & 309.19 & 220.87 & 252.53 & 394.91 & 14.14 & 11.66 & {12.77}  & {16.05} & {9.26} & {7.64} & {8.18}  & {11.32} \\[1.2pt]
\textsc{Vanilla} + \textbf{\textsc{FDS}} & 303.82 & 214.63 & 267.08 & 386.75 & 13.84 & 11.13 & 12.72  & 15.95 & 8.89  & \textbf{6.93}  & 8.05 & 11.19 \\[1.2pt]
\textsc{Vanilla} + \textbf{\textsc{LDS}} + \textbf{\textsc{FDS}} & \textbf{292.18} & \textbf{211.89} & \textbf{247.48} & \textbf{346.01} & \textbf{13.76} & \textbf{11.12} & \textbf{12.18}  & \textbf{15.07}     & \textbf{8.70}     & 6.94     & \textbf{7.60}      & \textbf{10.18} \\ \midrule\midrule
\textsc{Focal-R}  & 345.44 & 219.75 & 309.01 & 430.26 & 14.67 & 11.70  & 13.69  & 17.06 & 9.98 & 7.93 & 8.85 & 11.95 \\[1.2pt]
\textsc{Focal-R} + \textbf{\textsc{LDS}} & 317.39 & 242.18 & 270.04 & 411.73 & 14.49  & 12.01  & 12.43 & 16.57 & 9.98 & 7.89 & 8.59 & 11.40  \\[1.2pt]
\textsc{Focal-R} + \textbf{\textsc{FDS}} & 310.94 & \textbf{185.16} & 303.90 & 391.22 & 14.18 &  \textbf{11.06}  & 13.56 & 15.99 & 9.45 & \textbf{6.95} & 8.81 & 11.13 \\[1.2pt]
\textsc{Focal-R} + \textbf{\textsc{LDS}} + \textbf{\textsc{FDS}} & \textbf{297.85} & 193.42 & \textbf{259.33} & \textbf{375.16} & \textbf{14.02} & 11.08 &  \textbf{12.24} & \textbf{15.49}  & \textbf{9.32} & 7.18  & \textbf{8.10}  & \textbf{10.39}  \\ \midrule\midrule
\textsc{RRT}  & 354.75 & 274.01 & 308.83 & 408.47 & 14.78 & 12.43 & 14.01 & 16.48 & 10.12 & 8.05 & 9.71 & 11.96 \\[1.2pt]
\textsc{RRT} + \textbf{\textsc{LDS}} & 344.18 & 245.39 & 304.32 & 402.56 & {14.56} & 12.08 & {13.44}  & 16.45 & {9.89} & {7.85} & {9.18}  & {11.82} \\[1.2pt]
\textsc{RRT} + \textbf{\textsc{FDS}} & 328.66 & 239.83 & 298.71 & 397.25 & 14.36 & 11.97 & {13.33}  & 16.08 & 9.74 & 7.54 & 9.20  & 11.31 \\[1.2pt]
\textsc{RRT} + \textbf{\textsc{LDS}} + \textbf{\textsc{FDS}} & \textbf{313.58} & \textbf{238.07} & \textbf{276.50} & \textbf{380.64} & \textbf{14.33} & \textbf{11.96} & \textbf{12.47}  & \textbf{15.92}     & \textbf{9.63}     & \textbf{7.35}     & \textbf{8.74} & \textbf{11.17} \\ \midrule\midrule
\textsc{Inv} & 322.17  & 231.68 & 293.43 & 387.48 & 14.39 & 11.84 & 13.12  & 16.02 & 9.34 & 7.73 & 8.49  & 11.20 \\[1.2pt]
\textsc{Inv} + \textbf{\textsc{LDS}} & 309.19 & 220.87 & 252.53 & 394.91 & 14.14 & 11.66 & {12.77}  & {16.05} & {9.26} & {7.64} & {8.18}  & {11.32} \\[1.2pt]
\textsc{Inv} + \textbf{\textsc{FDS}} & 307.95 & 219.36 & 247.55 & 361.29 & 13.91 & \textbf{11.12} & {12.29}  & 15.53 & 8.94  & \textbf{6.91}  & {7.79}      & 10.65 \\[1.2pt]
\textsc{Inv} + \textbf{\textsc{LDS}} + \textbf{\textsc{FDS}} & \textbf{292.18} & \textbf{211.89} & \textbf{247.48} & \textbf{346.01} & \textbf{13.76} & \textbf{11.12} & \textbf{12.18}  & \textbf{15.07}     & \textbf{8.70}     & 6.94     & \textbf{7.60}      & \textbf{10.18} \\ \midrule\midrule
\textsc{\textbf{Ours~(best)} vs. Vanilla} & \textcolor{darkgreen}{\textbf{+77.00}} & \textcolor{darkgreen}{\textbf{+84.21}} & \textcolor{darkgreen}{\textbf{+63.97}} & \textcolor{darkgreen}{\textbf{+71.30}} & \textcolor{darkgreen}{\textbf{+1.60}} & \textcolor{darkgreen}{\textbf{+1.41}} & \textcolor{darkgreen}{\textbf{+1.80}} & \textcolor{darkgreen}{\textbf{+1.87}} & \textcolor{darkgreen}{\textbf{+1.93}} & \textcolor{darkgreen}{\textbf{+1.13}} & \textcolor{darkgreen}{\textbf{+1.99}} & \textcolor{darkgreen}{\textbf{+2.02}} \\
\bottomrule[1.5pt]
\end{tabular}
}
\end{center}
\vspace{-0.5cm}
\end{table}

\section{Further Analysis and Ablation Studies}
\label{appendix:ablation-visualization}

\subsection{Kernel Type for LDS \& FDS}
\label{appendix:ablation-kernel-type}

We study the effects of different kernel types for LDS and FDS when applying distribution smoothing, in addition to the default setting where Gaussian kernels are employed.
We select three different kernel types, i.e., Gaussian, Laplacian, and Triangular kernel, and evaluate their effects on both LDS and FDS.
We remain other hyper-parameters unchanged as in Sec.~\ref{appendix:experiment-details-implementation}, and report results on IMDB-WIKI-DIR in Table~\ref{table:appendix-imdb-kernel} and results on STS-B-DIR in Table~\ref{table:appendix-sts-kernel}.
In general, as both tables indicate, all kernel types can lead to notable gains compared to the vanilla model. Moreover, Gaussian kernel often delivers the best results among all kernel types, which is consistent for both LDS and FDS.

\begin{table}[tb]
\setlength{\tabcolsep}{7.5pt}
\caption{Ablation study of different kernel types for LDS \& FDS on IMDB-WIKI-DIR.}
\vspace{4pt}
\label{table:appendix-imdb-kernel}
\begin{center}
\small
\resizebox{0.95\textwidth}{!}{
\begin{tabular}{l|cccc|cccc|cccc}
\toprule[1.5pt]
Metrics      &\multicolumn{4}{c|}{MSE~$\downarrow$} & \multicolumn{4}{c|}{MAE~$\downarrow$}       & \multicolumn{4}{c}{GM~$\downarrow$}    \\ \midrule
Shot         & All   & Many  & Med. & Few & All   & Many  & Med. & Few   & All   & Many  & Med. & Few   \\ \midrule\midrule
\textsc{Vanilla}   & 138.06 &  108.70  & 366.09  & 964.92  & 8.06 & 7.23 & 15.12  & 26.33 & 4.57 & 4.17 & 10.59  & 20.46 \\ \midrule\midrule
\multicolumn{9}{l}{\emph{\textbf{LDS:}}} \\ \midrule
\textsc{Gaussian Kernel}   & 131.65 & 109.04 & 298.98 & 834.08  & {7.83} & 7.31 & {12.43}  & {22.51} & {4.42} & {4.19} & {7.00}  & {13.94} \\[1.2pt]
\textsc{Triangular Kernel} & 133.77 & 110.24  & 309.70 & 850.74 & 7.89 & 7.30 & 12.72 & 22.80 & 4.50 & 4.24 & 7.75 & 14.91  \\[1.2pt]
\textsc{Laplacian Kernel} & 132.87 &  109.27  & 312.10 & 829.83 & 7.87 & 7.29 & 12.68 & 22.38 & 4.50 & 4.26 & 7.29 & 13.71\\ \midrule\midrule
\multicolumn{9}{l}{\emph{\textbf{FDS:}}} \\ \midrule
\textsc{Gaussian Kernel}      & 133.81 & 107.51 & 332.90 & 916.18 & 7.85 & 7.18 & 13.35 & 24.12 & 4.47 & 4.18 & 8.18 & 15.18 \\[1.2pt]
\textsc{Triangular Kernel} & 134.09 & 110.49 & 301.18 & 927.99 & 7.97 & 7.41 & 12.20 & 23.99 & 4.64 & 4.41 & 7.06 & 14.28  \\[1.2pt]
\textsc{Laplacian Kernel} & 133.00 & 104.26 & 352.95 & 968.62 & 8.05 & 7.25 & 14.78 & 26.16 & 4.71 & 4.33 & 10.19 & 19.09 \\
\bottomrule[1.5pt]
\end{tabular}
}
\end{center}
\vspace{-0.4cm}
\end{table}

\begin{table}[tb]
\setlength{\tabcolsep}{4pt}
\caption{Ablation study of different kernel types for LDS \& FDS on STS-B-DIR.}
\vspace{4pt}
\label{table:appendix-sts-kernel}
\begin{center}
\small
\resizebox{0.95\textwidth}{!}{
\begin{tabular}{l|cccc|cccc|cccc|cccc}
\toprule[1.5pt]
Metrics      &\multicolumn{4}{c|}{MSE~$\downarrow$} & \multicolumn{4}{c|}{MAE~$\downarrow$}       & \multicolumn{4}{c|}{Pearson correlation (\%)~$\uparrow$} & \multicolumn{4}{c}{Spearman correlation (\%)~$\uparrow$}   \\ \midrule
Shot         & All   & Many  & Med. & Few & All   & Many  & Med. & Few    & All   & Many  & Med. & Few & All   & Many  & Med. & Few   \\ \midrule\midrule
\textsc{Vanilla}   & 0.974 & 0.851 & 1.520 & 0.984 & 0.794 & 0.740 & 1.043 & 0.771 & 74.2 & 72.0 & 62.7 & 75.2 & 74.4 & 68.8 & 50.5 & 75.0 \\ \midrule\midrule
\multicolumn{9}{l}{\emph{\textbf{LDS:}}} \\ \midrule
\textsc{Gaussian Kernel}   & 0.914 & 0.819 & 1.319 & 0.955 & 0.773 & 0.729 & 0.970 & 0.772  & 75.6 & 73.4 & 63.8 & 76.2 & 76.1 &  70.4 & 55.6 & 74.3 \\[1.2pt]
\textsc{Triangular Kernel} & 0.938 & 0.870 & 1.193 & 1.039 & 0.786 & 0.754 & 0.929 & 0.784  & 74.8 & 72.4 & 64.1 & 74.0 & 75.2 & 69.3 & 54.1 & 73.9  \\[1.2pt]
\textsc{Laplacian Kernel} & 0.938 & 0.829 & 1.413 & 0.962 & 0.782 & 0.731 & 1.014 & 0.773 & 75.7 & 73.0 & 65.8 & 76.5 & 76.0 & 70.0 & 52.3 & 75.2 \\ \midrule\midrule
\multicolumn{9}{l}{\emph{\textbf{FDS:}}} \\ \midrule
\textsc{Gaussian Kernel}      & 0.916 & 0.875 & 1.027 & 1.086 & 0.767 & 0.746 & 0.840 & 0.811  & 75.5 & 73.0 & 67.0 & 72.8 & 75.8 & 69.9 & 54.4 & 72.0  \\[1.2pt]
\textsc{Triangular Kernel} & 0.935 & 0.863 & 1.239 & 0.966 & 0.762 & 0.725 & 0.912 & 0.788 & 74.6 & 72.4 & 64.8 & 75.9 & 74.4 & 69.1 & 48.4 & 75.4  \\[1.2pt]
\textsc{Laplacian Kernel} & 0.925 & 0.843 & 1.247 & 1.020 & 0.771 & 0.733 & 0.929 & 0.800 & 75.0 & 72.6 & 64.7 & 74.2 & 75.4 & 70.1 & 53.5 & 73.5  \\
\bottomrule[1.5pt]
\end{tabular}
}
\end{center}
\vspace{-0.4cm}
\end{table}

\subsection{Training Loss for LDS \& FDS}
\label{appendix:ablation-diff-training-loss}

In the main paper, we fix the training loss function used for each dataset (e.g., MSE loss is used for experiments on STS-B-DIR). In this section, we investigate the influence of different training loss functions on LDS \& FDS. We select three common losses used for regression tasks, i.e., $L_1$ loss, MSE loss, and the Huber loss~(also referred to as smoothed $L_1$ loss). We show the results on STS-B-DIR in Table~\ref{table:appendix-loss-sts}, where similar results are obtained for all the losses, with no significant performance differences observed between loss functions, indicating that FDS \& LDS are robust to different loss functions.

\vspace{-0.4cm}
\begin{table}[H]
\setlength{\tabcolsep}{5pt}
\caption{Ablation study of different loss functions used during training for LDS \& FDS on STS-B-DIR.}
\vspace{4pt}
\label{table:appendix-loss-sts}
\begin{center}
\small
\resizebox{0.95\textwidth}{!}{
\begin{tabular}{l|cccc|cccc|cccc|cccc}
\toprule[1.5pt]
Metrics      &\multicolumn{4}{c|}{MSE~$\downarrow$} & \multicolumn{4}{c|}{MAE~$\downarrow$}       &  \multicolumn{4}{c|}{Pearson correlation (\%)~$\uparrow$} & \multicolumn{4}{c}{Spearman correlation (\%)~$\uparrow$}   \\ \midrule
Shot         & All   & Many  & Med. & Few   & All   & Many  & Med. & Few & All   & Many  & Med. & Few & All   & Many  & Med. & Few   \\ \midrule\midrule
\multicolumn{9}{l}{\emph{\textbf{LDS:}}} \\ \midrule
\textsc{L1}   & 0.893 & 0.808 & 1.241 & 0.964 & 0.765 & 0.727 & 0.938 & 0.758 & 76.3 & 73.9 & 66.0 & 75.9 & 76.7 & 71.1 & 54.5 & 75.6 \\[1.2pt]
\textsc{MSE} & 0.914 & 0.819 & 1.319 & 0.955 & 0.773 & 0.729 & 0.970 & 0.772  & 75.6 & 73.4 & 63.8 & 76.2 & 76.1 &  70.4 & 55.6 & 74.3  \\[1.2pt]
\textsc{Huber Loss} & 0.902 & 0.811 & 1.276 & 0.978 & 0.761 & 0.718 & 0.954 & 0.751 & 76.1 & 74.2 & 64.7 & 75.5 & 76.5 & 71.6 & 52.9 & 74.3 \\ \midrule\midrule
\multicolumn{9}{l}{\emph{\textbf{FDS:}}} \\ \midrule
\textsc{L1}    & 0.918 & 0.860 & 1.105 & 1.082 & 0.762 & 0.733 & 0.859 & 0.833 & 75.5 & 73.7 & 65.3 & 72.3 & 75.6 & 70.9 & 52.1 & 71.5    \\[1.2pt]
\textsc{MSE}  & 0.916 & 0.875 & 1.027 & 1.086 & 0.767 & 0.746 & 0.840 & 0.811  & 75.5 & 73.0 & 67.0 & 72.8 & 75.8 & 69.9 & 54.4 & 72.0  \\[1.2pt]
\textsc{Huber Loss} & 0.920 & 0.867 & 1.097 & 1.052 & 0.765 & 0.741 & 0.858 & 0.800 & 75.3 & 72.9 & 66.6 & 73.6 & 75.3 & 69.7 & 52.3 & 73.6  \\
\bottomrule[1.5pt]
\end{tabular}
}
\end{center}
\vspace{-0.4cm}
\end{table}

\subsection{Hyper-parameters for LDS \& FDS}
\label{appendix:ablation-hyper-param}

In this section, we study the effects of different hyper-parameters on both LDS and FDS. As we mainly employ the Gaussian kernel for distribution smoothing, we extensively study different choices of the kernel size $l$ and the standard deviation $\sigma$.
Specifically, we conduct controlled experiments on IMDB-WIKI-DIR and STS-B-DIR, where we vary the choices of these hyper-parameters as $l\in\{\texttt{5,9,15}\}$ and $\sigma\in\{\texttt{1,2,3}\}$, and leave other training hyper-parameters unchanged.

\begin{table}[tb]
\setlength{\tabcolsep}{7.5pt}
\caption{Hyper-parameter study on kernel size $l$ and standard deviation $\sigma$ for LDS \& FDS on IMDB-WIKI-DIR.}
\vspace{4pt}
\label{table:appendix-hyper-imdb}
\begin{center}
\small
\resizebox{0.85\textwidth}{!}{
\begin{tabular}{c|c|cccc|cccc|cccc}
\toprule[1.5pt]
\multicolumn{2}{l|}{Metrics}      &\multicolumn{4}{c|}{MSE~$\downarrow$} & \multicolumn{4}{c|}{MAE~$\downarrow$}       & \multicolumn{4}{c}{GM~$\downarrow$}    \\ \midrule
\multicolumn{2}{l|}{Shot}         & All   & Many  & Med. & Few & All   & Many  & Med. & Few   & All   & Many  & Med. & Few   \\ \midrule\midrule
\multicolumn{2}{l|}{\textsc{Vanilla}}   & 138.06 &  108.70  & 366.09  & 964.92  & 8.06 & 7.23 & 15.12  & 26.33 & 4.57 & 4.17 & 10.59  & 20.46 \\ \midrule\midrule
$l$ & $\sigma$ &  \\ \midrule\midrule
\multicolumn{9}{l}{\emph{\textbf{LDS:}}} \\ \midrule
\texttt{5}&\texttt{1}   & 132.08 & 108.53 & 309.03 & 843.53   & 7.80 & 7.22 & 12.61 & 22.33 &4.42 & 4.19 & 7.16 & 12.54 \\[1.2pt]
\texttt{9}&\texttt{1} & 135.04 & 112.32 & 307.90 & 803.15   & 7.97 & 7.39 & 12.74 & 22.19 & 4.55 & 4.30 & 7.53 & 14.11  \\[1.2pt]
\texttt{15}&\texttt{1} & 134.06 & 110.49 & 308.83 &  864.30 & 7.84 & 7.28 & 12.35 & 22.81 & 4.44 & 4.22 & 6.95 & 14.22\\[1.2pt] 
\texttt{5}&\texttt{2}   & 131.65 & 109.04 & 298.98 & 834.08  & 7.83 & 7.31 & 12.43  & 22.51 & 4.42 & 4.19 & 7.00  & 13.94 \\[1.2pt]
\texttt{9}&\texttt{2} & 136.78 & 112.41 & 322.65 & 850.47   & 8.02 & 7.41 & 13.00 & 23.23 & 4.55 & 4.29 & 7.55 & 15.65  \\[1.2pt]
\texttt{15}&\texttt{2} & 135.66 & 111.68 & 319.20 & 833.02   & 7.98 & 7.40 & 12.74 & 22.27 & 4.60 & 4.37 & 7.30 & 12.92\\[1.2pt] 
\texttt{5}&\texttt{3}   & 137.56 & 113.50 & 322.47 & 831.38 & 8.07 & 7.47 & 13.06 & 22.85 & 4.63 & 4.36 & 7.87 & 15.11 \\[1.2pt]
\texttt{9}&\texttt{3} & 138.91 & 114.89 & 319.40 & 863.16   & 8.18 & 7.57 & 13.19 & 23.33 & 4.71 & 4.44 & 8.09 & 15.17  \\[1.2pt]
\texttt{15}&\texttt{3} & 138.86 & 114.25 & 326.97 & 856.27   & 8.18 & 7.54 & 13.53 & 23.17 & 4.77 & 4.47 & 8.52 & 15.25\\ \midrule\midrule
\multicolumn{9}{l}{\emph{\textbf{FDS:}}} \\ \midrule
\texttt{5}&\texttt{1}   & 133.63 & 104.80 & 354.24 & 972.54 & 7.87 & 7.06 & 14.71 & 25.96 & 4.42 & 4.04 & 9.95 & 18.47 \\[1.2pt]
\texttt{9}&\texttt{1} & 134.34 & 105.97 & 356.54 & 919.16 & 7.95 & 7.18 & 14.58 & 24.80 & 4.54 & 4.20 & 9.56 & 15.13  \\[1.2pt]
\texttt{15}&\texttt{1} & 136.32 & 107.47 & 355.84 & 948.71 & 7.97 & 7.23 & 14.81 & 25.59 & 4.60 & 4.23 & 9.99 & 17.60 \\[1.2pt]
\texttt{5}&\texttt{2}   & 133.81 & 107.51 & 332.90 & 916.18 & 7.85 & 7.18 & 13.35 & 24.12 & 4.47 & 4.18 & 8.18 & 15.18 \\[1.2pt]
\texttt{9}&\texttt{2} & 133.99 & 105.01 & 357.31 & 963.79 & 7.94 & 7.11 & 14.95 & 25.97 & 4.48 & 4.09 & 10.49 & 18.19  \\[1.2pt]
\texttt{15}&\texttt{2} & 136.61 & 107.93 & 361.08 & 973.56 & 7.98 & 7.23 & 14.68 & 25.21 & 4.61 & 4.24 & 10.14 & 17.91 \\[1.2pt] 
\texttt{5}&\texttt{3}   & 136.81 & 107.76 & 359.08 & 953.16 & 7.98 & 7.18 & 14.85 & 24.94 & 4.53 & 4.15 & 10.27 & 17.33 \\[1.2pt]
\texttt{9}&\texttt{3} & 133.48 & 104.14 & 359.80 & 972.29 & 7.94 & 7.09 & 15.04 & 25.87 & 4.48 & 4.09 & 10.40 & 16.85  \\[1.2pt]
\texttt{15}&\texttt{3} & 132.55 & 103.08 & 360.39 & 970.43 & 8.03 & 7.22 & 14.86 & 25.40 & 4.67 & 4.33 & 10.04 & 13.86 \\
\bottomrule[1.5pt]
\end{tabular}
}
\end{center}
\vspace{-0.6cm}
\end{table}

\begin{table}[H]
\setlength{\tabcolsep}{6pt}
\caption{Hyper-parameter study on kernel size $l$ and standard deviation $\sigma$ for LDS \& FDS on STS-B-DIR.}
\vspace{4pt}
\label{table:appendix-hyper-sts}
\begin{center}
\small
\resizebox{0.9\textwidth}{!}{
\begin{tabular}{c|c|cccc|cccc|cccc|cccc}
\toprule[1.5pt]
\multicolumn{2}{l|}{Metrics}      &\multicolumn{4}{c|}{MSE~$\downarrow$} & \multicolumn{4}{c|}{MAE~$\downarrow$}       & \multicolumn{4}{c|}{Pearson correlation (\%)~$\uparrow$} & \multicolumn{4}{c}{Spearman correlation (\%)~$\uparrow$}   \\ \midrule
\multicolumn{2}{l|}{Shot}   & All   & Many  & Med. & Few   & All   & Many  & Med. & Few & All   & Many  & Med. & Few & All   & Many  & Med. & Few   \\ \midrule\midrule
\multicolumn{2}{l|}{\textsc{Vanilla}}   & 0.974 & 0.851 & 1.520 & 0.984 & 0.794 & 0.740 & 1.043 & 0.771  & 74.2 & 72.0 & 62.7 & 75.2 & 74.4 & 68.8 & 50.5 & 75.0 \\ \midrule\midrule
$l$ & $\sigma$ &  \\ \midrule\midrule
\multicolumn{9}{l}{\emph{\textbf{LDS:}}} \\ \midrule
\texttt{5}&\texttt{1}   & 0.942 & 0.825 & 1.431 & 1.023 & 0.781 & 0.726 & 1.016 & 0.809 & 75.1 & 73.2 & 61.8 & 74.5 & 75.3 & 70.2 & 52.2 & 72.5 \\[1.2pt]
\texttt{9}&\texttt{1} & 0.931 & 0.840 & 1.323 & 0.962 & 0.785 & 0.744 & 0.972 & 0.773 & 75.0 & 72.7 & 63.3 & 75.8 & 75.6 & 70.1 & 53.6 & 74.8  \\[1.2pt]
\texttt{15}&\texttt{1} & 0.941 & 0.833 & 1.413 & 0.953 & 0.781 & 0.728 & 1.014 & 0.776 & 75.0 & 72.8 & 62.6 & 76.3 & 75.5 & 70.2 & 52.0 & 74.6 \\[1.2pt] 
\texttt{5}&\texttt{2}   & 0.914 & 0.819 & 1.319 & 0.955 & 0.773 & 0.729 & 0.970 & 0.772 & 75.6 & 73.4 & 63.8 & 76.2 & 76.1 &  70.4 & 55.6 & 74.3 \\[1.2pt]
\texttt{9}&\texttt{2} & 0.926 & 0.823 & 1.379 & 0.944 & 0.782 & 0.733 & 1.003 & 0.764 & 75.5 & 73.4 & 63.6 & 76.8 & 76.0 & 70.5 & 53.5 & 76.2  \\[1.2pt]
\texttt{15}&\texttt{2} & 0.949 & 0.831 & 1.452 & 1.005 & 0.788 & 0.735 & 1.023 & 0.782 & 74.9 & 72.9 & 63.0 & 74.7 & 75.4 & 70.1 & 52.5 & 73.6 \\[1.2pt]
\texttt{5}&\texttt{3}   & 0.928 & 0.845 & 1.250 & 1.041 & 0.775 & 0.733 & 0.951 & 0.798 & 75.1 & 73.3 & 63.2 & 73.8 & 75.3 & 70.4 & 51.4 & 72.6 \\[1.2pt]
\texttt{9}&\texttt{3} & 0.939 & 0.816 & 1.462 & 1.000 & 0.786 & 0.732 & 1.030 & 0.783 & 75.3 & 73.5 & 62.6 & 74.7 & 75.9 & 70.9 & 53.0 & 73.7  \\[1.2pt]
\texttt{15}&\texttt{3} & 0.927 & 0.824 & 1.348 & 1.010 & 0.774 & 0.726 & 0.982 & 0.780 & 75.2 & 73.4 & 62.2 & 74.6 & 75.7 & 70.7 & 53.0 & 72.3 \\ \midrule\midrule
\multicolumn{9}{l}{\emph{\textbf{FDS:}}} \\ \midrule
\texttt{5}&\texttt{1}   & 0.943 & 0.869 & 1.217 & 1.066 & 0.776 & 0.742 & 0.914 & 0.799 & 74.4 & 71.7 & 65.6 & 72.5 & 74.2 & 68.4 & 51.1 & 71.2 \\[1.2pt]
\texttt{9}&\texttt{1} & 0.927 & 0.851 & 1.193 & 1.096 & 0.770 & 0.736 & 0.896 & 0.822 & 74.9 & 72.8 & 65.8 & 71.6 & 74.8 & 69.7 & 52.3 & 68.3  \\[1.2pt]
\texttt{15}&\texttt{1} & 0.926 & 0.854 & 1.202 & 1.029 & 0.776 & 0.743 & 0.914 & 0.800 & 74.9 & 72.6 & 66.1 & 74.0 & 75.1 & 69.8 & 49.5 & 73.6 \\[1.2pt] 
\texttt{5}&\texttt{2}   & 0.916 & 0.875 & 1.027 & 1.086 & 0.767 & 0.746 & 0.840 & 0.811 & 75.5 & 73.0 & 67.0 & 72.8 & 75.8 & 69.9 & 54.4 & 72.0 \\[1.2pt]
\texttt{9}&\texttt{2} & 0.933 & 0.888 & 1.068 & 1.081 & 0.776 & 0.752 & 0.855 & 0.839 & 74.8 & 72.0 & 67.9 & 72.2 & 74.9 & 68.9 & 53.3 & 72.0  \\[1.2pt]
\texttt{15}&\texttt{2} & 0.944 & 0.890 & 1.125 & 1.078 & 0.783 & 0.761 & 0.864 & 0.822 & 74.4 & 71.8 & 65.8 & 72.2 & 74.5 & 68.9 & 53.1 & 70.9 \\[1.2pt]
\texttt{5}&\texttt{3}   & 0.924 & 0.860 & 1.190 & 0.964 & 0.771 & 0.740 & 0.897 & 0.790 & 75.0 & 72.7 & 64.4 & 76.1 & 75.1 & 69.4 & 53.8 & 76.5 \\[1.2pt]
\texttt{9}&\texttt{3} & 0.932 & 0.878 & 1.149 & 0.982 & 0.770 & 0.746 & 0.876 & 0.780 & 74.8 & 72.5 & 63.8 & 75.3 & 74.8 & 69.3 & 50.2 & 75.6  \\[1.2pt]
\texttt{15}&\texttt{3} & 0.956 & 0.915 & 1.110 & 1.016 & 0.784 & 0.767 & 0.855 & 0.803 & 74.4 & 72.1 & 63.7 & 75.5 & 74.3 & 68.7 & 50.0 & 74.6 \\
\bottomrule[1.5pt]
\end{tabular}
}
\end{center}
\vspace{-0.3cm}
\end{table}

\textbf{IMDB-WIKI-DIR.}
We first report the results on IMDB-WIKI-DIR in Table~\ref{table:appendix-hyper-imdb}.
The table reveals the following observations. First, both LDS and FDS are robust to different hyper-parameters within the given range, where similar performance gains are obtained across different choices of $\{l, \sigma\}$. Specifically, for LDS, the relative MAE improvements in the few-shot regions range from $11.4\%$ to $15.7\%$, where a smaller $\sigma$ usually leads to slightly better results over all regions. As for FDS, similar conclusion can be made, while a smaller $l$ often obtains slightly higher improvements.
Interestingly, we can also observe that LDS leads to larger gains w.r.t. the performance in medium-shot and few-shot regions, while with minor degradation in many-shot regions. In contrast, FDS equally boosts all the regions, with slightly smaller improvements in medium-shot and few-shot regions compared to LDS. Finally, for both LDS and FDS, setting $l=\texttt{5}$ and $\sigma=\texttt{2}$ exhibits the best results.

\textbf{STS-B-DIR.}
Further, we show the results of different hyper-parameters on STS-B-DIR in Table~\ref{table:appendix-hyper-sts}. Similar to the results on IMDB-WIKI-DIR, we observe that both LDS and FDS are robust to the hyper-parameter changes, where the performance gaps between $\{l, \sigma\}$ pairs become smaller. In summary, the overall MSE gains range from $3.3\%$ to $6.2\%$ compared to the vanilla model, with $l=\texttt{5}$ and $\sigma=\texttt{2}$ exhibiting the best results for both LDS and FDS.

\subsection{Robustness to Diverse Skewed Label Distributions}
\label{appendix:ablation-diff-skewed-label-density}

\begin{figure*}[ht]
\centering
\subfigure[Curated skewed label distribution, with 1 Gaussian peak]{
    \label{fig:skewed_peak_1}
    \includegraphics[height=0.295\textwidth]{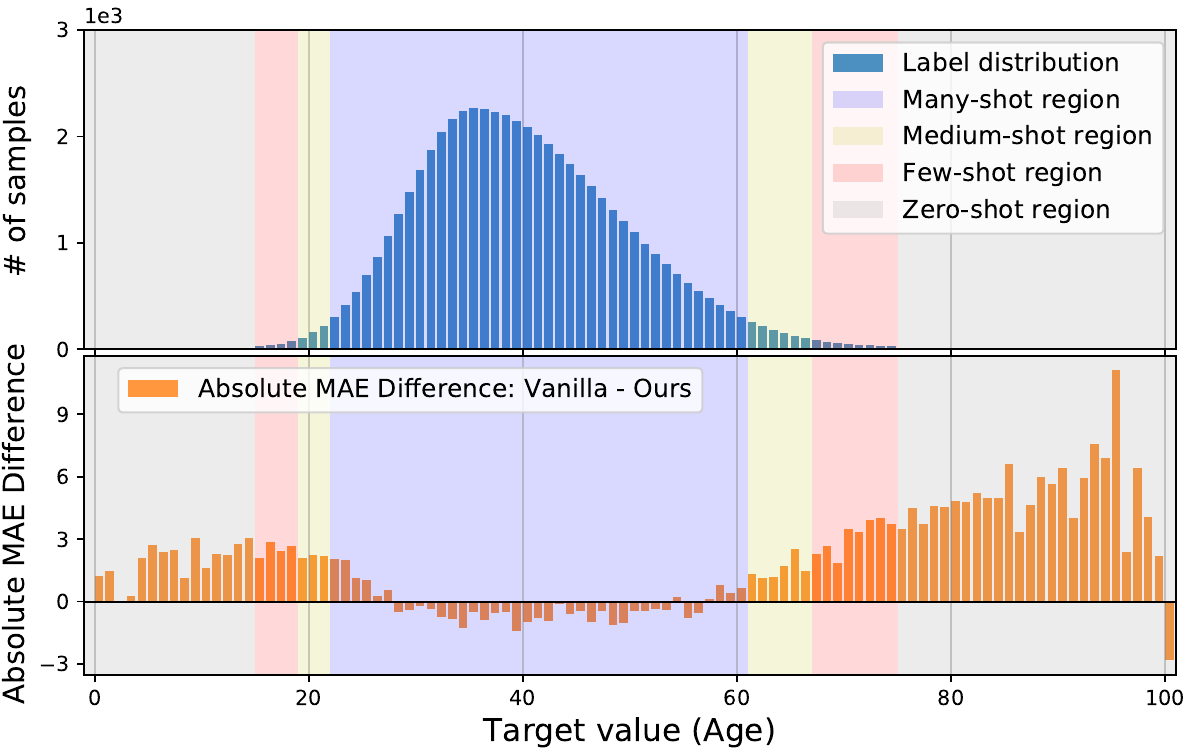}
}
\hspace{1.5ex}
\subfigure[Curated skewed label distribution, with 2 Gaussian peaks]{
    \label{fig:skewed_peak_2}
    \includegraphics[height=0.295\textwidth]{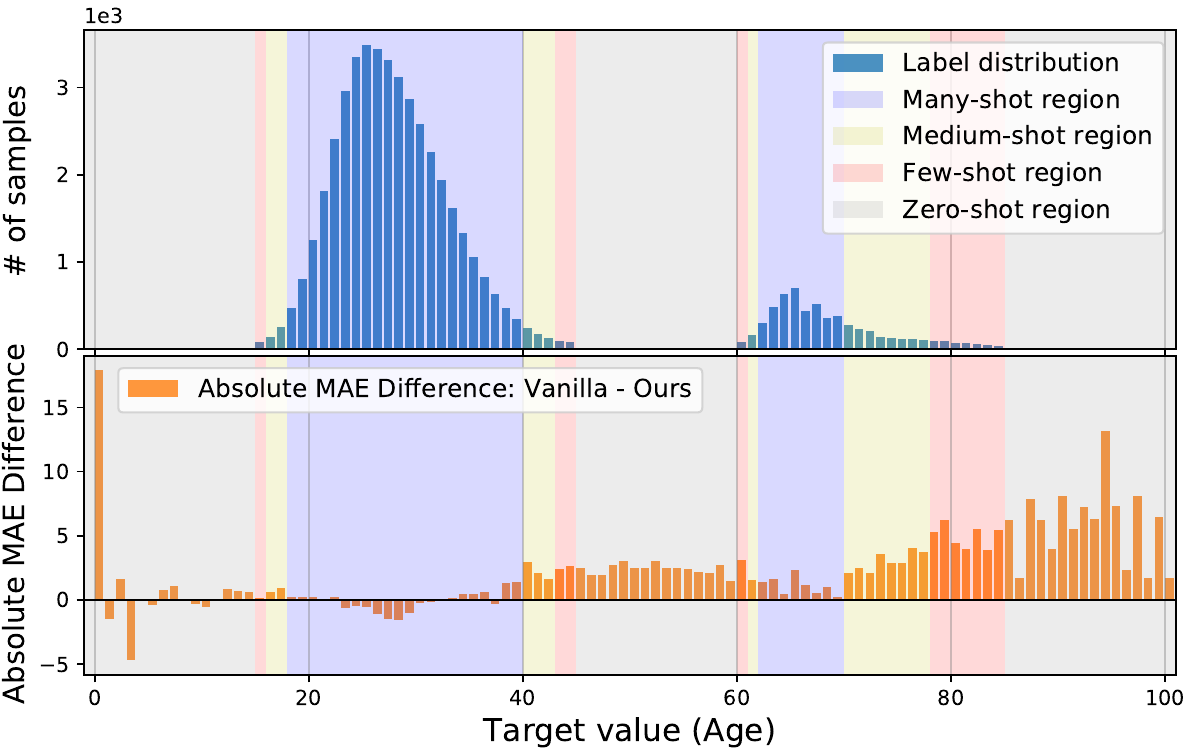}
}\\ \vspace{-0.1cm}
\subfigure[Curated skewed label distribution, with 3 Gaussian peaks]{
    \label{fig:skewed_peak_3}
    \includegraphics[height=0.295\textwidth]{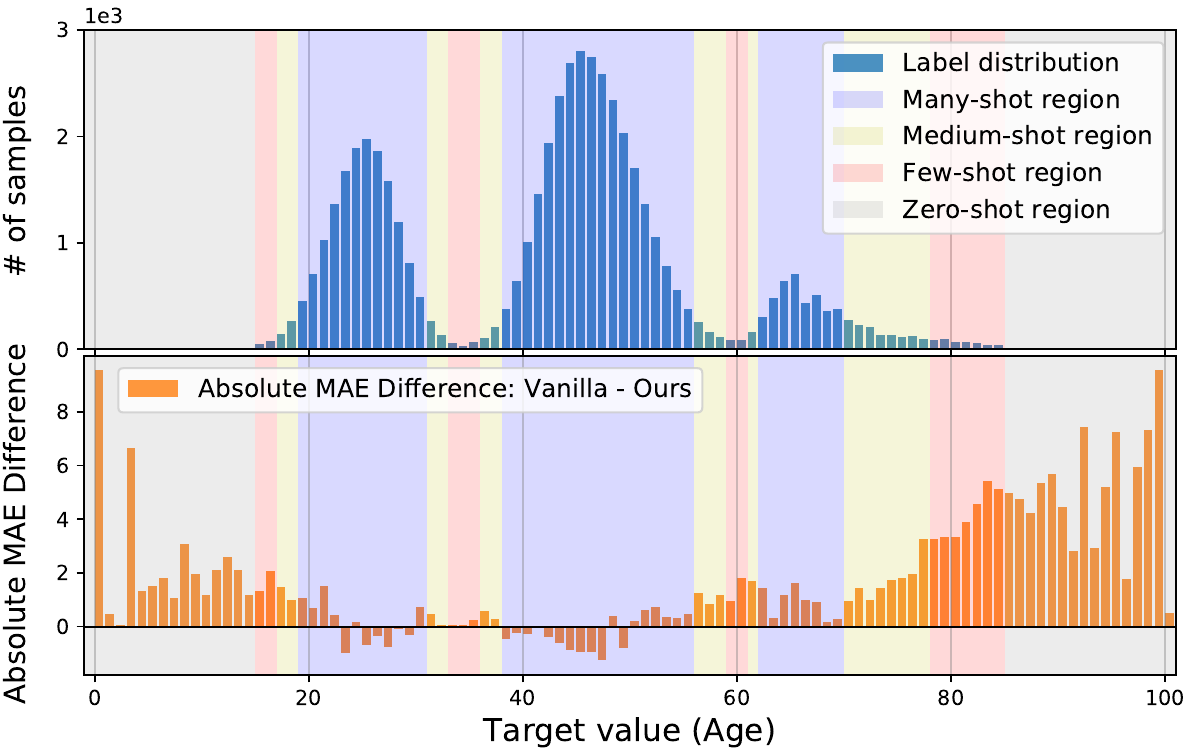}
}
\hspace{1.5ex}
\subfigure[Curated skewed label distribution, with 4 Gaussian peaks]{
    \label{fig:skewed_peak_4}
    \includegraphics[height=0.295\textwidth]{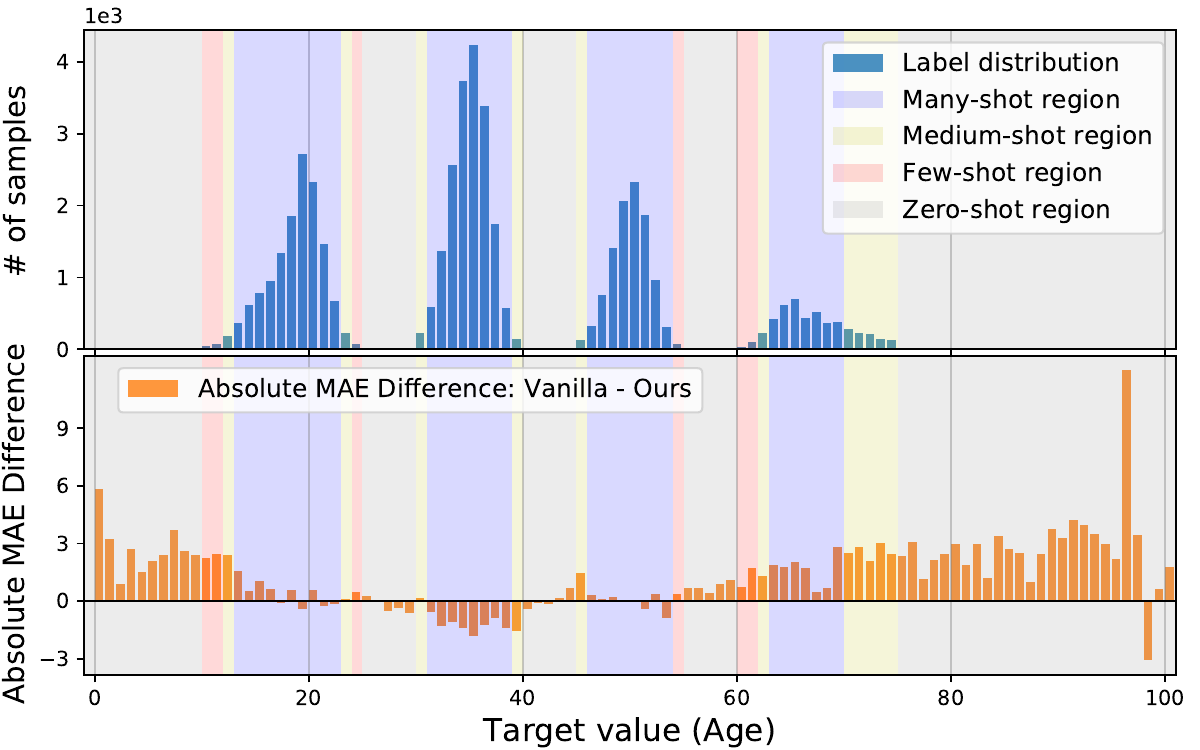}
}
\vspace{-0.2cm}
\caption{The absolute MAE gains of LDS $+$ FDS over the vanilla model under different skewed label distributions. We curate different imbalanced label distributions on IMDB-WIKI-DIR using different number of skewed Gaussians over the target space. We confirm that LDS and FDS are robust to distribution change, and can consistently bring improvements under different imbalanced label distributions.
}
\label{fig:appendix-diff-skewed-dist}
\vspace{-0.1cm}
\end{figure*}

We analyze the effects of different skewed label distributions on our techniques for DIR tasks.
We curate different imbalanced label distributions for IMDB-WIKI-DIR by combining different number of skewed Gaussians over the target space.
Precisely, as shown in Fig.~\ref{fig:appendix-diff-skewed-dist}, we create new training sets with $\{\texttt{1,2,3,4}\}$ disjoint skewed Gaussian distributions over the label space, with potential missing data in certain target regions, and evaluate the robustness of LDS and FDS to the distribution change.

\vspace{-0.3cm}

We verify in Table~\ref{table:appendix-skewed-dist} that even under different imbalanced label distributions, LDS and FDS consistently bring improvements compared to the vanilla model. Substantial improvements are established not only on regions that have data, but more prominent on those without data, i.e., zero-shot regions that require target interpolation or extrapolation.
We further visualize the absolute MAE gains of our methods over the vanilla model for the curated skewed distributions in Fig.~\ref{fig:appendix-diff-skewed-dist}. Our methods provide a comprehensive treatment to the many, medium, few, as well as zero-shot regions, where remarkable performance gains are achieved across all skewed distributions, confirming the robustness of LDS and FDS under distribution change.

\begin{table}[t]
\setlength{\tabcolsep}{4.5pt}
\caption{Ablation study on different skewed label distributions on IMDB-WIKI-DIR.}
\vspace{4pt}
\label{table:appendix-skewed-dist}
\small
\begin{center}
\resizebox{0.95\textwidth}{!}{
\begin{tabular}{l|ccccccc|ccccccc}
\toprule[1.5pt]
Metrics      & \multicolumn{7}{c|}{MAE~$\downarrow$}       & \multicolumn{7}{c}{GM~$\downarrow$}    \\ \midrule
Shot         & All   & Many & Med. & Few & Zero & Interp. & Extrap.   & All   & Many & Med. & Few & Zero & Interp. & Extrap.   \\ \midrule\midrule
\multicolumn{9}{l}{\emph{\textbf{1 peak:}}} \\ \midrule
\textsc{Vanilla}       & 11.20 & 6.05 & 11.43 & 14.76 & 22.67 & $-$ & 22.67 & 7.02 & \textbf{3.84} & 8.67 & 12.26 & 21.07 & $-$ & 21.07 \\ [1.2pt]
\textsc{Vanilla} + \textbf{\textsc{LDS}}  & 10.09 & 6.26 & 9.91 & 12.12 & 19.37 & $-$ & 19.37 & 6.14 & 3.92 & 6.50 & 8.30 & 16.35 & $-$ & 16.35 \\[1.2pt]
\textsc{Vanilla} + \textbf{\textsc{FDS}}  & 11.04 & \textbf{5.97} & 11.19 & 14.54 & 22.35 & $-$ & 22.35 & 6.96 & \textbf{3.84} & 8.54 & 12.08 & 20.71 & $-$ & 20.71 \\[1.2pt]
\textsc{Vanilla} + \textbf{\textsc{LDS}} + \textbf{\textsc{FDS}}  & \textbf{10.00} & 6.28 & \textbf{9.66} & \textbf{11.83} & \textbf{19.21} & $-$ & \textbf{19.21} & \textbf{6.09} & 3.96 & \textbf{6.26} & \textbf{8.14} & \textbf{15.89} & $-$ & \textbf{15.89}  \\ \midrule\midrule
\multicolumn{9}{l}{\emph{\textbf{2 peaks:}}} \\ \midrule
\textsc{Vanilla}      & 11.72 & 6.83 & 11.78 & 15.35 & 16.86 & 16.13  & 18.19 & 7.44 & 3.61 & 8.06 & 12.94 &15.21&  14.41  & 16.74 \\ [1.2pt]
\textsc{Vanilla} + \textbf{\textsc{LDS}} &  10.54  & 6.72 & 9.65 & 12.60 & 15.30&   14.14 & 17.38  & 6.50    & 3.65 & \textbf{5.65} & 9.30 & 13.20 & 12.13 & 15.36 \\[1.2pt]
\textsc{Vanilla} + \textbf{\textsc{FDS}} & 11.40  & 6.69 & 11.02 & 14.85 & 16.61 & 15.83  & 18.01  & 7.18   & \textbf{3.50} & 7.49 & 12.73 & 14.86 & 14.02 & 16.48 \\[1.2pt]
\textsc{Vanilla} + \textbf{\textsc{LDS}} + \textbf{\textsc{FDS}} & \textbf{10.27}  & \textbf{6.61} & \textbf{9.46} & \textbf{11.96} & \textbf{14.89} & \textbf{13.71} & \textbf{17.02}  & \textbf{6.33}   & 3.54 & 5.68 & \textbf{8.80} & \textbf{12.83} & \textbf{11.71}  & \textbf{15.13}  \\ \midrule\midrule
\multicolumn{9}{l}{\emph{\textbf{3 peaks:}}} \\ \midrule
\textsc{Vanilla}      & 9.83  & 7.01 & 9.81 & 11.93 & 20.11 & $-$ & 20.11 & 6.04 & 3.93 & 6.94 & 9.84 & 17.77 & $-$ & 17.77 \\ [1.2pt]
\textsc{Vanilla} + \textbf{\textsc{LDS}} & 9.08 & \textbf{6.77} & 8.82 & 10.48 & 18.43 & $-$ & 18.43 & \textbf{5.35} & \textbf{3.78} & 5.63 & 7.49 & 15.46 & $-$ & 15.46 \\[1.2pt]
\textsc{Vanilla} + \textbf{\textsc{FDS}} & 9.65 & 6.88 & 9.58 & 11.75 & 19.80 & $-$ & 19.80 & 5.86 & 3.83 & 6.68 & 9.48 & 17.43 & $-$ & 17.43 \\[1.2pt]
\textsc{Vanilla} + \textbf{\textsc{LDS}} + \textbf{\textsc{FDS}} & \textbf{8.96} & 6.88 & \textbf{8.62} & \textbf{10.08} & \textbf{17.76} & $-$ & \textbf{17.76} & 5.38  & 3.90 & \textbf{5.61} & \textbf{7.36} & \textbf{14.65} & $-$ & \textbf{14.65}  \\ \midrule\midrule
\multicolumn{9}{l}{\emph{\textbf{4 peaks:}}} \\ \midrule
\textsc{Vanilla}      & 9.49 & 7.23 & 9.73 & 10.85 & 12.16 & 8.23 & 18.78 & 5.68 & 3.45 & 6.95 & 8.20 & 9.43 & 6.89 & 16.02 \\ [1.2pt]
\textsc{Vanilla} + \textbf{\textsc{LDS}} & 8.80 & \textbf{6.98} & 8.26 & 10.07 & 11.26 & 8.31 & \textbf{16.22} & 5.10 & \textbf{3.33} & \textbf{5.07} & 7.08 & 8.47 & 6.66 & \textbf{12.74} \\[1.2pt]
\textsc{Vanilla} + \textbf{\textsc{FDS}} & 9.28 & 7.11 & 9.16 & 10.88 & 11.95 & 8.30 & 18.11 & 5.49 & 3.36 & 6.35 & 8.15 & 9.21 & 6.82 & 15.30 \\[1.2pt]
\textsc{Vanilla} + \textbf{\textsc{LDS}} + \textbf{\textsc{FDS}} & \textbf{8.76} & 7.07 & \textbf{8.23} & \textbf{9.54} & \textbf{11.13} & \textbf{8.05} & 16.32 & \textbf{5.05} & 3.36 & \textbf{5.07} & \textbf{6.56} & \textbf{8.30} & \textbf{6.34} & 13.10  \\ 
\bottomrule[1.5pt]
\end{tabular}
}
\end{center}
\vspace{-0.2cm}
\end{table}

\subsection{Additional Study on Test Set Label Distributions}
\label{appendix:diff-testset-label-distributions}

We define the evaluation of DIR as generalizing to a testset that is balanced over the entire target range, which is also aligned with the evaluation in the class imbalance setting~\cite{liu2019large}. In this section, we further investigate the performance under different test set label distributions. Specifically, we consider the test set to have exactly the same label distribution as the training set, i.e., the test set also exhibits skewed label distribution (see IMDB-WIKI-DIR in Fig.~\ref{fig:dataset-info}).
We show the results in Table~\ref{table:appendix-diff-test-dist}. As the table indicates, in the balanced testset case, using LDS and FDS can consistently improve the performance of all the regions, demonstrating that our approaches provide a comprehensive and unbiased treatment to all the target values, achieving substantial improvements. Moreover, when the testset has the same label distribution as the training set, we observe that adding LDS and FDS leads to minor degradation in the many-shot region, but drastically boosts the performance in medium-shot and few-shot regions. Note that when testset also exhibits skewed label distribution, the overall performance is dominated by the many-shot region, which can result in biased and undesired evaluation for DIR tasks.

\vspace{-0.2cm}
\begin{table}[ht]
\setlength{\tabcolsep}{6.5pt}
\caption{Additional study of performance on different test set label distributions on IMDB-WIKI-DIR.}
\vspace{4pt}
\label{table:appendix-diff-test-dist}
\small
\begin{center}
\resizebox{0.95\textwidth}{!}{
\begin{tabular}{l|cccc|cccc|cccc}
\toprule[1.5pt]
Metrics      & \multicolumn{4}{c|}{MSE~$\downarrow$}       & \multicolumn{4}{c|}{MAE~$\downarrow$}       & \multicolumn{4}{c}{GM~$\downarrow$}    \\ \midrule
Shot         & All & Many & Med. & Few & All & Many & Med. & Few & All & Many & Med. & Few \\ \midrule\midrule
\multicolumn{9}{l}{\emph{\textbf{Balanced:}}} \\ \midrule
\textsc{Vanilla} & 138.06 & 108.70 & 366.09 & 964.92 & 8.06 & 7.23 & 15.12 & 26.33 & 4.57 & 4.17 & 10.59 & 20.46 \\ [1.2pt]
\textsc{Vanilla} + \textbf{\textsc{LDS}} + \textbf{\textsc{FDS}} & \textbf{129.35} & \textbf{106.52} & \textbf{311.49} & \textbf{811.82} & \textbf{7.78} & \textbf{7.20} & \textbf{12.61} & \textbf{22.19} & \textbf{4.37} & \textbf{4.12} & \textbf{7.39}  & \textbf{12.61}  \\ \midrule\midrule
\multicolumn{9}{l}{\emph{\textbf{Same as training set:}}} \\ \midrule
\textsc{Vanilla} & \textbf{68.44} & \textbf{62.10} & 320.52 & 1350.01 & \textbf{5.84} & \textbf{5.72} & 15.11 & 30.54 & \textbf{3.44} & \textbf{3.40} & 11.76 & 24.06 \\ [1.2pt]
\textsc{Vanilla} + \textbf{\textsc{LDS}} + \textbf{\textsc{FDS}} & 69.86 & 63.43 & \textbf{161.97} & \textbf{1067.89} & {5.90} & {5.77} & \textbf{9.94}  & \textbf{25.17} & {3.48} & {3.44} & \textbf{7.03} & \textbf{15.95} \\
\bottomrule[1.5pt]
\end{tabular}
}
\end{center}
\vspace{-0.2cm}
\end{table}

\subsection{Further Comparisons to Imbalanced Classification Methods}
\label{appendix:compare-imb-classification}

We provide additional study on comparisons to imbalanced classification methods. For DIR tasks that are appropriate (e.g., limited target value ranges), imbalanced classification methods can also be plugged in by discretizing the continuous label space. To gain more insights on the intrinsic difference between imbalanced classification and imbalanced regression problems, we directly apply existing imbalanced classification schemes on several appropriate DIR datasets, and show empirical comparisons with imbalanced regression approaches.
Specifically, we select the subsampled IMDB-WIKI-DIR (see Fig.~\ref{fig:lds-motivate-label-error}), STS-B-DIR, and NYUD2-DIR for comparison. We compare with \textsc{CB}~\cite{cui2019class} and \textsc{cRT}~\cite{kang2020decoupling}, which are the state-of-the-art methods for imbalanced classification. We also denote the vanilla classification method as \textsc{Cls-vanilla}. For fair comparison, the classes are set to the same bins used in LDS and FDS.
Table~\ref{table:appendix-compare-imb-classification} confirms that LDS and FDS outperform imbalanced classification schemes by a large margin across all DIR datasets, where the errors for few-shot regions can be reduced by up to $50\%$ to $60\%$. Interestingly, the results also show that imbalanced classification schemes often perform \emph{worse} than even the vanilla regression model (i.e., \textsc{Reg-vanilla}), which confirms that regression requires different approaches for data imbalance than simply applying classification methods.

We note that imbalanced classification methods could fail on regression problems for several reasons. First, they ignore the similarity between data samples that are close w.r.t. the continuous target; Treating different target values as distinct classes is unlikely to yield the best results because it does not take advantage of the similarity between nearby targets. Moreover, classification methods cannot extrapolate or interpolate in the continuous label space, therefore unable to deal with missing data in certain target regions.

\vspace{-0.1cm}
\begin{table}[ht]
\setlength{\tabcolsep}{7pt}
\caption{Additional study on comparisons to imbalanced classification methods across several appropriate DIR datasets.}
\vspace{4pt}
\label{table:appendix-compare-imb-classification}
\small
\begin{center}
\resizebox{0.95\textwidth}{!}{
\begin{tabular}{l|cccc|cccc|cccc}
\toprule[1.5pt]
Dataset & \multicolumn{4}{c|}{IMDB-WIKI-DIR (subsampled)} & \multicolumn{4}{c|}{STS-B-DIR} & \multicolumn{4}{c}{NYUD2-DIR} \\ \midrule
Metric & \multicolumn{4}{c|}{MAE~$\downarrow$} & \multicolumn{4}{c|}{MSE~$\downarrow$} & \multicolumn{4}{c}{RMSE~$\downarrow$} \\ \midrule
Shot & All & Many & Med. & Few & All & Many & Med. & Few & All & Many & Med. & Few \\ \midrule\midrule
\multicolumn{9}{l}{\emph{\textbf{Imbalanced Classification:}}} \\ \midrule
\textsc{Cls-vanilla} & 15.94 & 15.64 & 18.95 & 30.21 & 1.926 & 1.906 & 2.022 & 1.907 & 1.576 & 0.596 & 1.011 & 2.275 \\ [1.2pt]
\textsc{CB}~\cite{cui2019class} & 22.41 & 22.32 & 22.05 & 32.90 & 2.159 & 2.194 & 2.028 & 2.107 & 1.664 & 0.592 & 1.044 & 2.415 \\ [1.2pt]
\textsc{cRT}~\cite{kang2020decoupling} & 15.65 & 15.33 & 17.52 & 29.54 & 1.891 & 1.906 & 1.930 & 1.650 & 1.488 & 0.659 & 1.032 & 2.107 \\ \midrule\midrule
\multicolumn{9}{l}{\emph{\textbf{Imbalanced Regression:}}} \\ \midrule
\textsc{Reg-vanilla} & 14.64 & 13.98 & 17.47 & 30.29 & 0.974 & 0.851 & 1.520 & 0.984 & 1.477 & \textbf{0.591} & 0.952 & 2.123 \\ [1.2pt]
\textsc{LDS} & {14.03} & {13.72} & {15.93} & {26.71} & 0.914 & 0.819 & 1.319 & 0.955 & 1.387 & 0.671 & 0.913 & 1.954 \\ [1.2pt]
\textsc{FDS} & {13.97} & {13.55} & {16.42} & {24.64} & 0.916 & 0.875 & \textbf{1.027} & 1.086 & 1.442 & 0.615 & 0.940 & 2.059 \\ [1.2pt]
\textsc{LDS + FDS} & \textbf{13.32} & \textbf{13.14} & \textbf{15.06} & \textbf{23.87} & \textbf{0.907} & \textbf{0.802} & 1.363 & \textbf{0.942} & \textbf{1.338} & 0.670 & \textbf{0.851} & \textbf{1.880} \\
\bottomrule[1.5pt]
\end{tabular}
}
\end{center}
\vspace{-0.1cm}
\end{table}

\subsection{Complete Visualization for Feature Statistics Similarity}
\label{appendix:visualize-feat-sim-all-ages}

\begin{figure*}[ht]
\centering
\subfigure[Feature statistics similarity for age $0$, without FDS]{
    \label{fig:feat_sim_fds_base_0}
    \includegraphics[height=0.265\textwidth]{figures/feat_sim_fds_base_0.pdf}
}
\hspace{2.5ex}
\subfigure[Feature statistics similarity for age $0$, with FDS]{
    \label{fig:feat_sim_fds_ours_0}
    \includegraphics[height=0.265\textwidth]{figures/feat_sim_fds_ours_0.pdf}
}\\
\subfigure[Feature statistics similarity for age $30$, without FDS]{
    \label{fig:feat_sim_fds_base_30}
    \includegraphics[height=0.265\textwidth]{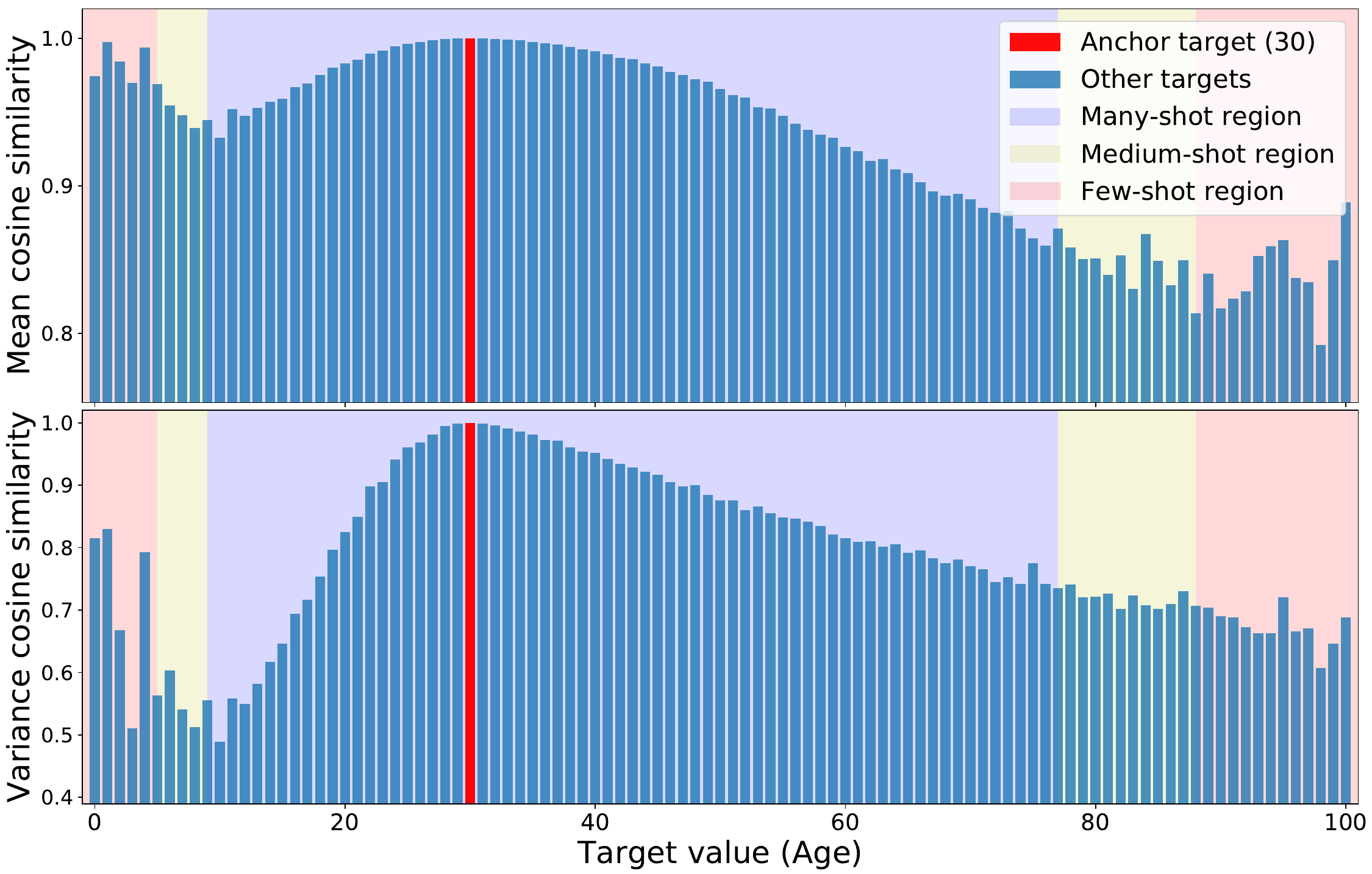}
}
\hspace{2.5ex}
\subfigure[Feature statistics similarity for age $30$, with FDS]{
    \label{fig:feat_sim_fds_ours_30}
    \includegraphics[height=0.265\textwidth]{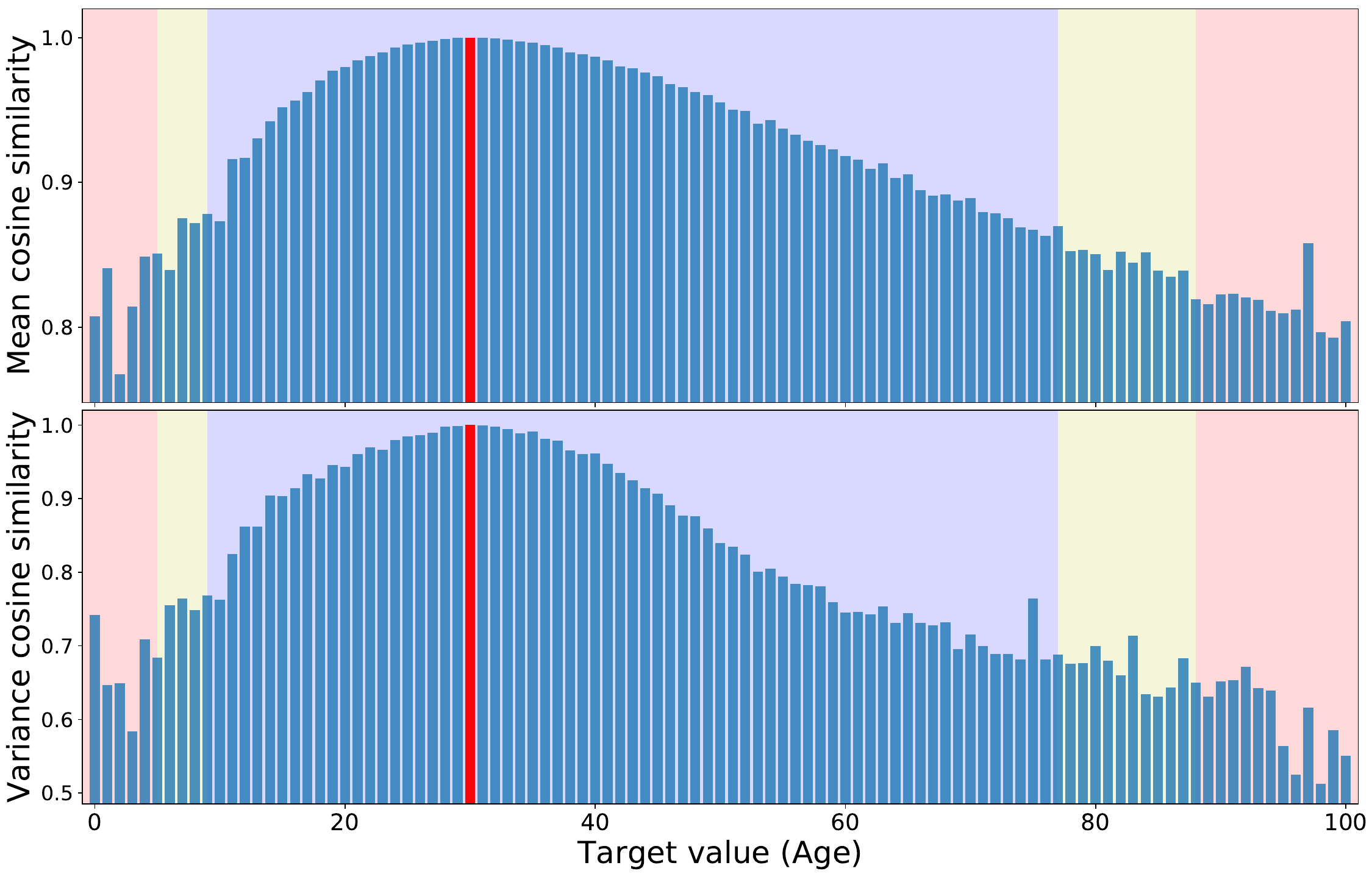}
}\\
\subfigure[Feature statistics similarity for age $60$, without FDS]{
    \label{fig:feat_sim_fds_base_60}
    \includegraphics[height=0.265\textwidth]{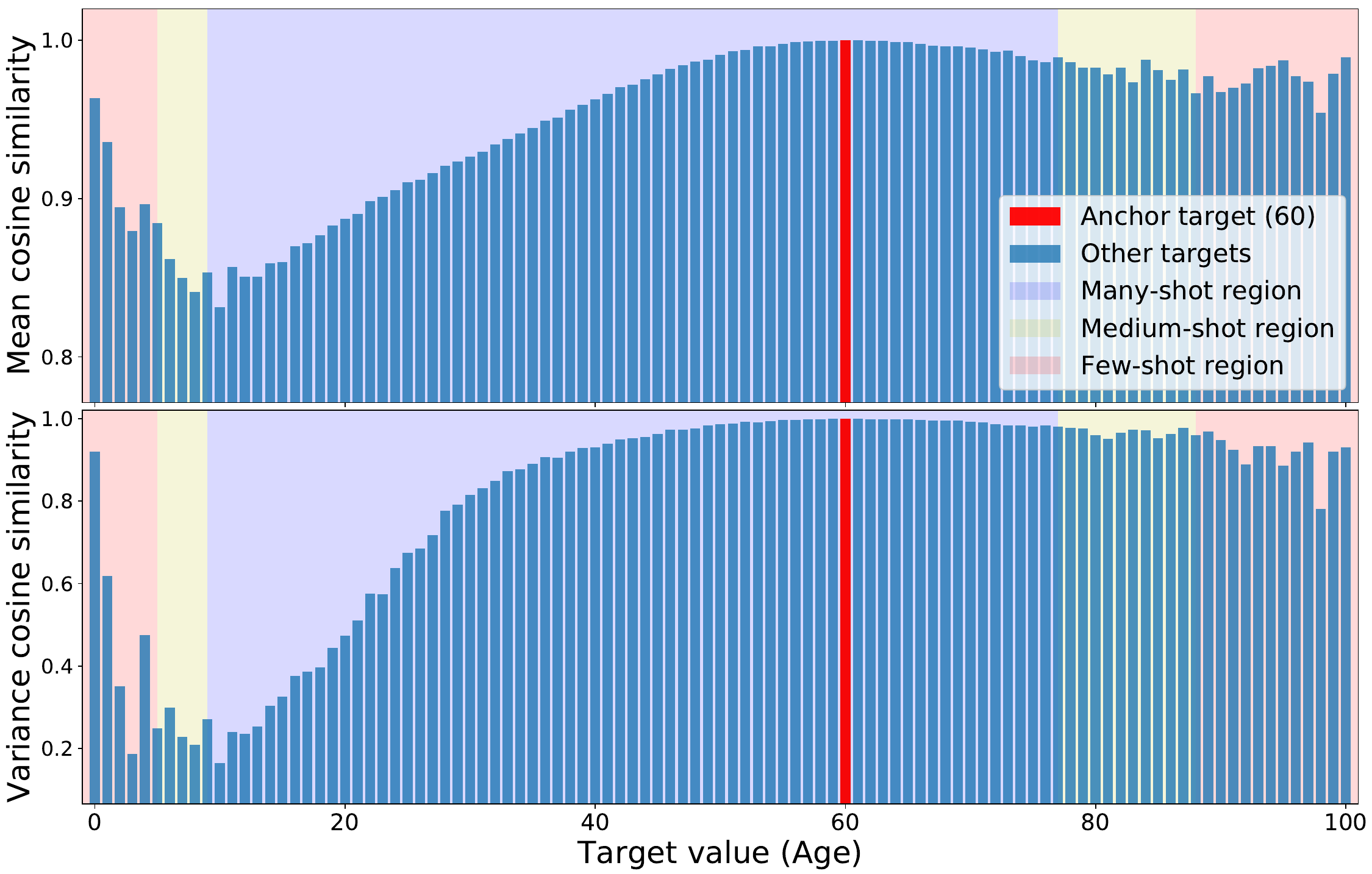}
}
\hspace{2.5ex}
\subfigure[Feature statistics similarity for age $60$, with FDS]{
    \label{fig:feat_sim_fds_ours_60}
    \includegraphics[height=0.265\textwidth]{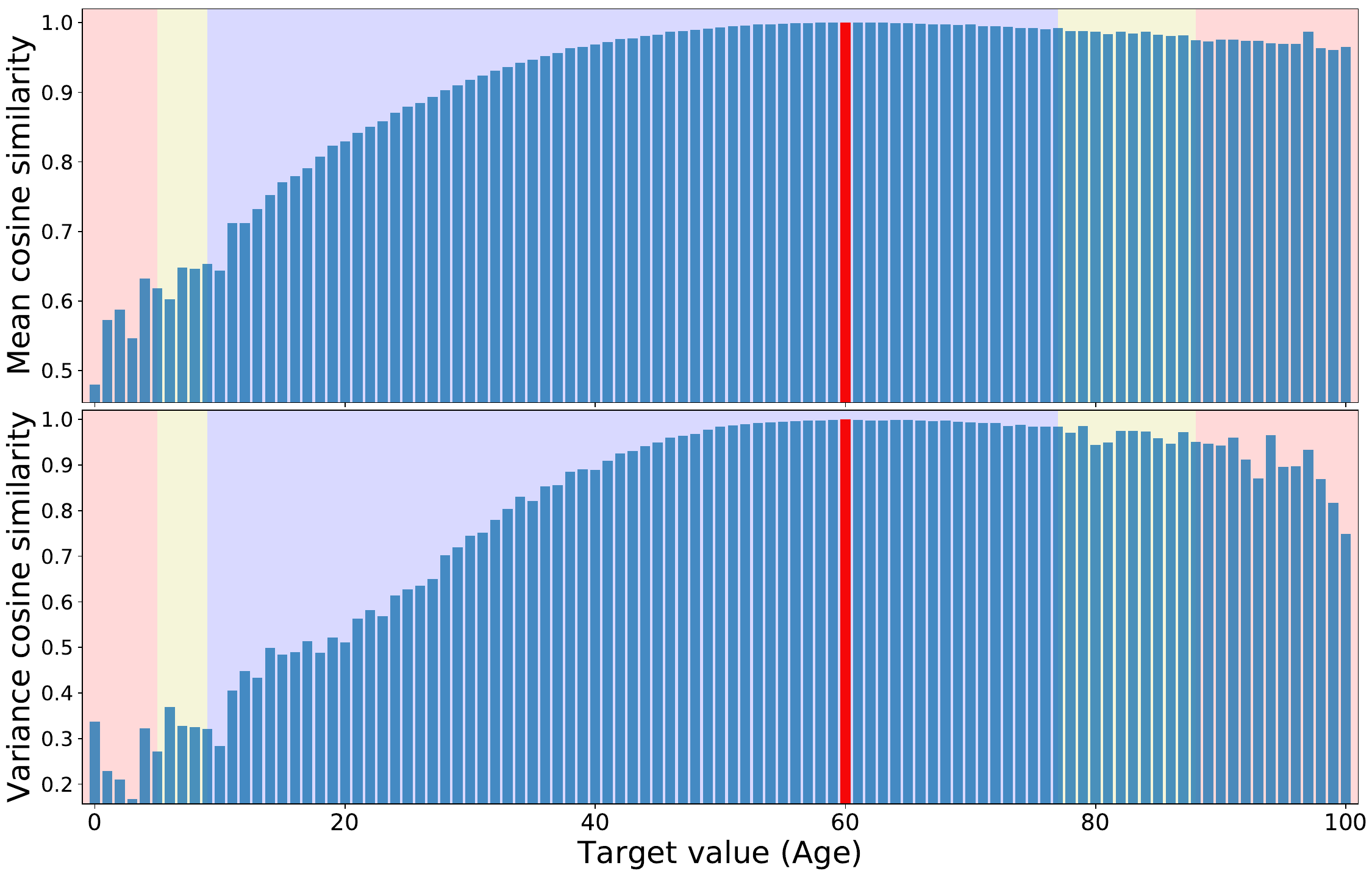}
}\\
\subfigure[Feature statistics similarity for age $90$, without FDS]{
    \label{fig:feat_sim_fds_base_90}
    \includegraphics[height=0.265\textwidth]{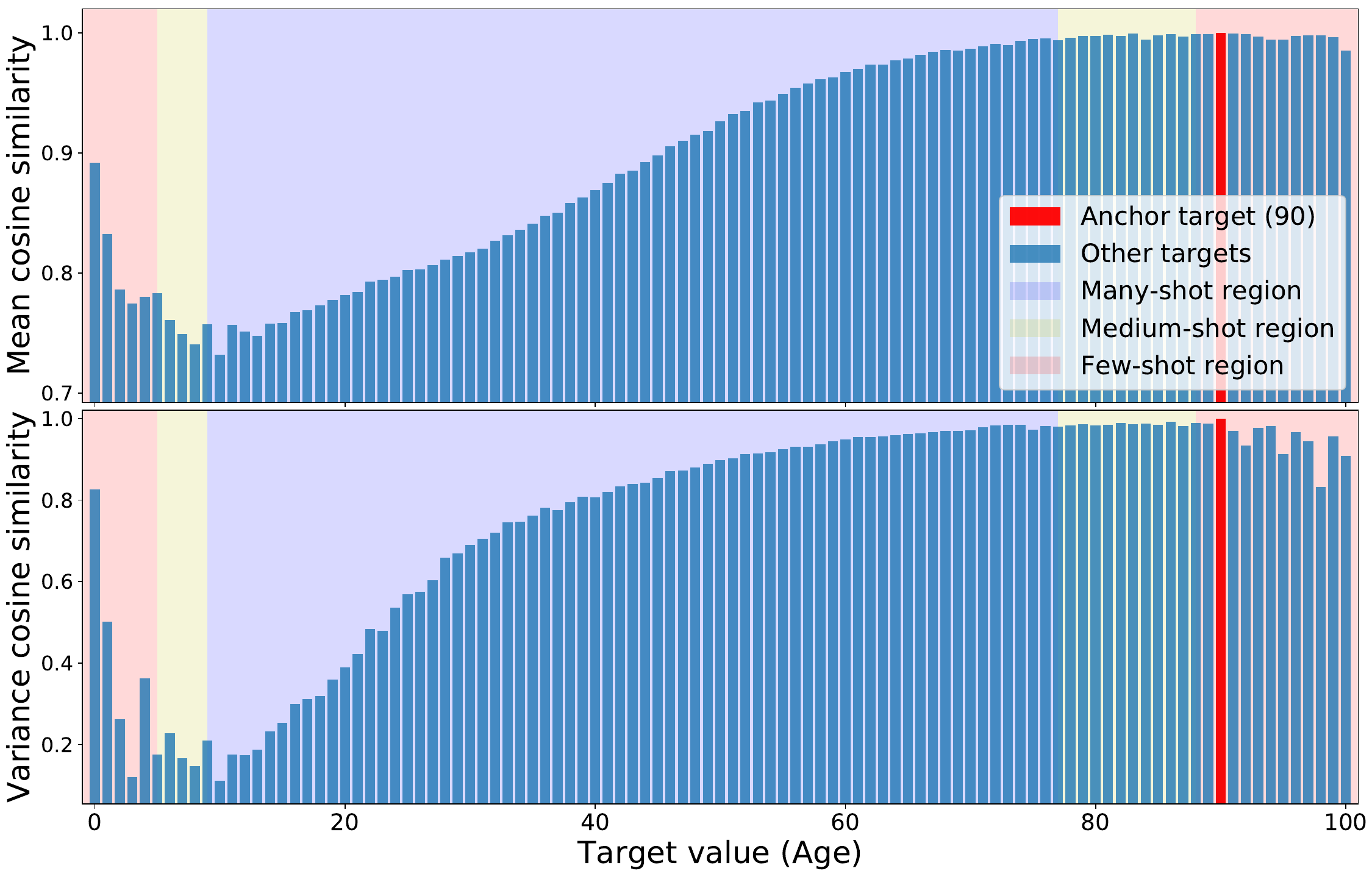}
}
\hspace{2.5ex}
\subfigure[Feature statistics similarity for age $90$, with FDS]{
    \label{fig:feat_sim_fds_ours_90}
    \includegraphics[height=0.265\textwidth]{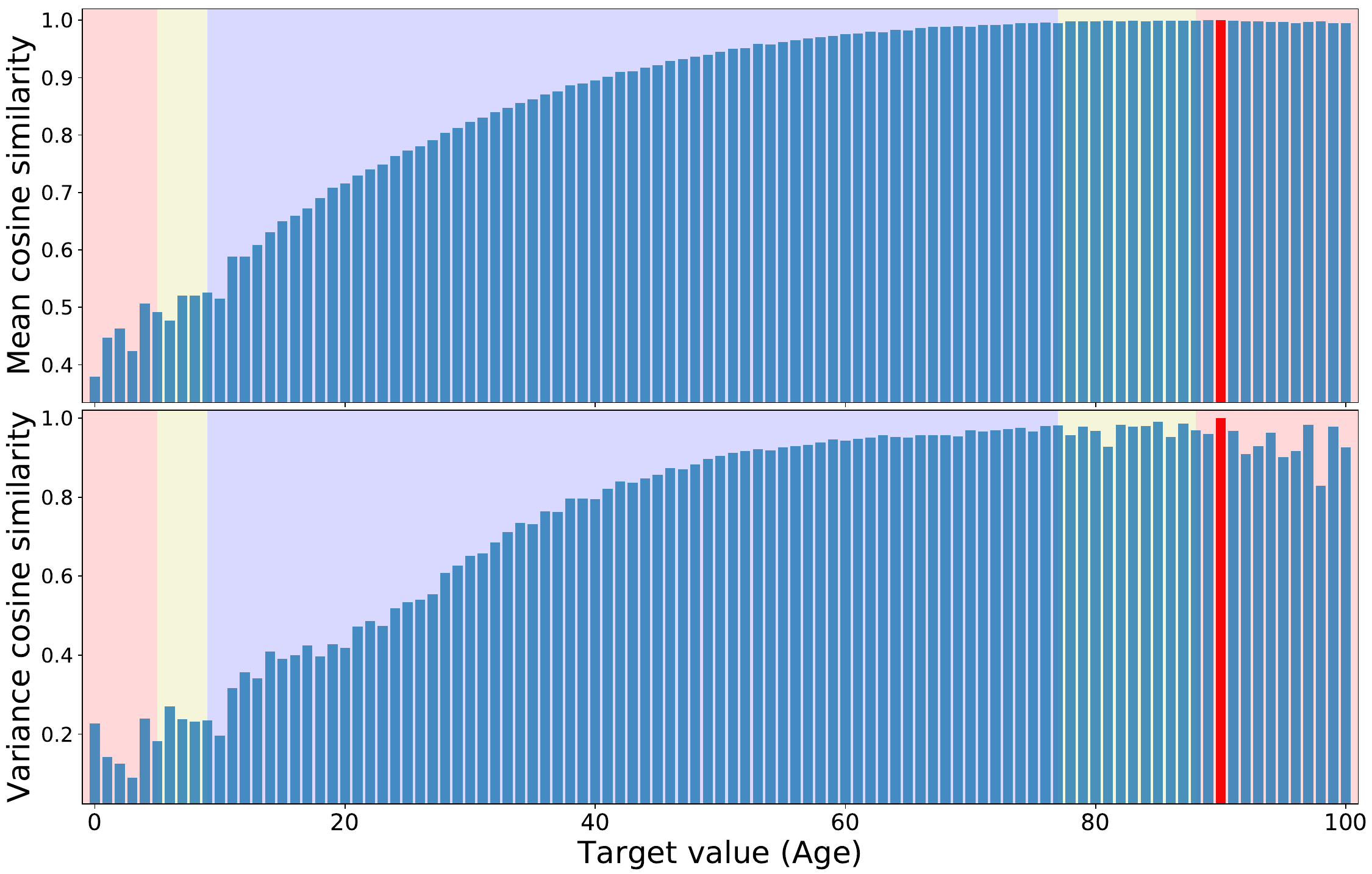}
}
\vspace{-0.3cm}
\caption{Analysis on how FDS works. \textbf{First column:} Feature statistics similarity for anchor ages $\{0,30,60,90\}$, using model trained without FDS. \textbf{Second column:} Feature statistics similarity for anchor ages $\{0,30,60,90\}$, using model trained with FDS. We show that using FDS, the statistics are better calibrated for all anchor ages, leading to a high similarity only in the neighborhood, and a gradually decreasing similarity score as target value becomes smaller or larger.
}
\label{fig:appendix-fds-analysis-full}
\end{figure*}

We provide additional results for understanding FDS, i.e., how FDS influences the feature statistics. In Fig.~\ref{fig:appendix-fds-analysis-full}, we plot the similarity of the feature statistics for different anchor ages in $\{0,30,60,90\}$, using models trained without and with FDS.
As the figure indicates, for the vanilla model (i.e., Fig.~\ref{fig:feat_sim_fds_base_0}, \ref{fig:feat_sim_fds_base_30}, \ref{fig:feat_sim_fds_base_60}, and \ref{fig:feat_sim_fds_base_90}), there exists unexpected high similarities between the anchor ages and the regions that have very few data samples. For example, in Fig.~\ref{fig:feat_sim_fds_base_0} where the anchor age is 0, the highest similarity is obtained with age range between 40 and 80, rather than its nearby ages. Moreover, for anchor ages that lie in the many-shot regions (e.g., Fig.~\ref{fig:feat_sim_fds_base_30}, \ref{fig:feat_sim_fds_base_60}, and \ref{fig:feat_sim_fds_base_90}), they also exhibit unjustified feature statistics similarity with samples from age range 0 to 6, which is due to data imbalance.
In contrast, by adding FDS (i.e., Fig.~\ref{fig:feat_sim_fds_ours_0}, \ref{fig:feat_sim_fds_ours_30}, \ref{fig:feat_sim_fds_ours_60}, and \ref{fig:feat_sim_fds_ours_90}), the statistics are better calibrated for all anchor ages, leading to a high similarity only in the neighborhood, and a gradually decreasing similarity score as target value becomes smaller or larger.

\end{document}